\documentclass{article}

\usepackage[final]{neurips_2025}

\usepackage[utf8]{inputenc} % allow utf-8 input
\usepackage[T1]{fontenc}    % use 8-bit T1 fonts
\usepackage{hyperref}       % hyperlinks
\usepackage{url}            % simple URL typesetting
\usepackage{booktabs}       % professional-quality tables
\usepackage{amsfonts}       % blackboard math symbols
\usepackage{nicefrac}       % compact symbols for 1/2, etc.
\usepackage{microtype}      % microtypography
\usepackage{xcolor}         % colors
\usepackage{soul}           % for the command \hl
\usepackage{graphicx}
\usepackage{amsmath}
\usepackage{amssymb}
\usepackage{wrapfig}
\usepackage{caption}
\DeclareMathOperator{\E}{\mathbb{E}}
\newcommand{\name}{\texttt{EchoNeRF}}
\newcommand{\red}[1]{#1}

\newcommand{\NeRFs}{\texttt{NeRF2}}
\newcommand{\NeWRF}{\texttt{NeWRF}}
\newcommand{\MLP}{\texttt{MLP}}
\newcommand{\ZYBase}{\texttt{Heatmap Segmentation}}
\newcommand{\nameLoS}{\name\_\texttt{LoS}}

\newcommand{\Tx}{\textit{Tx}}
\newcommand{\Rx}{\textit{Rx}}
\renewcommand{\hl}[1]{#1}
\DeclareMathOperator*{\argmin}{argmin}

\title{Can NeRFs \hspace{0.01pt} ``See'' \hspace{0.01pt} without Cameras?}

\author{
\begin{tabular}{ccc}
\textbf{Chaitanya Amballa}\textsuperscript{1} &
\textbf{Sattwik Basu}\textsuperscript{1}\thanks{Equal contribution.} &
\textbf{Yu\mbox{-}Lin Wei}\textsuperscript{1}\footnotemark[1] \\
\texttt{amballa2@illinois.edu} &
\texttt{sattwik2@illinois.edu} &
\texttt{yulinlw2@illinois.edu} \\
\\[-0.25em]
\textbf{Zhijian Yang}\textsuperscript{2} &
\textbf{Mehmet Ergezer}\textsuperscript{2} &
\textbf{Romit Roy Choudhury}\textsuperscript{1,2} \\
\texttt{yzhijian@amazon.com} &
\texttt{mergezer@amazon.com} &
\texttt{croy@illinois.edu}
\vspace{0.2cm}
\end{tabular}
\\[0.45in]
\textsuperscript{1}University of Illinois Urbana–Champaign \quad
\textsuperscript{2}Amazon
}

\begin{document}

\maketitle

\vspace{-0.4cm}
\begin{abstract}
Neural Radiance Fields (NeRFs) have been remarkably successful at synthesizing novel views of 3D scenes by optimizing a volumetric scene function.
This scene function models how optical rays bring color information from a 3D object to the camera pixels.
% bring colors from the environment and deliver to the camera pixels they impinge upon.
% This scene function models how an optical ray accumulates colors on its path and eventually delivers this color to the camera pixel it impinges upon.
Radio frequency (RF) or audio signals can also be viewed as a vehicle for delivering information about the environment to a sensor.
However, unlike camera pixels, an RF/audio sensor receives a mixture of signals that contain many environmental reflections (also called ``multipath'').
% signal rays that arrive after reflections in the environment.
Is it still possible to infer the environment using such multipath signals?
We show that with redesign, NeRFs can be taught to learn from multipath signals, and thereby ``see'' the environment.
As a grounding application, we aim to infer the indoor floorplan of a home from sparse WiFi measurements made at multiple locations inside the home.
Although a difficult inverse problem, our implicitly learnt floorplans look promising, and enables forward applications, such as indoor signal prediction and basic ray tracing.
% and our model, {\name}, enables various forward 
% downstream signal prediction applications.
% Our work also uncovers new research pathways for NeRF-based wireless imaging.
% \hl{Say simulator early on to set expectations.}
% Is it possible to infer the environment strength across the room using mixed measurements?
% We show that with redesign, the core NeRF framework can model the physical reflections and solve this inverse problem. 
% We focus on a specific application of inferring the indoor floorplan of a home from WiFi measurements made at multiple locations inside the home.
% Our inferred floorplans look promising and benefits downstream signal prediction applications.
% Our work also uncovers several new questions for further research.
\end{abstract}

\vspace{-0.15in}
\section{Introduction}
% \textbf{Motivation.} 
NeRFs \cite{mildenhall2021nerf, nerfInvProb, mipnerf, pixelnerf} have delivered impressive results in solving inverse problems, resulting in 3D scene rendering.
While NeRFs have mostly used pictures (from cameras or LIDARs) to infer a 3D scene, we ask if the core ideas can generalize to the case of wireless signals (such as RF or audio).
% As a specific instance, we are interested in inferring the floorplan of an indoor environment by measuring wireless signals from a mobile sensor.
% One could envision this as a user walking around with a phone in a home; can the phone record ambient WiFi signals (or audio music played from loudspeakers) and train a NeRF to implicitly learn the home’s floorplan?
% Generalization would mean implicitly learning the scene from wireless signals that arrive after reflecting off the scene.
% Generalization means that the NeRF is implicitly able to learn the scene using wireless signals that arrive after reflecting off of that scene.
Unlike camera pixels that receive line-of-sight (LoS) rays, a wireless receiver (e.g., a WiFi antenna on a smartphone) would receive a mixture of LoS and many reflections, called {\em multipath}.
If the receiver moves, it receives a sequence of $N$ measurements.
Using these $N$ wireless measurements, is it possible to learn a representation of the scene, such as the floorplan of the user's home?
% \footnote{Imagine a user walking inside a home with a smartphone; can the phone record ambient WiFi signals and train a NeRF to implicitly learn the home’s floorplan?}.
A vanilla NeRF understandably fails since it is not equipped to handle multipath.
This paper is focused on redesigning NeRFs so they can learn to image the environment, thereby solving the inverse problem from ambient wireless signals.
% Generalizing optical NeRFs to wireless sensing seems plausible since wireless measurements are just observations from inside the scene, while NeRF images are typically taken from the outside. 

A growing body of work \cite{specnerf, winert, zhao2023nerf2, newrf, luo2022learning, AV-NeRF} is investigating connections between NeRFs and wireless.
% has adopted NeRFs to wireless applications.
While none have concentrated on imaging, NeRF2 \cite{zhao2023nerf2} and NeWRF \cite{newrf} have augmented NeRFs to correctly synthesize WiFi signals at different locations inside an indoor space.
However, correct synthesis is possible without necessarily 
learning the correct signal propagation models.
% In other words, it is possible to learn an implicit representation of the scene that does not align with signal propagation models. 
We find that NeRFs with adequate model complexity can overfit a function to correctly predict signals at test locations, but this function does not embed the true behavior of multipath signal propagation.
We re-design the NeRF's objective function so that it learns the environment through line-of-sight (LoS) paths and reflections.
This teaches the NeRF an implicit representation of the scene, which can then be utilized for various forward tasks, including WiFi signal prediction and ray tracing.
% and, in turn, trains the NeRF to learn the environment's floorplan.
% % Our goal is to design a multipath power function  utilize the inherent potential of NeRFs by  to implicitly learn the environment walls that correctly explain the received signals.
% Once learnt, the floorplan can be utilized for various downstream tasks, such as signal prediction or basic ray tracing. 

In our model, {\name}, each voxel is parameterized by its opacity $\delta \in [0,1]$ and orientation $\omega \in [-\pi,\pi]$.
When trained perfectly, free-space air voxels should be transparent ($\delta=0$), wall voxels should be opaque ($\delta=1$), and each opaque voxel's orientation should match its wall's orientation.
As measurements, we use the received {\em signal power}.
% easily available from any receiver's hardware. 
% which to implicitly learn the floorplan representation.
% Our expected output is a 2D/3D floorplan as shown in Figure \ref{fig:bigpicture}.
Thus, the input to our {\name} model is the transmitter ($Tx$) location, a sequence of known receiver ($Rx$) locations, and the signal power measured at each $Rx$ location. 
The output of {\name} is an (implicitly learnt) floorplan of the indoor space.
We expect to visualize the floorplan by plotting the learnt voxel opacities.

\begin{figure}[!h]
  \centering
  \begin{minipage}{0.17\textwidth}
  \includegraphics[width=.9\columnwidth]{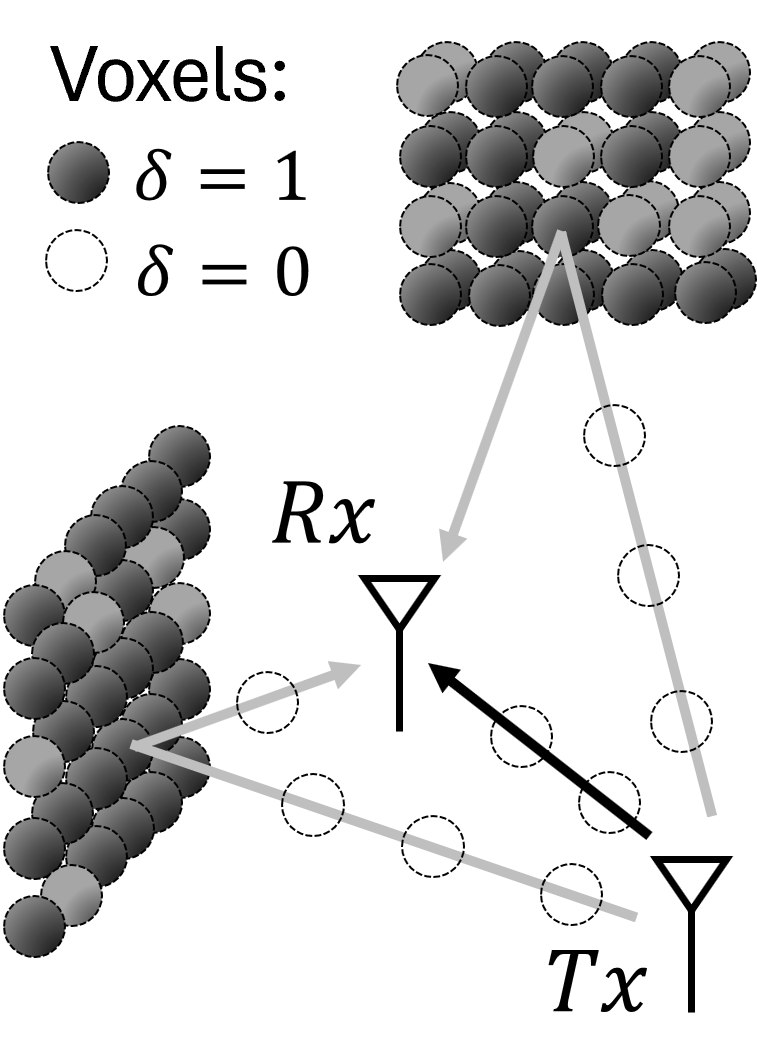} 
  \caption{LoS and correct multipath reflections.}
  \vspace{-0.05in}
  \label{fig:multipath}
  \end{minipage}
  \hfill
  \begin{minipage}{0.8\textwidth}
  \includegraphics[width=\columnwidth]{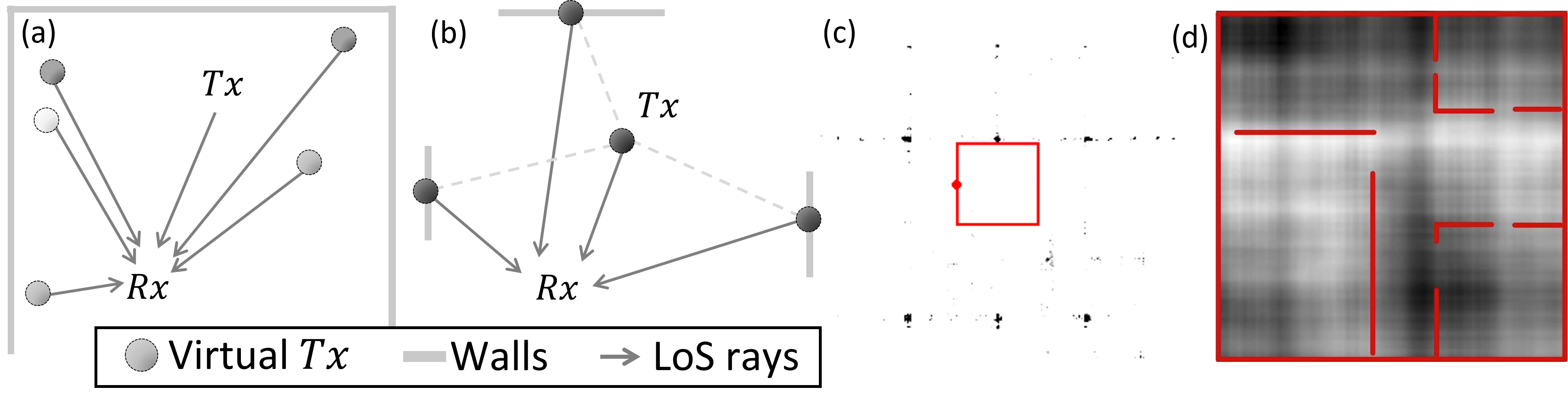} 
    \caption{
(a) Fitting signal power using virtual $Tx$s. (b) Ideally, virtual $Tx$s should be located on wall surfaces. (c) Sparse virtual $Tx$s learnt by NeWRF shown in black/gray dots. (d) Dense virtual $Tx$s learnt by NeRF2. 
Neither correspond to the true (red) floorplan.}
  \vspace{-0.05in}
  \label{fig:virtualTx}
  \end{minipage}
\end{figure}

% \begin{wrapfigure}{r}{0.\textwidth}
%   \includegraphics[width=0.48\textwidth]{ICCV2025-Author-Kit-Feb/figs/voxel_rf.png} 
%   \caption{Multipath signals through voxel lattice.}
% %   \vspace{-0.1in}
%   \label{fig:voxel}
% \end{wrapfigure}

Learning floorplans requires modeling the correct reflections (see Fig. \ref{fig:multipath}) since these reflections help reveal where the walls are. 
However, without knowledge of the walls, the reflections are difficult to model, leading to a type of chicken and egg problem.
Additionally, the number of wireless measurements is relatively sparse compared to the number of pixels measured in image-trained NeRFs.
% since a user may walk around with her phone for only a few minutes. 
% we envision a user walking around with a phone for a few minutes in her home, giving us say $M=800$ measurements along a randomly walked trajectory). 
Finally, measured signals will have ``blind spots'', meaning that rays that bounced off certain regions of the walls may not have arrived at any of the $Rx$ locations.
This leaves gaps or holes in the floorplan and NeRF's interpolation through these gaps will produce error or blur.

% \begin{wrapfigure}{r}{0.42\textwidth}
%   \centering
%   \includegraphics[width=0.4\textwidth]{ICCV2025-Author-Kit-Feb/figs/voxel_rf.png} 
%   \vspace{-0.1in}
%   \caption{Multipath signals through voxel lattice.}
%   \vspace{-0.1in}
%   \label{fig:voxel}
% \end{wrapfigure}

{\name} approaches this problem by modeling the received signal power as a combination of the LoS power and the power from first order reflections. 
The LoS model is inherited from classical NeRFs.
The main departure from past work is in modeling the reflections.
Since opaque voxels are unknown during training, the reflection surfaces are not known; hence, the reflection power at the $Rx$ is modeled as an aggregate over all plausible reflections.
Given the planar structure of walls, the plausible set of reflections can be heavily pruned to reduce the optimization complexity.
% Fortunately, for a given $\langle Tx, Rx \rangle$ pair, the voxels that can cause reflections lie on a geometric manifold (under reasonable assumptions), reducing the plausible set. 
Reflections aggregated over this plausible set models the total (LoS + reflection) power at a receiver $Rx$.
% The reflections are aggregated over this plausible set to finally model the total (LoS + reflection) power. 

{\name} trains to minimize the loss between the modeled and measured power across all $Rx$ locations, and in the process, learns the voxel's opacities that best explain the measured dataset.
Some regularization is necessary to cope with sparse measurements and to ensure smoothness of walls.
% To cope with sparse measurements, regularizations are added to enforce smoothness among local voxels; a penalty is imposed to prevent learning multiple reflections from one manifold. 
% orientations
% sparsity of voxels along a ray, smoothness among voxel orientations, and a penalty against learning too many reflections from the manifold.
Lastly, to handle some gradient imbalance issues, {\name} freezes the LoS model once it converges, and uses this intermediate state to partly supervise the reflection model.

To evaluate {\name}, we train on $2.4$ GHz WiFi signals from NVIDIA's Sionna simulator \cite{hoydis2023sionna}, with floorplans from the Zillow's Indoor Dataset (ZIND) \cite{cruz2021zillow}.
Results show consistent improvement over baselines in terms of the estimated floorplan's IoU and F1 score.
% ; using the estimated floorplan, we show that {\name} can better predict the signal power at unknown locations.
% based on the learnt floorplan, our signal  even though the baselines use the more informative impulse response data.
Qualitative results show visually legible floorplans without any post-processing.
Applying forward functions on the floorplan, {\name} can predict the received signal power for new $\langle Tx, Rx \rangle$ locations (outperforming existing baselines).
Lastly, basic ray tracing explains the predictions, offering interpretability to its results.

\section{Related Work and Research Scope}
\label{sec:related}
\vspace{-0.1in}

\textbf{Wireless (WiFi) channel prediction using NeRFs.} 
\NeWRF \cite{newrf} and \NeRFs \cite{zhao2023nerf2} are recent papers that have used NeRFs to predict the wireless channel impulse response (CIR) \cite{Tse_Viswanath_2005} at unknown locations inside a room.
% Both utilize NeRFs to predict the wireless CIR.
Drawing a parallel to optical NeRFs, a voxel's color in optics becomes a voxel's transmit power in wireless. 
The voxel's density in optics remains the same in wireless, modeling how that voxel attenuates signals passing through it.
NeRF2 and NeWRF {\em assign transmit power and attenuation to each voxel} such that they best explain the measured CIR.
% (i.e., minimize $L_2$ loss over all training locations inside a room).
The authors explain that voxels assigned non-zero transmit power will be called {\em virtual transmitters}; these voxels represent the reflection points on the walls.
 % and will model the reflections in the environment.
% -- Figure \ref{fig:virtualTx} shows the expected assignment.
However, {\em many assignments are possible that fit the CIR training data}, especially when the data is sparse. Fig. \ref{fig:virtualTx}(a,b) illustrates 2 possible assignments.
While the predicted CIRs could achieve low error for all such assignments, only one of the assignments will model the true reflections, forcing the virtual transmitters to be located on the wall surfaces (Fig. \ref{fig:virtualTx}(b)).
We have plotted {\NeRFs} and {\NeWRF}'s assignment of voxel densities (see Fig. \ref{fig:virtualTx}(c,d)) to confirm that the high accuracy in CIR prediction is not an outcome of correctly learning the wall layout.
Our goal is to repair this important issue, i.e., assign voxel densities that obey the basic physics of wall reflections.
Correct voxel assignment leads to the correct layouts, which then makes the (forward) CIR prediction easy.

% measured signal powers, and basic multipath propagation physics, are both satisfied. 
% Our goal is to assign the voxel densities correctly using the simple NeRF framework, which should naturally yield the floorplan.

% This characterizes the room for research and, consequently, {\name}'s main contribution.
% We have plotted NeRF2 and NeWRF's assignment of virtual transmitters (see Figure \ref{}) to show that the floorplan in not learnt in the process of CIR prediction.
% {\name}'s model is designed to place the virtual transmitters on wall surfaces, thus learning the floorplan and the true behavior of multipath signal propagation.

% \begin{figure}[!h]
%   \centering
%   \includegraphics[width=\columnwidth]{figs/virtualTx2.png}
%   \caption{Modeling wireless reflections using virtual transmitters.  Ideally, the virtual $Tx$'s should be located on the wall surfaces.}
%   \vspace{-0.1in}
%   \label{fig:virtualTx}
% \end{figure}

% \begin{figure}[!h]
%   \centering
%   \includegraphics[width=0.4\columnwidth]{figs/NeWRF.png}
%   \hfill
%   \includegraphics[width=0.4\columnwidth]{figs/eval/main/3_N2.png}
%   \caption{Virtual transmitters learnt by (a) NeWRF and (b) NeRF2, shown as black/gray dots. The true floorplan is overlaid in red.}
%   \label{fig:others}
% \end{figure}

\textbf{Neural radiance fields for audio.}
% NAF followed by NACF, AV-NeRF, NeRAF
Another active line of research focuses on predicting room impulse response (RIR) for audio \cite{luo2022learning,INRAS, AV-NeRF,brunetto2024neraf}. 
Neural Acoustic Field (NAF) \cite{luo2022learning} extended the classical NeRF to train on RIR measurements in a room and predict the RIR (magnitude and phase) at new $\langle Tx, Rx \rangle$ locations.
NAF identified the possibility of overfitting to the RIR and proposed to learn, jointly, the local geometric features of the environment (as spatial latents) and the NAF parameters.
The spatial latents embed floorplan information but a decoder needs to be trained using the partial floorplan data.
{\name} requires no floorplan supervision, and secondly, relies entirely on signal power (less informative than RIR) to estimate the floorplan.

Follow up work are embracing more information about the surroundings (pictures \cite{NACF, INRAS, AV-NeRF, fewshotvisual}, LIDAR scans  \cite{oufqir2020arkit}, meshes and optical NeRFs \cite{brunetto2024neraf}), to boost RIR accuracy. 
Results are steadily improving, however, this sequence of ideas is unaligned with solving the core inverse problem.
Our goal is to first invert the signal power to a floorplan, which can then enable CIR/RIR predictions.

% Follow up work have focused more on improving the RIR accuracy and have utilized explicit information about the surroundings, namely pictures \cite{NACF, INRAS, AV-NeRF, fewshotvisual}, LIDAR scans \cite{oufqir2020arkit}, or optical NeRFs \cite{brunetto2024neraf}, to better guide their MLPs.
% RIR prediction results are steadily improving, however, this sequence of ideas are veering away from the inverse nature of the problem.
% Our goal is to first invert the signal power to a floorplan, which can then enable CIR/RIR predictions.
% % Our priority is to learn the signal propagation and the floorplan first, and then infer the CIR/RIR from them in the future.

\textbf{Modeling reflections in optical NeRFs}
Optical NeRFs have tackled reflections \cite{refneus, nerfds, flipnerf} for synthesizing glossy surfaces and mirrors, and for re-lighting \cite{rudnev2022nerf, relight}.
NeRFRen \cite{nerfren} proposes to decompose a viewed image into a transmitted and a reflected component.
% To synthesize novel views, the two-component images (learnt by separated branches of the MLP) are added with learnt weights to appropriately suppress the reflected component. 
Ref-NeRF \cite{verbin2022refnerf} also focuses on reflections through a similar decomposition of the transmitted and reflected color, however, the reflected color is modeled as a function of the viewing angle and the surface-normal, resulting in accurate models of specular reflection.
% Recent works like NeRF-Casting \cite{nerfcasting} and oRCA \cite{orca} have further improved multipath modeling on glossy and mirrored surfaces.
% Several other papers \cite{boss2021nerd, zhang2021nerfactor} have developed similar ideas, with the core insight centering around solving a two-component decomposition problem.
\red{Recent proposals such as NeRF-Casting \mbox{\cite{nerfcasting}} and oRCA \mbox{\cite{orca}} have further improved the models of multipath for glossy and mirrored surfaces, and others \mbox{\cite{boss2021nerd, zhang2021nerfactor}} have developed similar ideas, with the core insights centering around solving a two-component decomposition problem.}
% Several other papers have developed similar ideas \cite{boss2021nerd, zhang2021nerfactor} and the core insight centers around solving a two-component decomposition problem.
{\name} faces the challenge of not knowing the number of rays adding up from all possible directions in the environment.
Hence, {\name} must solve a many-component decomposition problem by leveraging the physics of multipath signal propagation. \\
% (an \hl{RF} specific opportunity).
% Finally, floorplan estimation, as a technology, has matured significantly with RGBD cameras \cite{roomnet,  layoutnet, 3dlayout, jia20223d, zhang2019edge, stekovic2020general}, LIDARs \cite{ericson2024beyond, xiao2014reconstructing}, and even smartphone cameras (in Apple's ARKit) \cite{apple2017arkit}. 
% However, using visual sensors indoors can be invasive to the user's privacy; 
% {\name} aims to infer floorplans through less invasive RF signals.
% % a user to walk to every corner of the home.
% More importantly, we are keen on helping NeRFs learn wireless signal propagation, especially $1^{st}$ order reflections; floorplan estimation is one application of this learning exercise.
\red{In the context of LIDARs, PlatoNeRF \mbox{\cite{platonerf}} and NeTF \mbox{\cite{nonlos}} cope with unknown reflections, however, LIDARs have very high time resolution (due to high clock frequency), and is therefore able to assign the incoming rays to different time buckets. 
This temporal separation allows the NeRF to make separate measurements for different surfaces in the  scene.
Since {\name} uses only signal power---a single scalar measurement that contains a mixture of the LoS and all the reflections---the inverse problem of recovering the voxel attributes from only the power is far more complex. 
}

\section{{\name} Model}

\textbf{Setup and Overview.}
At a $Rx$ location, we model the received signal power $\psi$ as
% The signal power $\psi$, received by a receiver $Rx$, can be \hl{modeled as}: 
\[ \psi  = \psi_{LoS} + \psi_{ref_1} + \cdots + \psi_{ref_n} \]
where $\psi_{LoS}$ is the power from the direct line-of-sight (LoS) path, and $\psi_{ref_k}$ is the aggregate power from all $k^{th}$ order reflections (i.e., all signal paths that underwent exactly $k$ reflections before arriving at the $Rx$) 
\footnote{This model is a simplification since it ignores the signal phase in estimating the received power. Appendix \ref{sec:modelling_signal_power} shows that with a moving wide-band receiver, like WiFi, the approximation may be tolerable for a sensing application like {\name}.}
% \footnote{This model is an approximation since it relies on averaging measurements as explained in Appendix \ref{sec:modelling_signal_power}. 
% However, for this sensing application involving moving receivers, and when using wideband-multipath systems such as WiFi, this approximation holds well.}.
% This model is an approximation; the accurate received power must be computed by adding signals based on their respective phases and amplitudes. 
% However, this approximation holds well for wideband signals like WiFi, the details of which are explained in Appendix \ref{}.
We assume $M$ fixed transmitters and move the $Rx$ to $N$ known locations and measure $\psi$ at each of them. 
% and initially assume their locations are also known (\hl{but we relax this assumption and show results in Appendix C} ).
{\name} accepts $M \times N$ measurements as input and outputs the 2D floor-plan $F$, a binary matrix of size $L \times L$, where $L$ denotes the maximum floorplan length. 

We train {\name} on the measured data using our proposed objective function. 
This function only models the LoS and the first order reflections.
We disregard the higher orders since they are very complex to model and contribute, on average, $<6\%$ of the total power (see statistics in Appendix \ref{sec:highorder}).
% We train {\name} on the measured data using our proposed {\em multipath power function} as part of the optimization objective. 
% This function models the LoS and the first order reflections.
% Higher order reflections are complex to model for real floorplans; moreover, they contribute, on average, $<6\%$ of the total power (see statistics in Appendix).
%(see Figure \ref{fig:glass_contributions}).
% Hence, we disregard higher order reflections.
The NeRF model we use is a remarkably simple MLP designed to predict the density $\delta \in [0,1]$ and orientation $\omega \in [-\pi, \pi]$ of a specified voxel in the indoor scene.
The orientation aids in modeling reflections.
The proposed objective function -- parameterized by voxel attributes $\langle \delta, \omega \rangle$ and the $\langle Tx, Rx \rangle$ locations -- models an approximation of the received power $\psi$ at that $Rx$ location.
Minimizing $L_2$ loss of this power across all $Rx$ locations  trains the MLP.
Plotting out all the voxel densities in 2D gives us the estimated floorplan $F$.

\subsection{The LoS Model}
Friss' equation \cite{antennaBook} from electromagnetics models the free-space received power as $P_r=\frac{K}{d^2}$ where $d$ is the distance of signal propagation, and $K$ is a product of transmit-power, wavelength, and antenna-related constants~\cite{antennaBook}.
% Friss' equation \cite{antennaBook} from electromagnetics models free-space received power as $P_r = P_t \frac{G_t G_r \lambda^2}{(4\pi d)^2}$ where \( P_t \) is the transmitted power, \( G_t \) and \( G_r \) are the gains of the $Tx$ and $Rx$ antennas, \( \lambda \) is the wavelength, and \( d \) is the distance of signal propagation \cite{Tse_Viswanath_2005}.
We model this free-space (LoS) behavior in the NeRF framework through the following equation.
\begin{equation} \label{eqn:rssi_los}
\psi_{LoS} = K \frac{\displaystyle \prod_{\{i| v_i \in {LoS}\}}(1-\delta_i)}{d^2}   
\end{equation}
where $K$ can be empirically measured, and $d$ is the known distance between the $\langle Tx, Rx \rangle.$
The numerator includes the product of voxel densities over all voxels along the LoS ray from $Tx$ to $Rx$ (with an abuse of notion, we write this as $v_i \in {LoS}$).
This models occlusions.
When the LoS path is completely free of any occlusions (i.e., $\delta_i = 0, \forall i$ where $\{i|v_i \in {LoS}\}$), we expect the received power to only be attenuated by the pathloss factor $d^2$ (in the denominator).
Eq. \ref{eqn:rssi_los} has a slight difference to classical NeRF's volumetric scene function.
In our case, voxels along the ray do not contribute to the received power (whereas in NeRF, each voxel's color is aggregated to model the final pixel color at the image).
In other words, we have modeled a single transmitter in Eq. \ref{eqn:rssi_los}.

\subsection{The Reflection Model}
To model reflections, consider a voxel $v_j$.
% For $v_j$ to be a valid reflector voxel from $Tx$ to $Rx$
Whether $v_j$ reflects a ray from the $Tx$ towards $Rx$ depends on (1) $v_j$'s density $\delta_j$ and orientation $\omega_j$, (2) the $\langle Tx, Rx \rangle$ locations, and (4) whether the path from $Tx$ to $v_j$, and from $v_j$ to $Rx$ are both occlusion-free.
Parameterized by these, Eq. \ref{eqn:psi_ref_r_pi} models $\psi_{ref}(v_j)$, which is the received power at $Rx$ due to the signal that reflected off voxel $v_j$. 

% \begin{equation} \label{eqn:psi_ref_r_pi}
% \psi_{ref_1}(m) = \delta_{p_{m}} \frac{f(\theta,\beta) \Pi_{j=Rx}^{m}(1-\delta_{p_j})   \Pi_{k=Tx}^{m}(1-\delta_{p_k}) }{d^2_{Tx:m+m:Rx}}
% \end{equation}

\begin{equation} \label{eqn:psi_ref_r_pi}
\psi_{ref}(v_j) = \delta_{j} f(\theta,\beta) \frac{\displaystyle \prod_{k \in \{Rx:v_j\}}(1-\delta_{k})   \prod_{l \in \{v_j:Tx\}}(1-\delta_{l}) }{\left(d_{Tx:v_j} + d_{v_j:Rx}\right)^2}
\end{equation}

Let us explain this equation briefly. 
The leading $\delta_j$ ensures that voxel $v_j$ is not a reflector when $\delta_j=0$.
The $f(\theta,\beta)$ term models the wave-surface interactions, i.e., how signals get attenuated as a function of the incident angle $\theta$ and how signals scatter as a function of the offset angle $\beta$ (which is the angle between the reflected ray and the direction of the $Rx$ from $v_j$). 
The next two product terms ensure that for the $Rx$ to receive this reflection, the voxels along the $2$ segments ($Tx$ to $v_j$ and $v_j$ to $Rx$) must be non-opaque; if any $\delta_k$ or $\delta_l$ equals $1$, that reflection path is blocked, producing no power contribution via this voxel $v_j$ to the receiver $Rx$.
Finally, the denominator is the squared distance from $Tx$ to $v_j$, and from $v_j$ to $Rx$, modeling signal attenuation.

To compute the full reflection power, the natural question is: {\em which voxels are contributing to the received power?}
Geometrically, any opaque voxel can be a plausible reflection point between any $\langle Tx, Rx \rangle$ pair.
This is because, for triangle formed by $Tx$, $v_j$, and $Rx$, the voxel orientation $\omega_j$ can be assigned a direction that bisects the angle at $v_j$.
For this $\omega_j$, the reflected ray will perfectly arrive at the $Rx$.
Thus, without the knowledge of orientation and density, the total first order reflection power at the $Rx$ should be modeled as the sum of reflections on all voxels.
This makes the optimization problem excessively under-determined.

\begin{figure}[!t]
  %\centering{0.45\columnwidth} % Use \columnwidth to make it fit within the single column
    \centering
    \begin{minipage}{0.42\textwidth}
    \vspace{1em}
\includegraphics[width=\columnwidth]{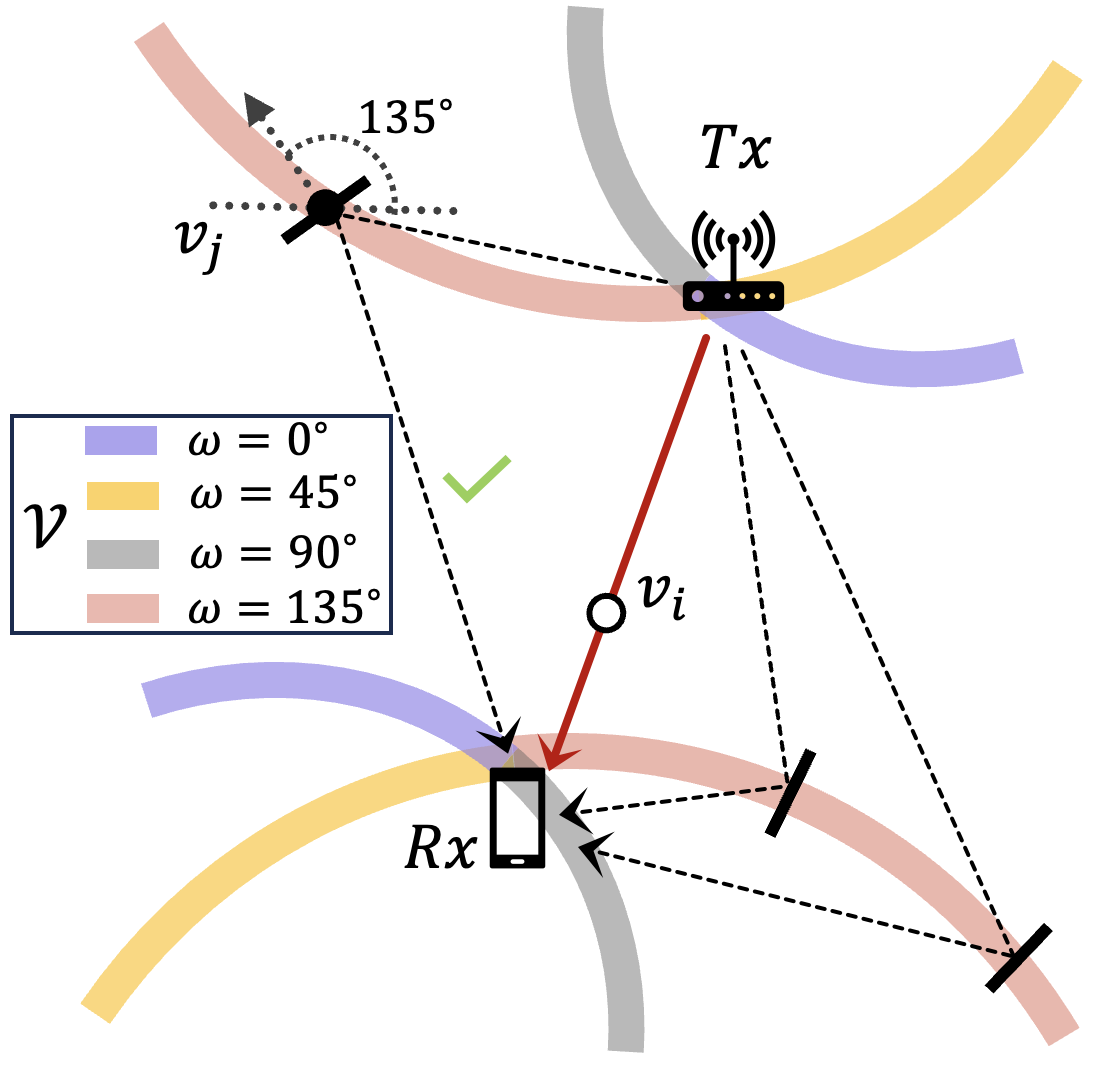}
\caption{Colored stripes define the manifold from which reflections are plausible between $\langle Tx, Rx \rangle$. Voxels located on the manifold form the plausible set $\mathcal{V}$. Dashed lines show plausible reflections.}
    % \caption{Plausible set $\mathcal{V}$ containing voxels that can potentially reflect the signal between \Tx\ and \Rx. Dashed lines show some plausible reflections.}
    \vspace{-0.1in}
    \label{fig:ray}
\end{minipage}
\hfill
\begin{minipage}{0.56\textwidth}
\includegraphics[width=\columnwidth]{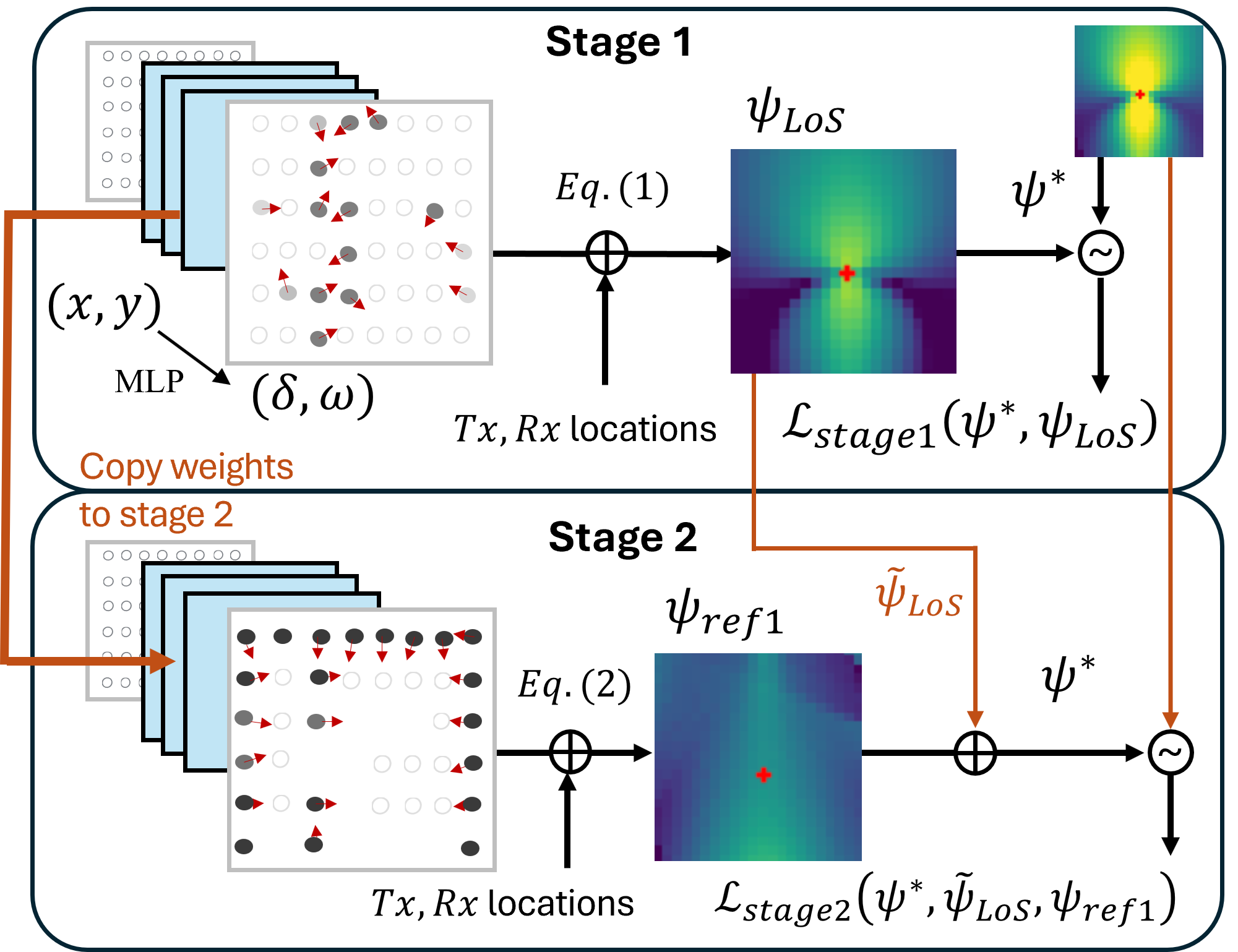}
  \caption{{\name}'s two-stage training approach: In Stage 1, the LoS model is trained using known $Rx$ locations and signal power. This provides a warm-start to the reflection model in Stage 2 which refines the learned voxel densities and orientation. }
  %During inference, {\name} outputs voxel densities at set of queried $Rx$ locations, representing the inferred floorplan.}
  \label{fig:flow}
  \vspace{-0.1in}
  \end{minipage}
\end{figure}

% We address this by modeling $\omega_j$ as a discrete value, i.e., $\omega_j \in \{ 1, 2 \dots K_\omega \}$.
% : $\omega_j \in \{ 1, 2 \dots K_\omega \}.$
\red{We address this by modeling $\omega_j$ as a discrete value---multiples of $\frac{\pi}{K_\omega}$}.
Larger $K_\omega$ is needed when the environment has complex surface orientations; however, most floorplans exhibit perpendicular walls \cite{manhattan1, manhattan2, manhattan3} and $K_\omega=4$ is adequate.
% To cope with this, we assume that surfaces in the environment are orientated discretely in one of $K_\omega$ angles.
% When $K_\omega = 4$, the walls can either be vertical, horizontal, tilted at $45^\circ$ or tilted at $135^\circ$.
Once $\omega_j$ becomes discrete, the voxels that can produce plausible reflections become far fewer -- we call this the ``plausible set'' $\mathcal{V}$.
Fig. ~\ref{fig:ray} visualizes $\mathcal{V}$ and shows $3$ out of many plausible reflections from voxels with 
% $\omega_j=45^\circ$.
\red{$\omega_j=135^\circ$}.
Eq. \ref{eqn: psi_ref} sums up the power from all reflections that occur on the plausible set:
\begin{equation} \label{eqn: psi_ref}
\psi_{ref_1} = \displaystyle \sum_{\{j|v_j \in \mathcal{V}\}} \psi_{ref}({v_j})
\end{equation}
Thus, the final modeled power at a specific $Rx$ location becomes $\tilde{\psi} = \psi_{LoS} + \psi_{ref_1}$.
%\begin{equation} \label{eqn: psi}

%\end{equation}

\subsection{Gradient Issues during Training}
\label{sec:gradientissues}

Training against a $L_2$ loss, $\mathcal{L} = \| \tilde{\psi} - \psi^* \|^2_2$ \footnote{Here, $\psi^*$ denotes the ground truth signal power}, did not generate legible floorplans.
We found that the $\psi_{LoS}$ dominated the loss term, drowning the reflection model's influence on learning. 
At a high level, the gradient of the LoS model (Eq. \ref{eqn:rssi_los}) w.r.t. $\delta_i$ has fewer terms in the numerator's product, and a smaller $d^2$ in the denominator.
The reflection model's gradient w.r.t. $\delta_i$ has many more terms since the reflection path is much longer; the denominator is also larger.
Since $(1- \delta_j) \leq 1$, their products force the gradient to decrease geometrically with more terms, causing the reflection gradient to be much smaller compared to LoS. 
% the voxel $v_i$'s gradient $\nabla_{\delta_i} \mathcal{L}$ contains fewer terms, since the product $\prod(1-\delta_i)$ in Equation \ref{} has fewer voxel \footnote{Note that $(1- \delta_j) \in [0,1]$ because $\delta_j \in [0,1]$; hence, the product of those terms grow smaller}, and a smaller $d^2$ in the denominator. This results in a relatively larger gradient magnitude. On the other hand, a reflecting voxel has a larger denominator and more terms in the product $\prod (1 - \delta_j), j \neq k$. This makes it's gradient quickly tends to zero, hindering proper training.
We formalize this explanation below by considering the LoS and reflection losses individually\footnote{For the ease of explanation, we use $\psi^*_{LoS}$ and $\psi^*_{ref}$ to denote the ground truth LoS and reflection powers, respectively. We do not need to know these terms in practice.}.
\begin{align*}
    \mathcal{L}_{LoS} &= \left\Vert \tilde{\psi}_{LoS}- \psi^*_{LoS} \right\Vert^2_2, \quad 
    \mathcal{L}_{ref} = \left\Vert \tilde{\psi}_{ref}- \psi^*_{ref} \right\Vert^2_2
\end{align*}
Consider the gradient of $\mathcal{L}_{LoS}$ w.r.t the density of $v_i$.
\begin{align}
\nabla_{\delta_i} \mathcal{L}_{LoS}&=2 (\tilde{\psi}_{LoS}- \psi^*_{LoS}) \sum_{\{n|i\in LoS^{(n)}\}}\nabla_{\delta_i}\psi^{(n)}_{LoS} \nonumber\\
\text{where} ~~ \nabla_{\delta_i}\psi^{(n)}_{LoS}&=-\frac{K}{{d^{(n)}}^2}\prod_{\substack{j\in LoS^{(n)}\\j\neq i}}(1-\delta_j)\label{eq:grad_LoS}
\end{align}
where $LoS^{(n)}$ is the $n$-th LoS path passing through voxel $v_i$.

The gradient of $\mathcal{L}_{ref}$ has a nearly identical expression with the only difference being many more product terms and that  $\nabla_{\delta_i} \psi_{ref}^{(n)}$ depends on where $v_i$ is present in the $n$-th set of reflection path voxels (denoted by $Ref^{(n)}$). 
For example,
\begin{align}
\nabla_{\delta_i} \psi_{ref}^{(n),Tx}&=-C\prod_{\substack{j\in Ref^{(n)}_{Tx:v}\\ j\neq i}}(1-\delta_j)\prod_{k\in Ref^{(n)}_{v:Rx}}(1-\delta_k) \label{eq:grad_refl} \\
C &=\frac{\delta^{(n)}f^{(n)}(\theta,\beta)}{(d^{(n)}_{Tx:v}+d^{(n)}_{v:Rx})^2}\nonumber
\end{align}
here $\nabla_{\delta_i} \psi_{ref}^{(n),Tx}$ denotes the gradient when $v_i$ is between \Tx\ to the reflection point $v$.

Finally, since the modeled power approximates the measured power, the residual error will remain non-zero even if the floorplan is accurately learnt. 
As a result, \hl{the optimization is biased} towards voxels of higher gradients, i.e., voxels on the LoS path, suppressing the importance of reflections. To address this, we train {\name} in $2$ stages.
% as discussed next. 
\vspace{-0.05in}
\vspace{0.05in}

\subsection{Multi-stage Training}
\noindent\textbf{Stage 1:} We first use the LoS model against the measured ground truth power $\psi^*$ (see Fig. \ref{fig:flow}).
This converges quickly because the network easily learns the transparent voxels ($\delta$=$0$) located along LoS paths. 
For LoS paths that are occluded, the network {\em incorrectly} learns excessive opaque voxels between the $\langle Tx, Rx \rangle$, 
% (i.e., very thick walls), 
but this does not affect the LoS error since the path is anyway occluded.
Hence, the outcome is a crude floorplan but a near-perfect LoS power estimate $\tilde{\psi}_{LoS}$.
We utilize this $\tilde{\psi}_{LoS}$ in stage $2$ (discussed soon).

As Stage $1$ training progresses, some opaque voxels emerge, offering crude contours of some walls.
We estimate a voxel's spatial gradient, $\nabla \delta_i$, and use it to supervise the orientation $\omega_i$ of that voxel.
The intuition is that a voxel's orientation -- needed to model reflections in Stage $2$ -- is essentially determined by the local surface around that voxel.
The gradient $\nabla \delta_i$ offers an opportunity for weak supervision.
Thus, the loss for Stage $1$ is:
\begin{align}
\mathcal{L}_{stage1} &= \left\Vert \psi^* - \psi_{LoS}\right\Vert^2_2 + \mathcal{L}^+ \\
\text{where} ~~ \mathcal{L}^+ &= \lambda_1 \underset{\forall j}{\sum} \left\Vert \nabla \delta_j - \omega_j \right\Vert_2^2 + \lambda_2 \mathcal{L}_{reg}
\end{align}

with $\lambda_1, \lambda_2 > 0$ being tunable hyperparameters. The regularization term $\mathcal{L}_{reg}$ will be discussed soon.
Finally, the near-perfect estimate of LoS power, denoted $\tilde{\psi}_{LoS}$, is also carried over to Stage $2$ to ensure the reflection model is penalized when it veers away from this LoS estimate. 
\vspace{0.05in}

% \textbf{Stage 1:} To address this, we train {\name} in $2$ stages. 
% In stage $1$, we only use the LoS model against the measured groundtruth power as below.
% \[
% \mathcal{L}_{stage1} = || \psi^* - \psi_{LoS}||^2_2
% \]
% This results in a crude floorplan, but importantly, the LoS power is almost perfectly estimated since the network easily learns the transparent ($\delta = 0$).
% For LoS paths that are truly occluded, the network incorrectly learns excessive opaque voxels between the $\langle Tx, Rx \rangle$, but this error does not affect the LoS power since the path is anyway occluded.
% \hl{Say psi tilde}.

\noindent\textbf{Stage 2} focuses on training the reflection model using the following loss function.
\[
\mathcal{L}_{stage2} = \left\Vert \tilde{\psi}_{LoS} - \psi_{LoS} \right\Vert^2_2 + \left\Vert \psi^* - \tilde{\psi}_{LoS} - \psi_{ref_{1}} \right\Vert^2_2  + \mathcal{L^+}
\]

The first term in the RHS ensures that the Stage $1$'s LoS estimate is honored in Stage $2$.
The second term subtracts Stage $1$'s LoS power from the measured power,  
($\psi^* - \tilde{\psi}_{LoS}$); this models the total power {\em only due to reflections}. 
Our (first order) reflection model $\psi_{ref_1}$ is trained to match this aggregate power ($L_2$ loss).
% The first term ensures that by learning to improve the floorplan based on reflections, the estimates from the first stage is not affected.
The supervision on orientation and the regularization terms are the same as in Stage $1$.

\noindent $\blacksquare$ \textbf{Regularization}: 
Floorplans demonstrate significant local similarity in orientation, hence we penalize differences in orientation among neighbors,  using a regularization (Eq. \ref{eqn:Lreg}) similar to Total Variation \cite{TV_reg}.
% enforce that nearby voxels should exhibit similar orientations. 
This can be achieved without additional computational cost to the neural network by directly utilizing voxel orientations obtained from each ray.  
% Our regularization term is shown in 
\begin{align} \label{eqn:Lreg}
\mathcal{L}_{\text{reg}} = \frac{1}{n_v(n_r-1)} \sum_{n=1}^{n_v} \sum_{i=1}^{n_r-1} \left\Vert {\omega}_{n,i+1} - {\omega}_{n,i}\right\Vert^2_2
\end{align} 
Here $n_v$ is the number of voxels queried from the plausible set $\mathcal{V}$ and $n_r$ is the number of voxels along the each ray.

\section{Experiments}

\textbf{Floorplan and Wireless Simulation Dataset.}
Floorplans are drawn from the Zillow Indoor Dataset ~\cite{cruz2021zillow}.
% ; we also generate floorplans from real apartments. 
% composed of \hl{one-story homes}.
% We also generate floorplans from realistic apartment layouts.
In each floorplan, we use the $A^*$ algorithm ~\cite{astar} to generate a walking trajectory that traverses all rooms.
% This mimics a user walking with a phone and collecting WiFi measurements.
% The rooms are devoid of furniture, however, we will later add a few toy shapes to mimic objects in the path of trajectories.
% Due to the difficulty in collecting large amounts of data for training, 
% We train and test the {\name} pipeline on simulation data. 
% The dataset is composed to two parts: (1) generated floor plans by ourselves according to real floor plan distribution; (2) floor plans from Zillow Indoor Dataset~\cite{cruz2021zillow}, which composes of thousands of 3D scans of real homes. We ensure the simulation mimics realistic indoor WiFi propagation by using state-of-the-art simulation engine.
% \noindent\textbf{Wireless Simulation Dataset.}
We use the NVIDIA Sionna RT~\cite{hoydis2022sionna, hoydis2023sionna} -- a ray tracer for radio propagation modeling -- to compute the ground truth signal power (also known as {\em received signal strength index} (RSSI)). 
We randomly place $M$ \Tx s, one in each room, denoted as $T_m$. 
To simulate omnidirectional transmissions at $2.4$GHz from each \Tx\ location, we shoot $10^7$ rays into the given floorplan. 
For receiver locations, we sample the user trajectory at a fixed time interval to obtain $N$ \Rx\ locations, denoted as $R_n$. 
% Only rays that reach the \Rx\ location within $5$ bounces (i.e., up to $5$-th order reflections) contribute to the RSSI; others attenuate excessively and are negligible.
Sionna accounts for specular reflections and refraction when these rays interact with walls in the specified floor plan; we use the default materials for the walls.
As with most WiFi simulators, Sionna does not model signal penetration through walls -- this means that a \Tx\ and \Rx\ located on opposite sides of a wall will not receive any RSSI.
% Using Sionna, we obtain signal power estimates $P_{m,n}$ by inputting $(FP, T_m, R_n)$, where $FP$ represents the floorplan in bitmap format, and $T_m$ and $R_n$ are transmitter and receiver locations.
% For each transmitter and receiver pair, we can get one RSSI simulated measurement. 
Overall, we gather $M\times N$ RSSI measurements $(T_m, R_n, \psi_{m,n})$ that serve as input to {\name}. 
% {\name} utilizes these measurements as input to estimate $FP$.
% To evaluate {\name}'s performance in signal power prediction, we split the measurements into an 80-20 ratio for training and testing.

\noindent\textbf{Baselines} used for comparison are: 
% Results are compared against the following baselines:
\begin{enumerate}
\item \textbf{\NeRFs} \cite{zhao2023nerf2}: Models WiFi reflections via virtual transmitters to predict channel impulse response (CIR).
\item \textbf{\ZYBase} \cite{liu2012review}: Interpolates CIR across the whole floorplan and applies an image segmentation algorithm (on the interpolated RSSI heatmap) to isolate each room.
Essentially, the algorithm identifies the contours of sharp RSSI change since such contours are likely to correspond to walls.
% Estimates the floorplan based on one assumption: \Rx s on different sides of a wall will observe a large signal power gap. 
% By grouping \Rx s with similar signal power values, \ZYBase\ infers walls as boundaries between groups.
Implementation details are included in the Appendix \ref{sec:segmentation}.
\item \textbf{\MLP}: Trains an MLP network to directly estimate the RSSI based on \Tx\ and \Rx\ locations. 
\item {\nameLoS}: Reports {\name}'s result considering only LoS path (ablation study). 
\end{enumerate}
% \vspace{0.05in}

% We compare {\name} with a recent work, \NeRFs\ ~\cite{zhao2023nerf2}, and 3 baseline / ablation methods, \ZYBase, \MLP, and \nameLoS. Each baseline model is trained and tested with the same dataset and train-test split.

\noindent\textbf{Metrics.} We evaluate using $3$ metrics:

\textbf{(A) Wall Intersection over Union (Wall\_IoU)}: This metric measures the degree to which the predicted walls and the true walls superimpose over each other in the 2D floorplan.
The following equation defines the metric:
\begin{equation*}
        \text{\texttt{Wall\_IoU}}=\frac{WP\cap WP^*}{WP \cup WP^*}
\end{equation*}
where $WP$ denotes the set of predicted wall pixels and $WP^*$ denotes the true wall pixels.
This is a {\em harsh metric} given wall pixels are a small fraction of the total floorplan; if a predicted wall is even offset by one pixel from the true wall, the \texttt{Wall\_IoU} drops significantly.
{IoU} \cite{iou} has often been defined in terms of room pixels (instead of wall pixels); this is an overestimate in our opinion, since predicting even an empty floorplan results in an impressively high IoU.

\textbf{(B) F1 score} \cite{metrics}: 
Defined as $F1=\frac{2\times P\times R}{P+R}$, where $P$ is the {\em precision} and $R$ is the {\em recall} of the bitmap. 
$P$ and $R$ are defined based on wall pixels, similar as above.

\textbf{(C) RSSI Prediction Error (RPE)}: 
We split all $Rx$ locations into a training and test set. 
RPE reports the average median RSSI error over all the test locations across floorplans.

\subsection{Overall Summarized Results}

\begin{table*}[!hbt]
    \centering
    \begin{tabular}{| l | c | c | c | c | c | c |}
        \hline
         & \multicolumn{3}{c|}{2000 receiver locations} & \multicolumn{3}{c|}{1000 receiver locations} \\ \hline
        Method & Wall\_IoU $\uparrow$ & F1 Score $\uparrow$ & RPE $\downarrow$ & Wall\_IoU $\uparrow$ & F1 Score $\uparrow$ & RPE $\downarrow$ \\ \hline
        \MLP & - & - & 1.03 & - & - & 0.65 \\ 
        \texttt{Heatmap Seg.}& 0.12 $\pm$ 0.03 & 0.21 $\pm$ 0.05 & 1.32 & 0.09 $\pm$ 0.02 & 0.16 $\pm$ 0.04 & 1.46 \\ 
        \NeRFs & 0.14 $\pm$ 0.02 & 0.24 $\pm$ 0.03 & 4.36 & 0.12 $\pm$ 0.02 & 0.21 $\pm$ 0.04 & 4.2 \\ 
        \nameLoS & 0.27 $\pm$ 0.07 & 0.42 $\pm$ 0.10 & 9.12 & 0.25 $\pm$ 0.04 & 0.39 $\pm$ 0.06 & 10.86 \\ 
        \name & 0.38 $\pm$ 0.06 & 0.55 $\pm$ 0.06 & 3.56 & 0.32 $\pm$ 0.06 & 0.48 $\pm$ 0.05 & 4.32 \\ 
        \hline
    \end{tabular}
    \caption{Performance Results for Wall\_IoU, F1 Score, and RPE}
    \vspace{-0.2in}
    \label{tab:performance_results}
\end{table*}

Table~\ref{tab:performance_results} reports comparative results between {\name} and baselines, averaged over $20$ different experiments, using all $3$ metrics.
The number of measurements are sparse ($N=2000$ and $N=1000$), given that apartment sizes in our dataset are more than $250,000$ pixels.
% With $N$=$2000$ $Rx$ locations, the user is expected to walk for around $8$ minutes, assuming her phone is sampling at $4$Hz.
% We repeat the experiment for fewer measurements of $N$=$1000$, reducing the user's burden to $4$ minutes of walking; this is sparse measurements given most floorplans are more than $250,000$ pixels.
Mean and standard deviation are reported in the table.
% floor plans and compares {\name} with baselines. 
% For each floor plan, we show results on 3 different scenarios by varying the number of receivers locations - we sample $500$, $1000$, and $2000$ receiver locations on user trajectories.
{\name} outperforms all models in terms of \texttt{Wall\_IoU} and \texttt{F1 Score}. 
Compared to \nameLoS, \name\ demonstrates visible improvements, highlighting the advantage of modeling reflections.
The absolute \texttt{Wall\_IoU} values are understandably low because {\em the metric penalizes small errors.}

\NeRFs\ is unable to predict the floor plan (opaque voxels) well and is only able to achieve better RPE than \nameLoS. 
\name\ outperforms both \nameLoS\ and \NeRFs.
% However, they predict the RSSI more accurately (lower RPE) than {\name} since they are explicitly trained on it.
Interestingly, MLP incurs a lower RPE than \NeRFs\, suggesting that RSSI is amenable to interpolation, and \NeRFs's implicit representation may not be an advantage for this interpolation task.

\begin{figure*}[t]
  % \vspace{-1em}
    \begin{tabular}{ @{\hskip 5pt} l @{\hskip 5pt} c@{ } c @{ } c @{ } c @{ } c @{ } c }
    % \hline
    \parbox[c][0.6cm][c]{0.1\textwidth}{\centering \vspace{-2cm}Ground Truth
    } & 
    \includegraphics[width=0.14\textwidth]{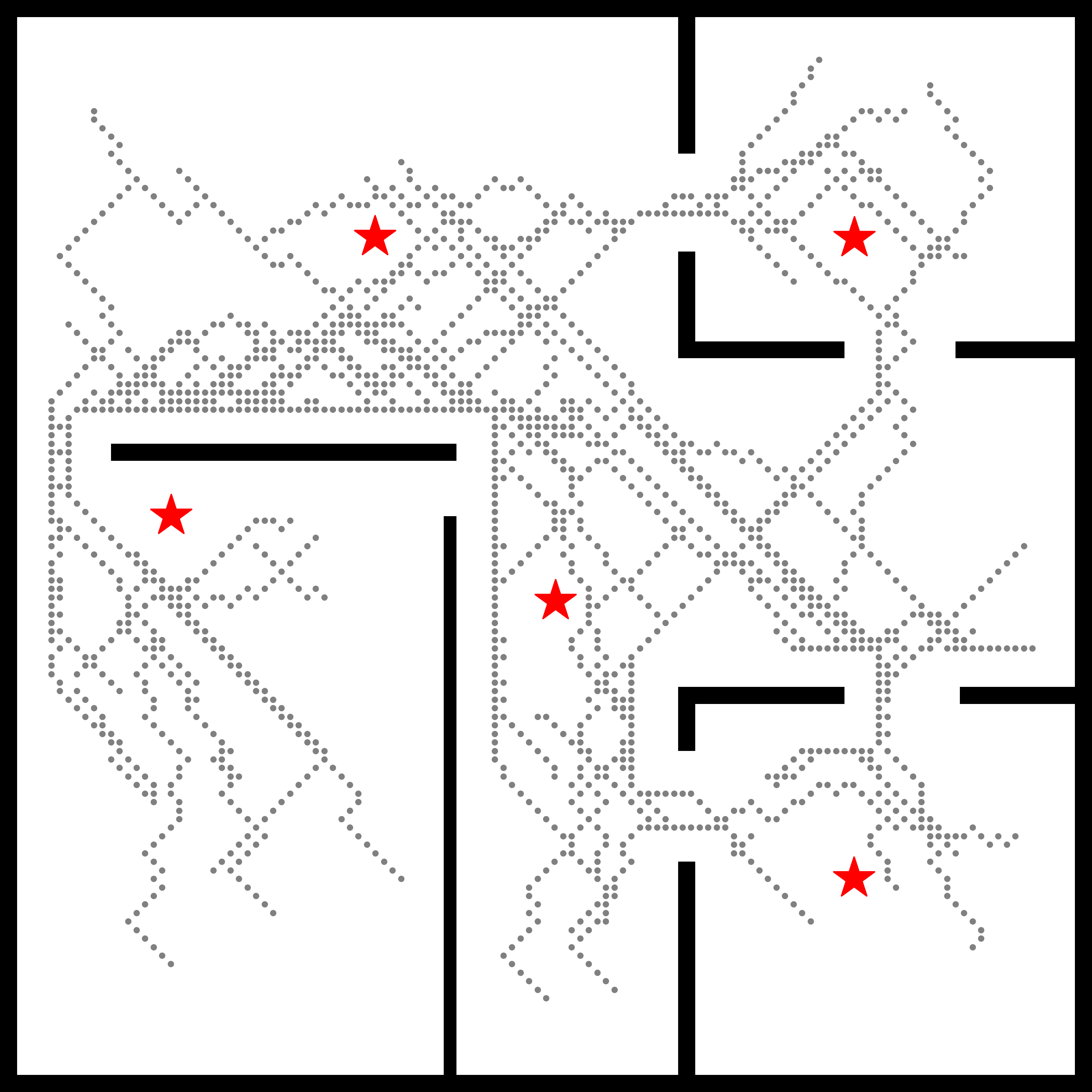}&
    \includegraphics[width=0.14\textwidth]{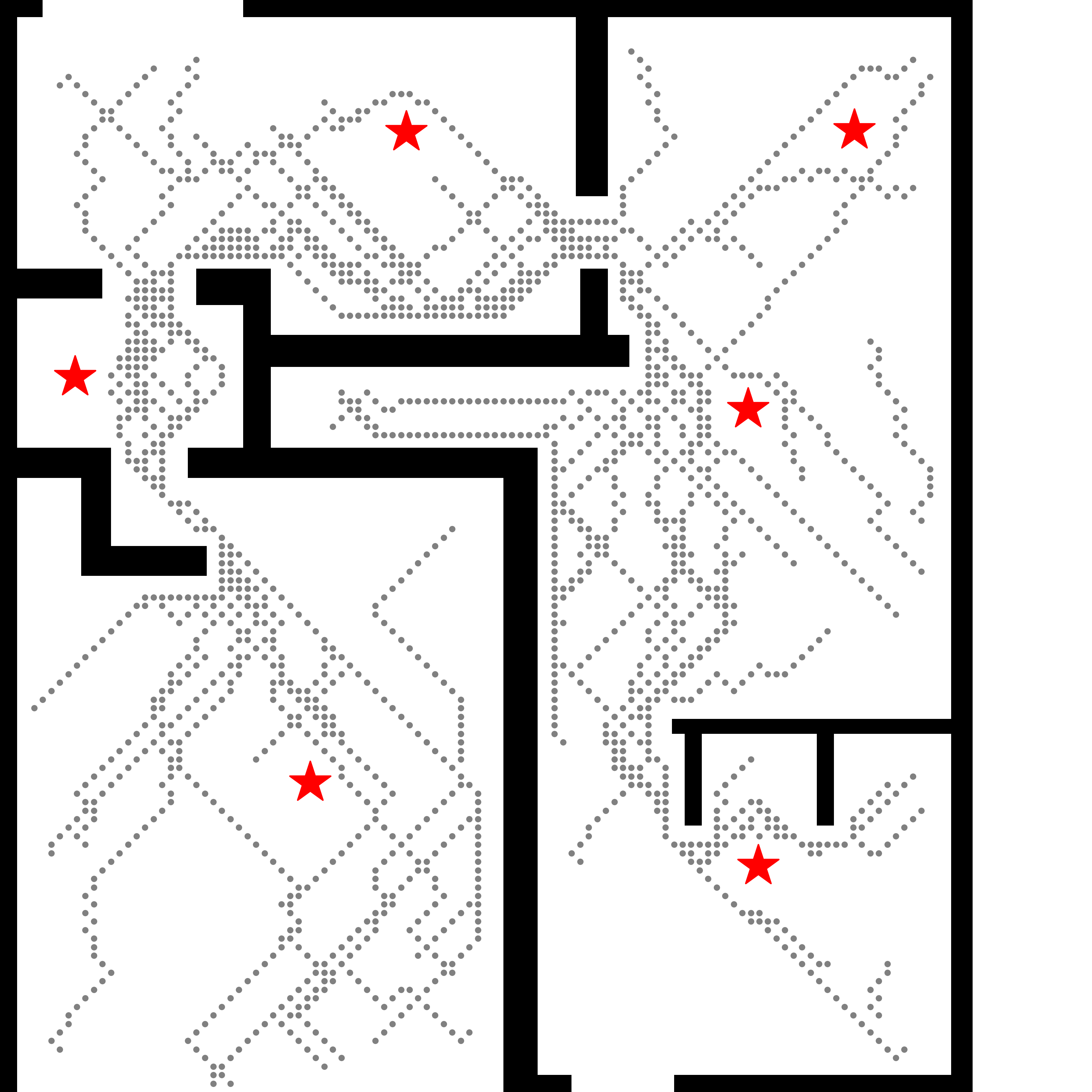}&
    \includegraphics[width=0.14\textwidth]{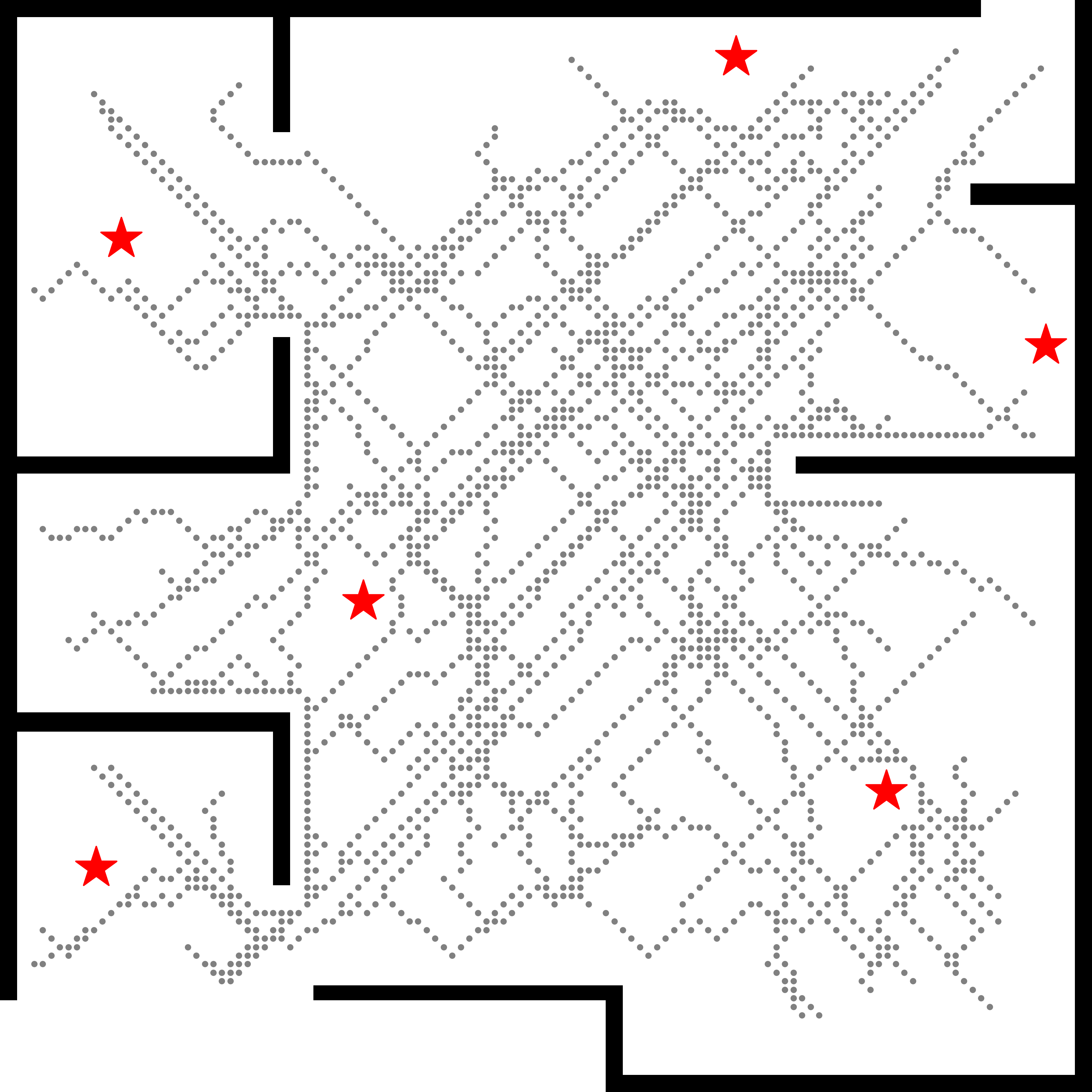}&
    \includegraphics[width=0.14\textwidth]{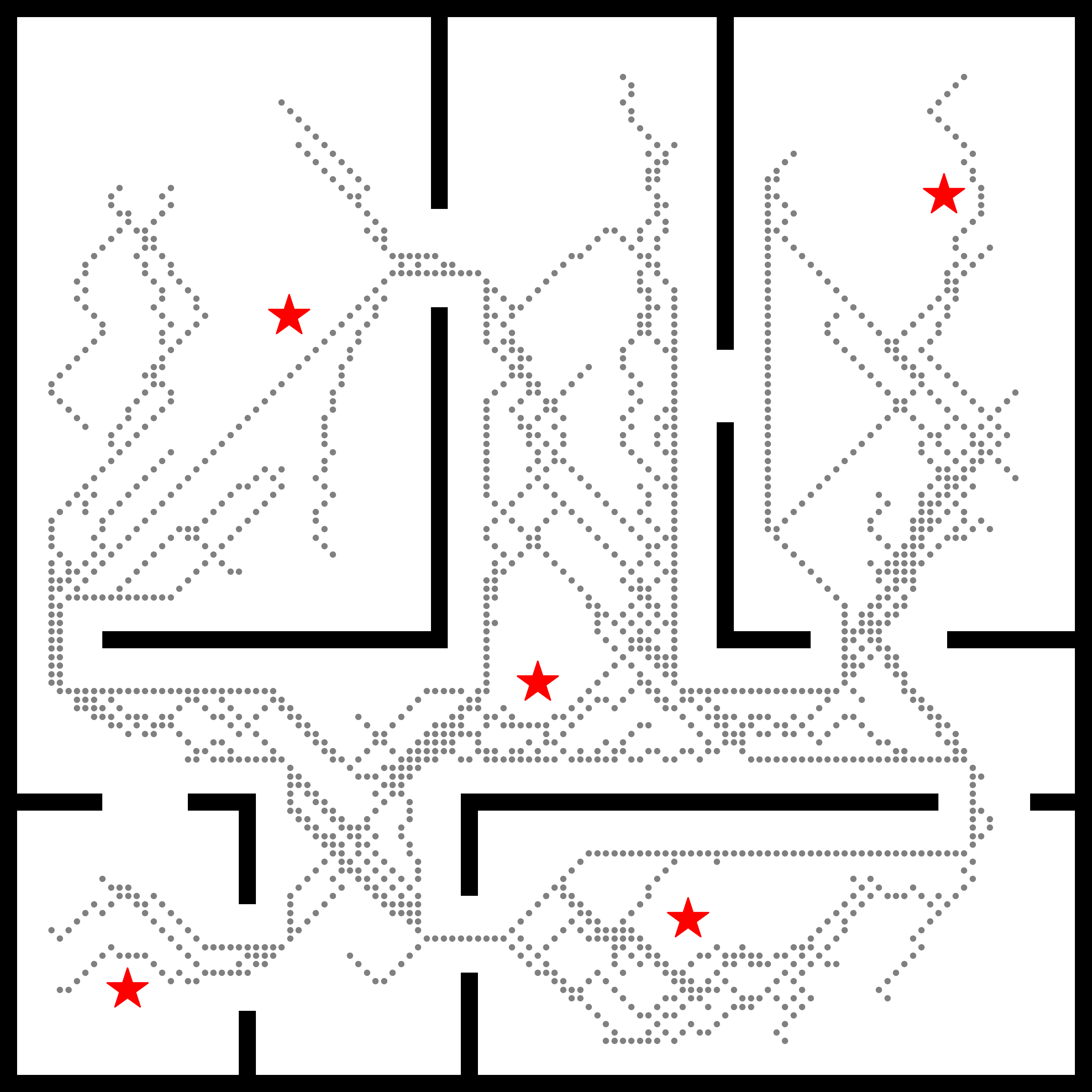}& 
    \includegraphics[width=0.14\textwidth]{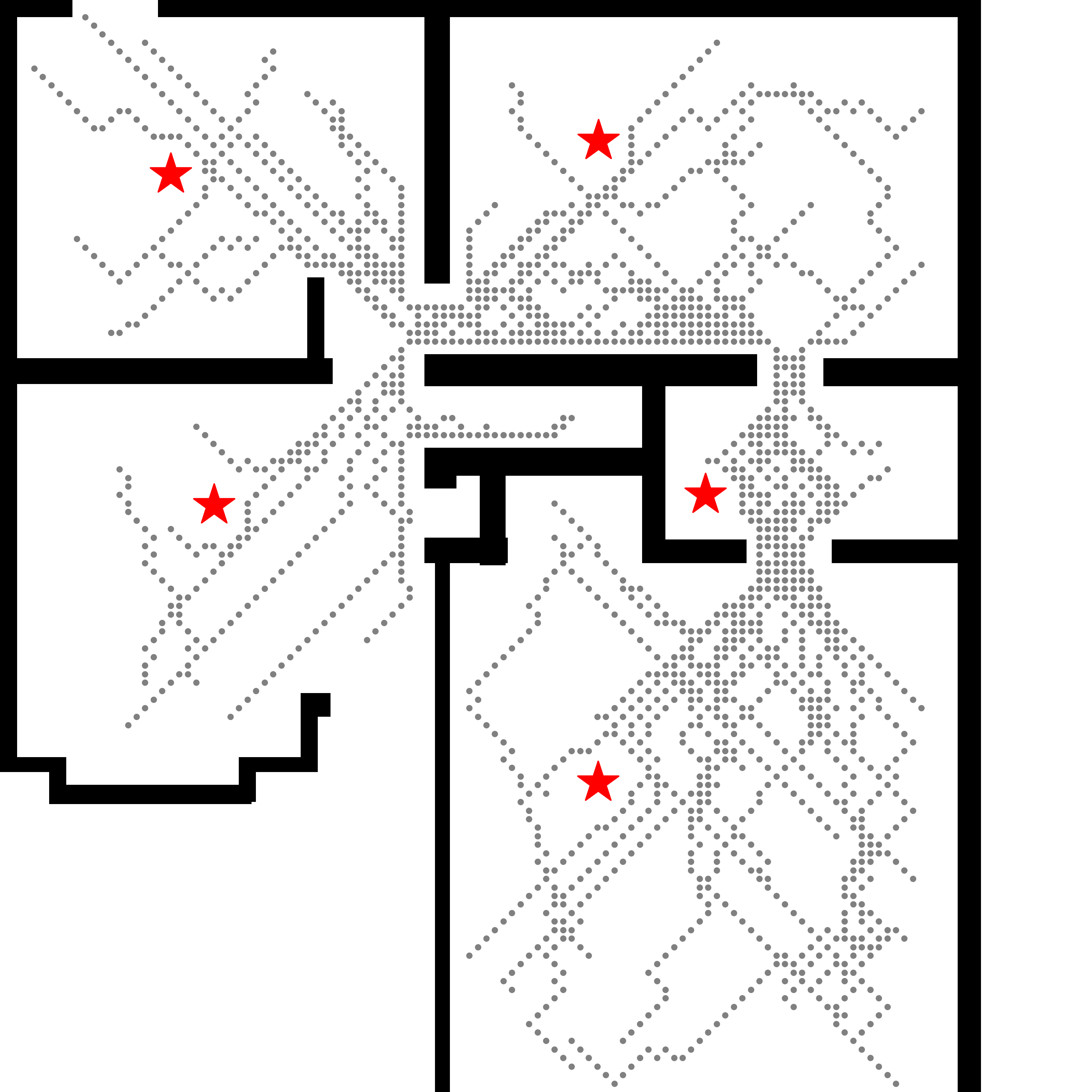}&
    \includegraphics[width=0.14\textwidth]{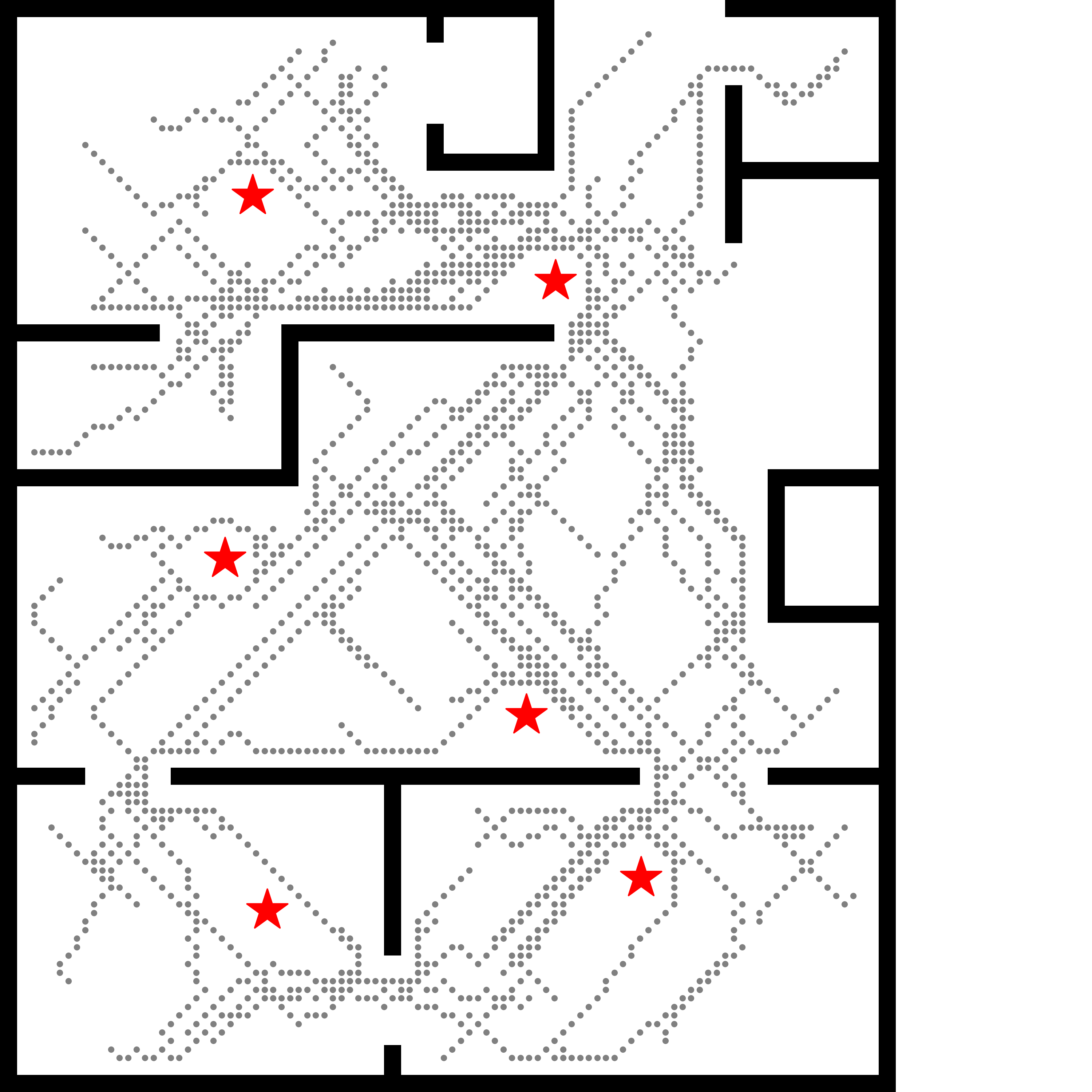}
    \\
    %\hline
    \parbox[c][.6cm][c]{0.1\textwidth}{\centering \vspace{-2cm} \texttt{Heatmap Seg.}} & 
    \includegraphics[width=0.14\textwidth]{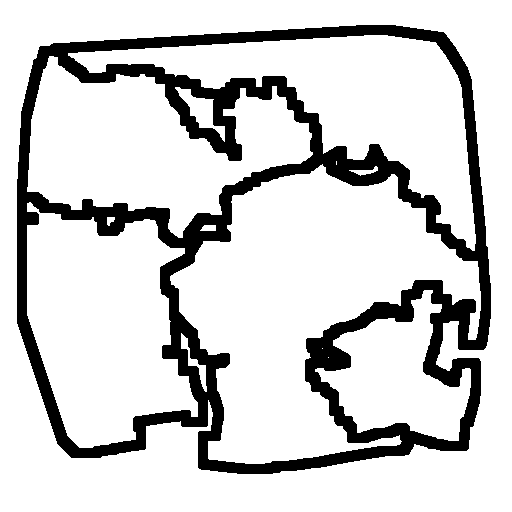}&
    \includegraphics[width=0.14\textwidth]{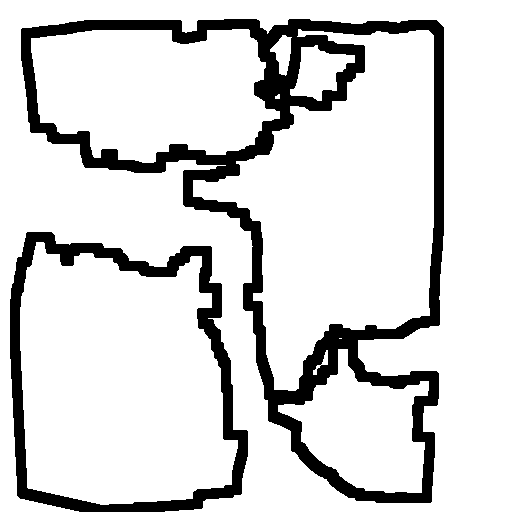}&
    \includegraphics[width=0.14\textwidth]{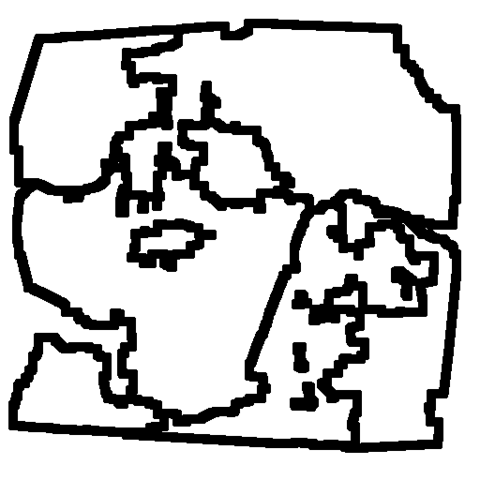}&
    \includegraphics[width=0.14\textwidth]{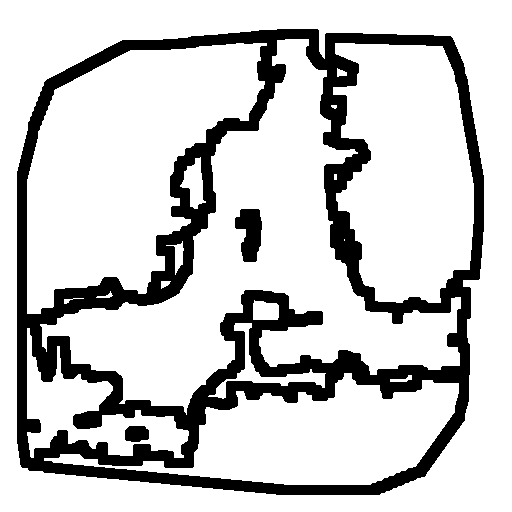}&
    \includegraphics[width=0.14\textwidth]{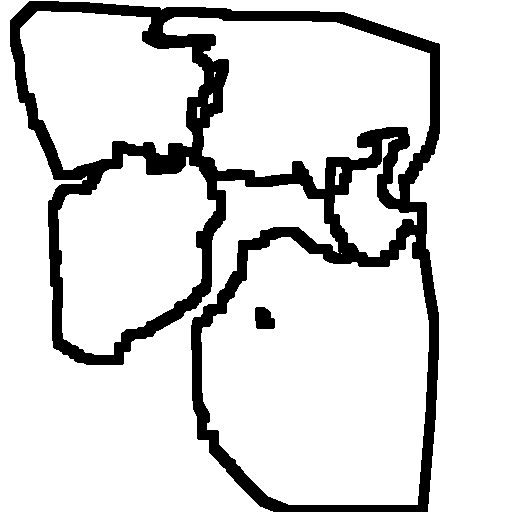}&
    \includegraphics[width=0.14\textwidth]{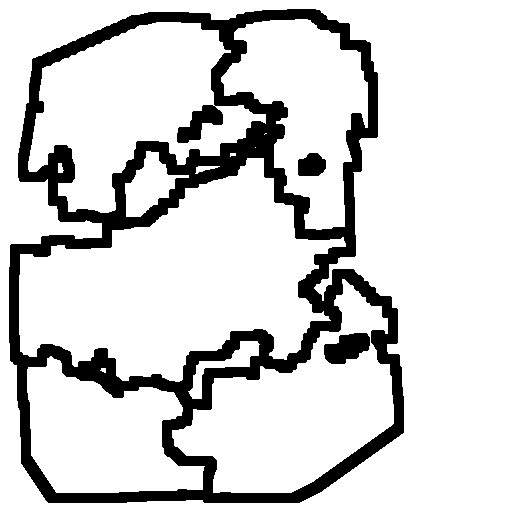}
    \\
    \parbox[c][.6cm][c]{0.1\textwidth}{\centering \vspace{-2cm} \NeRFs} & 
    \includegraphics[width=0.14\textwidth]{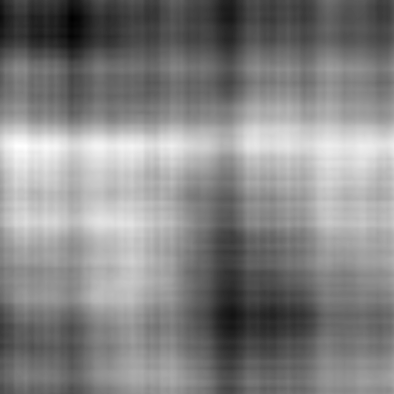}&
    \includegraphics[width=0.14\textwidth]{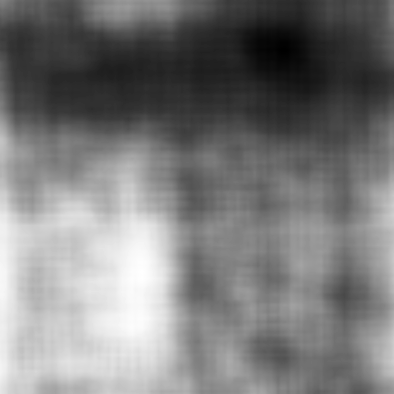}&
    \includegraphics[width=0.14\textwidth]{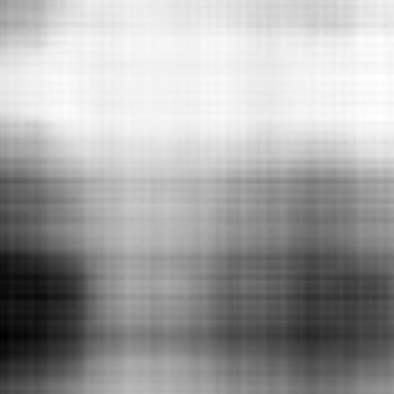}&
    \includegraphics[width=0.14\textwidth]{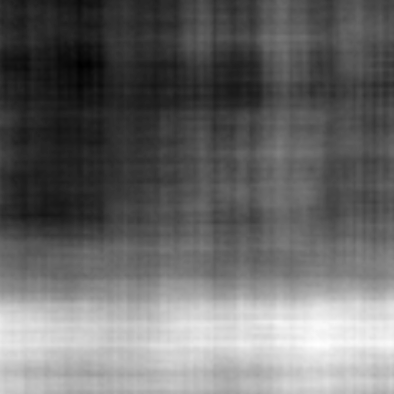}&
    \includegraphics[width=0.14\textwidth]{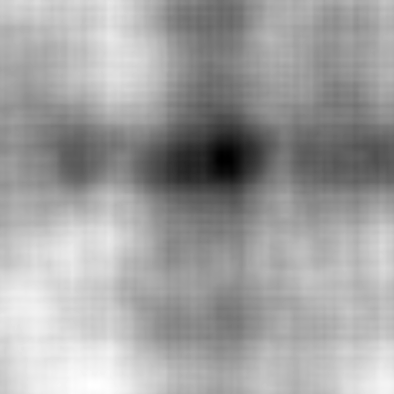}&
    \includegraphics[width=0.14\textwidth]{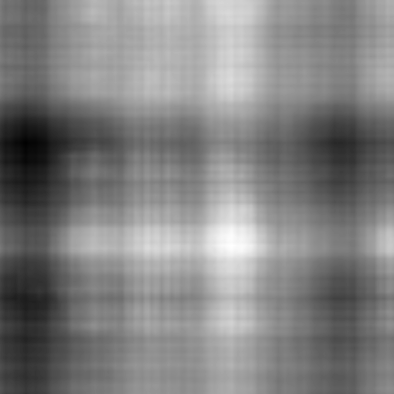}
    \\
    \parbox[c][1.2cm][c]{0.1\textwidth}{\centering \vspace{-2cm} \texttt{EchoNeRF LoS}} & 
    \includegraphics[width=0.14\textwidth]{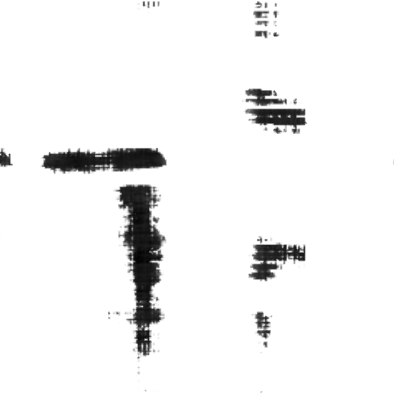}&
    \includegraphics[width=0.14\textwidth]{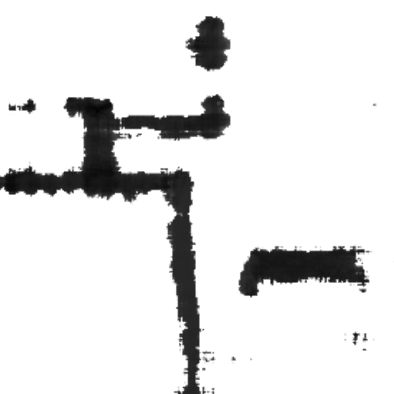}&
    \includegraphics[width=0.14\textwidth]{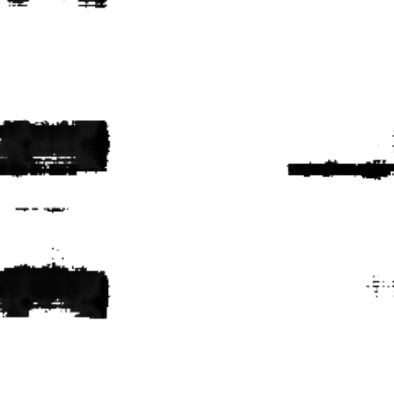}&
    \includegraphics[width=0.14\textwidth]{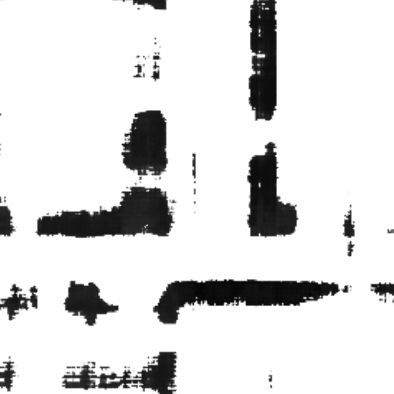}&
    \includegraphics[width=0.14\textwidth]{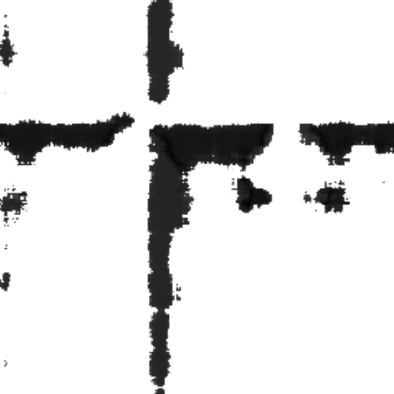}&
    \includegraphics[width=0.14\textwidth]{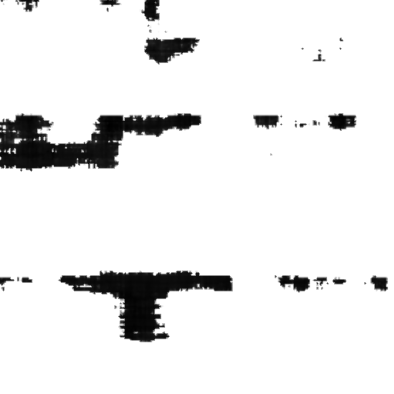}
    \\
    \parbox[c][.6cm][c]{0.1\textwidth}{\centering \vspace{-2cm} \name} &
    \includegraphics[width=0.14\textwidth]{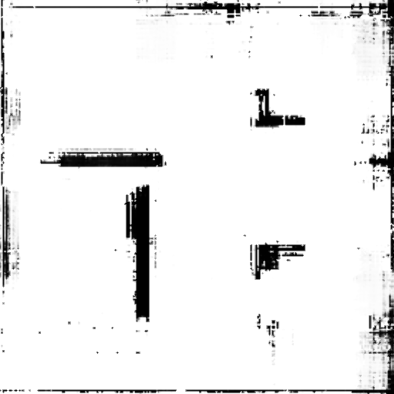}&
    \includegraphics[width=0.14\textwidth]{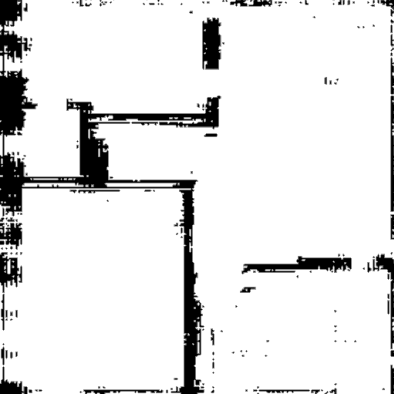}&
    \includegraphics[width=0.14\textwidth]{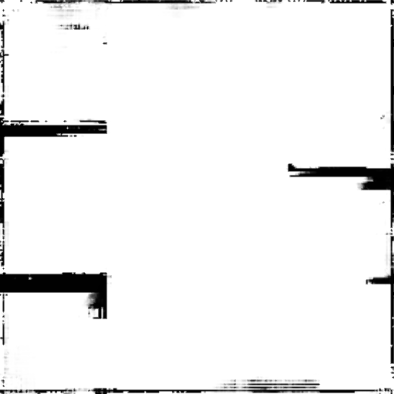}&
    \includegraphics[width=0.14\textwidth]{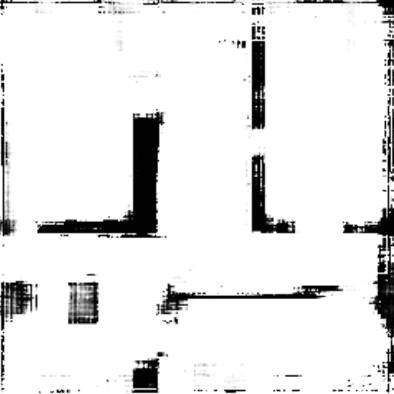}&
    \includegraphics[width=0.14\textwidth]{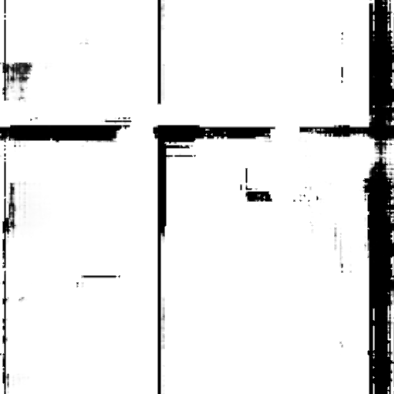}&
    \includegraphics[width=0.14\textwidth]{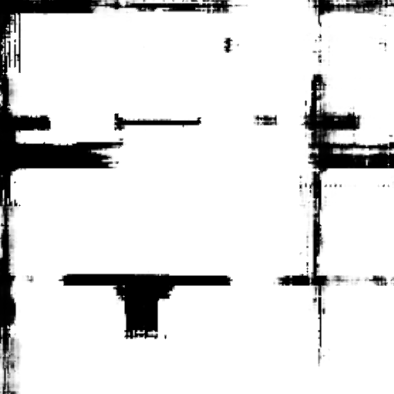}
  \end{tabular}
  % \vspace{-0.1in}
  \caption{Qualitative comparison of ground truth floorplans against baselines.
  \red{In the first row, red stars denote \Tx\ locations and light gray dots denote \Rx\ measurement locations}. The bottom two rows show floorplans learnt by {\nameLoS} \red{(i.e., Stage 1)} and {\name} \red{(i.e., Stage 2)} with sharper walls and boundaries. More visualizations available at \url{https://echonerf.github.io/} }
  \label{fig:ph2}
  % \vspace{-0.2in}
\end{figure*}

% \NeRFs\ relaxes physical reflection paths by modeling all voxels as potential virtual transmitters to approximate signal power, which leaves the opaque voxel locations under-determined based on the signal power measurements alone. 
% While the opaque virtual transmitters may fit the signal power data, these voxels do not represent true walls, as they do not adhere to specular reflection principles.

% With $N=2000$ $Rx$ locations, non-ray tracing methods like \ZYBase\ shows a smaller gap with {\name}. 
% This is expected because dense measurements increase the likelihood of receivers being located closely to opposite sides of walls, 
% resulting in a sharp boundary on the heatmap.
% \ZYBase’s performance degrades quickly when number of receiver locations decreases (e.g., 500 receiver locations).
% This is because, while signal power measurements can still be grouped, 
% the boundaries become less distinct as the distance between receivers increases. 
% In contrast, \name's performance is less sensitive to the distances between \Rx\ s and walls, as the reflection path length is incorporated into the loss function.

% For Signal Power Error, \name\ demonstrates performance comparable to \NeRFs. 
% \hl{This outcome confirms that NeRF can optimize Signal Power Error without accurately estimating the floor plan, as it does so by placing virtual transmitters throughout the scene.}

\subsection{Qualitative Results: Visual floorplans, RSSI heatmap, and basic ray tracing}
% We report qualitative results, including visual floorplans, predicted RSSI heatmap, and ray-tracing visualization.

\noindent $\blacksquare$ \textbf{Visual floorplans.} 
Figure~\ref{fig:ph2} presents visualization from all baselines and a comparison with our LoS-only model (as ablation).
All the floorplans use $N=2000$ receiver locations.
We make the following observations.
(1) {\ZYBase} leverages the difference of RSSI on opposite sides of a wall, however, reflections pollute this pattern, especially at larger distances between $Tx$ and $Rx$. 
Further, signals leak through open doors, injecting errors in the room boundaries.
(2) {\NeRFs} performs poorly since its MLP learns one among many possible assignments of virtual transmitters to fit the RSSI training data.
The virtual transmitters hardly correlate to the walls of the environment.
(3) {\nameLoS} can infer the position of inner walls.
However, these walls are thick and slanted because while {\nameLoS} can identify occlusions between a $\langle Tx, Rx \rangle$ pair, it cannot tell the shape and pattern of these occlusions.
Crucially, {\nameLoS} also cannot infer the boundary walls since no receivers are located outside the house.
(4) {\name} outperforms the baselines, sharpens the inner walls compared to {\nameLoS}, and constructs the boundary walls well.

\noindent \textbf{Shortcomings:} Recall that some parts of the floorplan are in the ``blind spots'' of our dataset since no reflection arrives from those parts to any of our sparse $Rx$ locations (e.g., see bottom left corner of the $1^{st}$ floorplan; no signals reflect off this region to arrive at any of the $Rx$ locations).
Hence, {\name} is unable to construct the bottom of the left wall in this floorplan.
Finally, note that areas outside the floorplan (e.g., the regions on the right of $6^{th}$ floorplan) cannot be estimated correctly since no measurements are available from those regions (hence, those voxels do not influence the gradients).

% We believe {\name} learns the floorplans better than baselines.
% {\name} $6$ floorplans estimated by \name\, the baselines ({\ZYBase} and \NeRFs), and one ablation version using only line-of-sight path \nameLoS. The visualizations are using \hl{$N=2000$} receiver locations. 
% As shown, \name\ consistently outperforms the baseline across all types of floor plans.

% Compared to {\name}, {\nameLoS} produces irregular and bulky wall structures for two main reasons: 
% (a) Like {\ZYBase}, the line-of-sight (LoS) model can only detect the presence of walls between a \Tx\ and a \Rx\ but cannot accurately estimate wall positions along the path. 
% Consequently, it may place a large wall between a \Tx\ and multiple \Rx\ s, leading to a consistent LoS loss. 
% (b) LoS paths are generally shorter and fewer than reflection paths, impacting fewer voxels. 
% Many voxels may not be reached by any LoS paths, allowing them to be either opaque or transparent without affecting the loss, which results in irregular boundaries.

% An additional advantage of including the reflection model is its ability to infer outer walls. 
% The LoS model requires \Tx\ s and \Rx\ s to be positioned on opposite sides of a wall to detect it, which is not feasible for outer boundaries, as all \Tx\ s and \Rx\ s are within the building. By leveraging reflection paths, the reflection model can instead estimate outer boundaries effectively.

\begin{figure*}[t]
  \centering
  % First figure (left)
  \begin{minipage}[t]{0.72\textwidth}
    \centering
    \vspace{0pt} 
    \includegraphics[width=1\columnwidth]{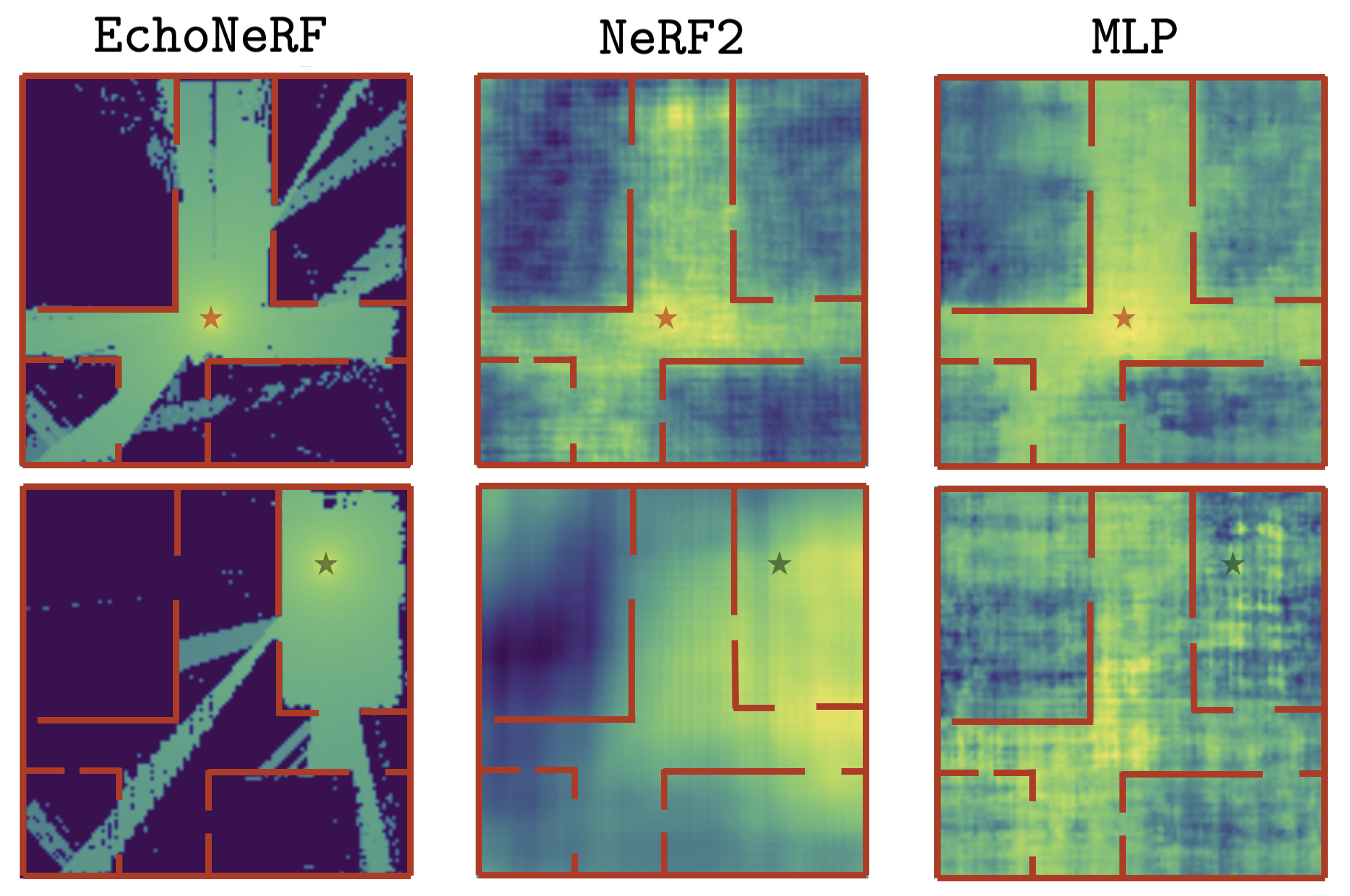}
    \caption{Heatmaps highlighting {\name}'s ability to learn signal propagation. (Top row) Inferred RSSI heatmaps with $Tx$ (red star) as used in training. (Bottom row) A new $Tx$ (green star) degrades {\NeRFs} and {\MLP} while {\name} shows accurate predictions.}
    \label{fig:ph4}
  \end{minipage}
  \vspace{-0.1in}
  \hfill
  % Second figure (right)
  \begin{minipage}[t]{0.24\textwidth}
    \centering
    \vspace{1em} 
    \includegraphics[width=.93\columnwidth]{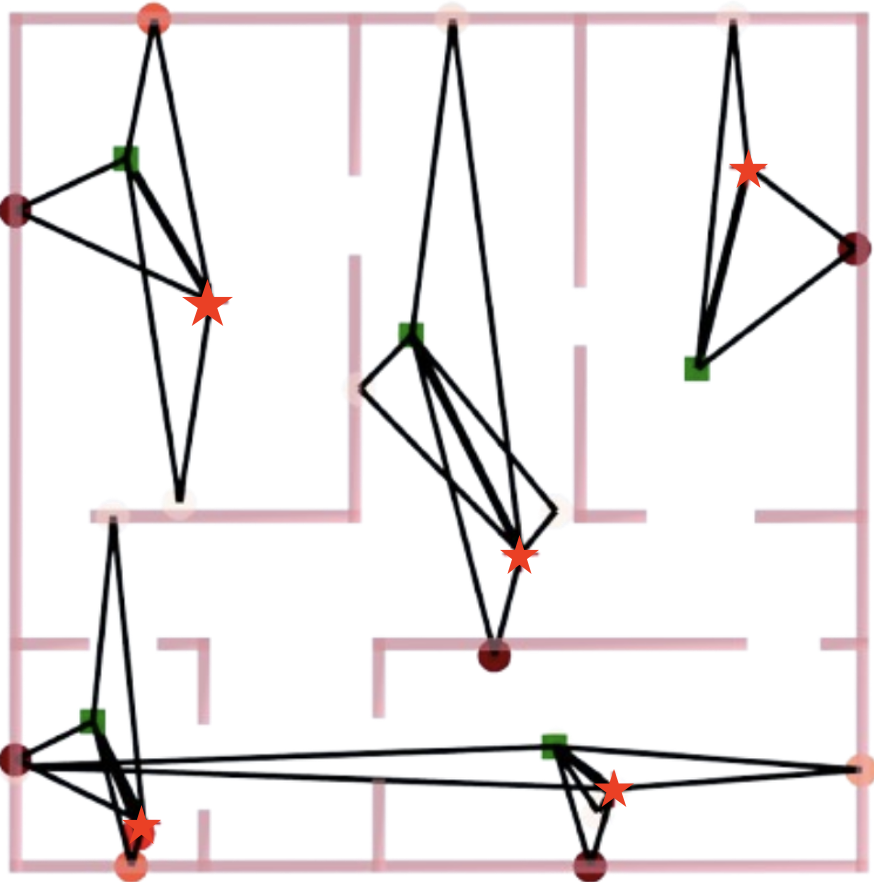}
    % \vspace{3em} 
    \includegraphics[width=.93\columnwidth]{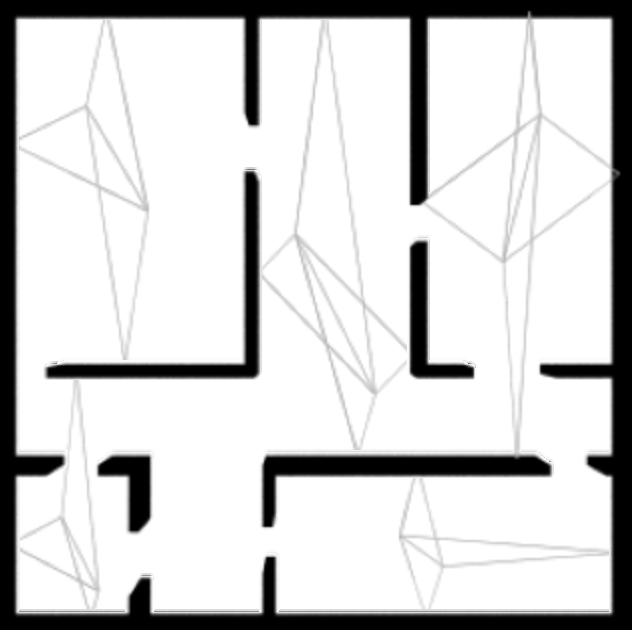}
    \caption{(a) Tracing reflections on the learnt floorplan. (b) True reflections from Sionna.}
    \label{fig:ph3}
  \end{minipage}
  \vspace{-0.1in}
\end{figure*}

\noindent $\blacksquare$ \textbf{RSSI prediction.} 
Figure~\ref{fig:ph4} visualizes and compares predicted RSSI.
The top row shows predictions at new $Rx$ locations with the $Tx$ held at the trained location; the bottom row shows predictions when both $Tx$ and $Rx$ are moved to new locations. Two key observations emerge:
(1) {\name} is limited by Sionna’s inability to simulate through-wall signal penetration; {\NeRFs} has access to an expensive license for a through-wall simulation and shows better predictions inside the rooms. However, in areas that {\name} can "see" (e.g., corridors in the top row), the awareness of reflecting surfaces leads to significantly better predictions. 
(2) When the $Tx$ location differs from that used in training, {\name}'s improvement over {\NeRFs} is significant.
This is the core advantage of first solving the inverse problem and then leveraging it for the (forward) RSSI prediction.

\noindent $\blacksquare$ \textbf{Learning reflected rays.}  
For a given $\langle Tx, Rx \rangle$ pair, we examine the points in the plausible set $\mathcal{V}$ that contribute to the reflections. 
Fig.~\ref{fig:ph3} compares the ray-tracing results from the NVIDIA Sionna simulator (we pick only first order reflections). 
{\name} captures many of the correct reflections.
Of course, some are incorrect -- a false positive occurs in the bottom right room since some wall segment is missing in our estimate; false negatives also occur in the top right room where again some parts of the wall are missing.

\subsection{Relaxing Assumptions \& Sensitivity Study}

\noindent $\blacksquare$ \textbf{Transmitter's location.} 
We assumed knowledge of $Tx$ locations, however, we relax this by applying maximum likelihood estimation on observed RSSI power, $\psi^*$ (see Appendix \ref{sec:relax}).
On average, the estimated \Tx\ location error is $2.08$ pixels in floorplans of sizes $\approx 512 \times 512$ pixels.

\begin{wrapfigure}{r}{0.45\textwidth}
  \centering
  \vspace{-1em} % optional: tighten top spacing
  \begin{tabular}{|c|c|c|c|c|}
    \hline
    Error $\sigma$(m) & 0 & 0.5 & 1 & 2 \\ \hline
    \texttt{Wall\_IoU} & 0.38 & 0.35 & 0.33 & 0.29 \\ \hline
  \end{tabular}
  \captionof{table}{Estimated Wall\_IoU at various levels of injected noise $\sigma$}
  \label{tab:ph5.2}

  \vspace{1.3em}

  \includegraphics[width=0.24\linewidth]{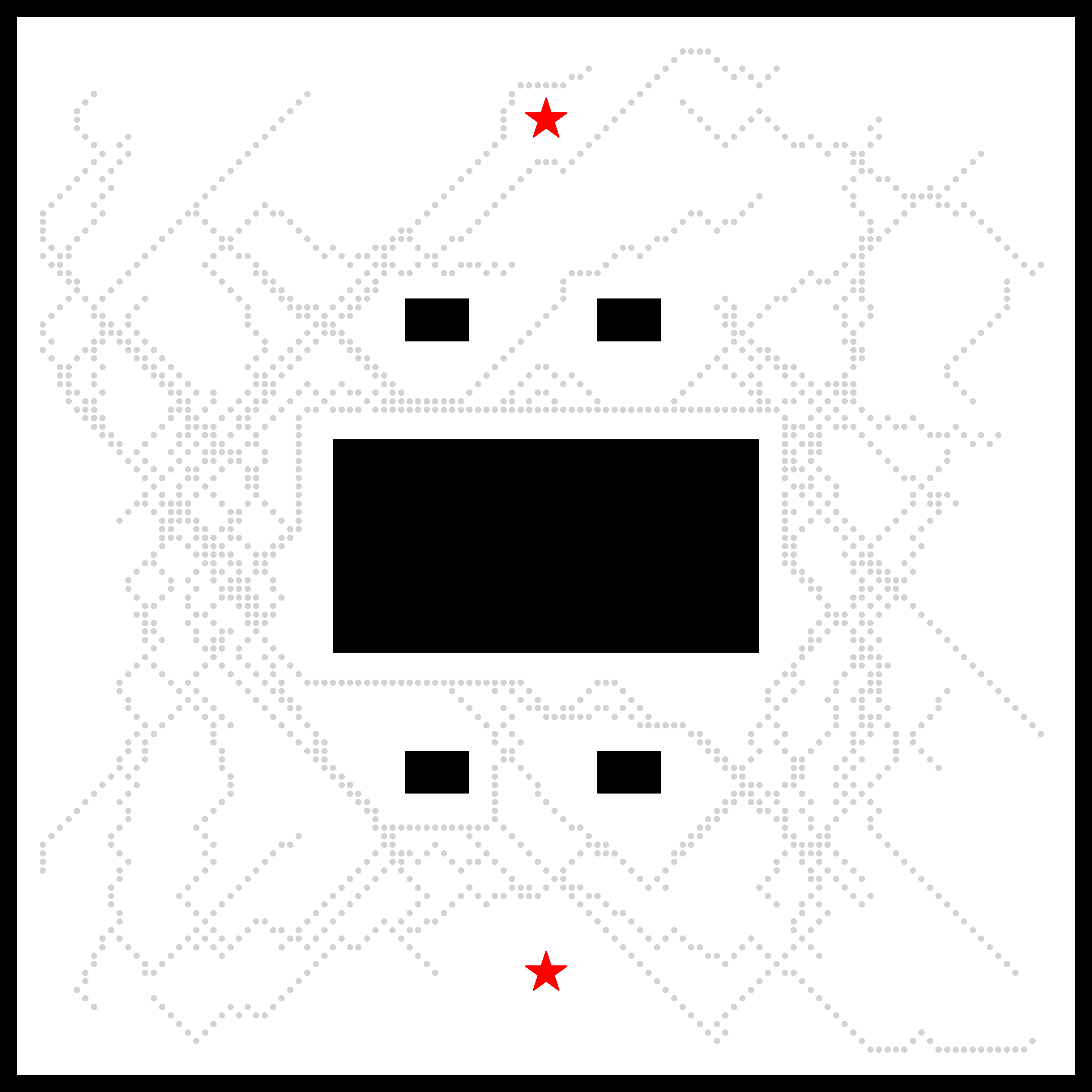}
  \includegraphics[width=0.24\linewidth]{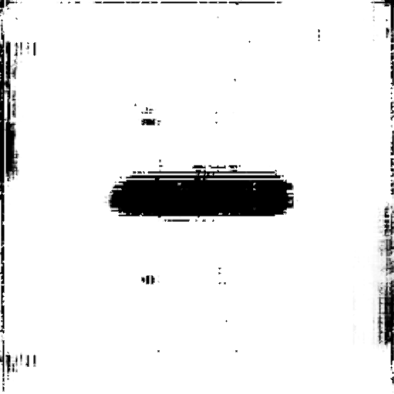}
  \includegraphics[width=0.24\linewidth]{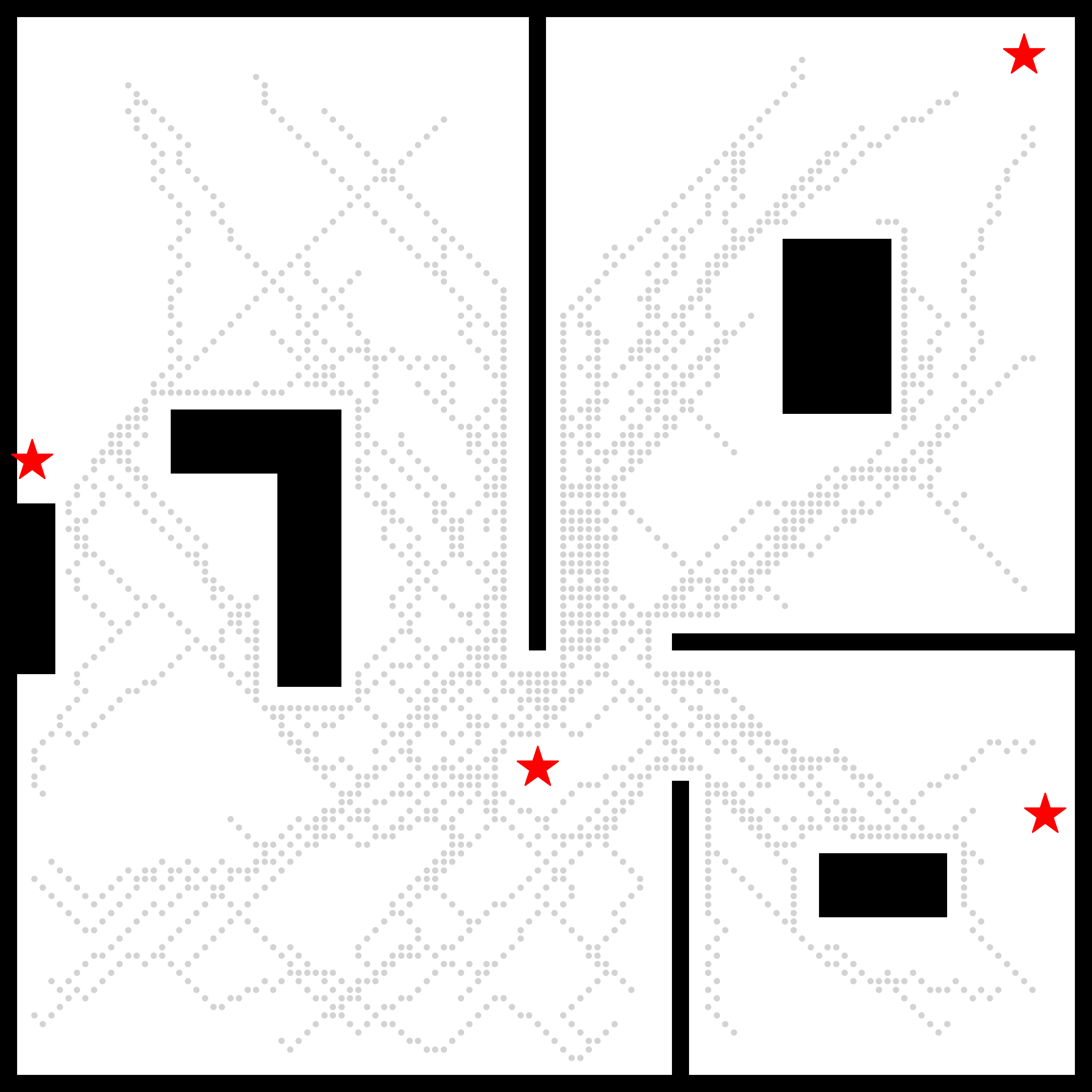}
  \includegraphics[width=0.24\linewidth]{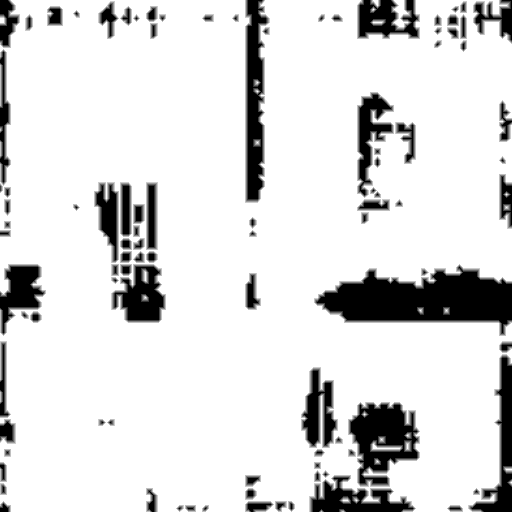}

  \captionof{figure}{\name's floorplan inference with furniture in conference (left) and apartment (right) layouts.}
  \label{fig:ph6}
  \vspace{-1.5em}
\end{wrapfigure}

\noindent $\blacksquare$ \textbf{Receiver location error.} 
% Indoor positioning systems are able to localize users but still incur some error.
Table \ref{tab:ph5.2} shows {\name}'s sensitivity to $Rx$ location errors. 
We inject Gaussian noise $\mathcal{N}(0,\sigma^2I)$ to the $Rx$ locations;   $\sigma=1$ implies a physical error of $1$m. 
\texttt{Wall\_IoU} accuracy obviously drops with error but $0.5$ meter of error is tolerable without destroying the floorplan structure. 
Advancements in WiFi positioning systems have demonstrated robust sub-meter error.

\noindent $\blacksquare$ \textbf{Effect of Furniture.}
Fig.~\ref{fig:ph6} visualizes inferred floorplans when toy objects are scattered in open spaces ($Rx$ locations remain $N$=$2000$).
{\name} is able to identify some of the object blobs but sharpening the small objects is challenging due to more higher order reflections from furniture.
Follow up work is needed, either in modeling $2^{nd}$ order reflections or by imposing stronger regularizations.
\vspace{-0.05in}

\red{\noindent $\blacksquare$ \textbf{Material Sensitivity.}
Table \ref{tab:material_sensitivity} shows the mean \texttt{Wall\_IoU} of \name\ on five different materials averaged across six scenes (shown in Fig. \ref{fig:ph2}) 
% to assess the effect of material reflectivity.
Materials with higher reflectivity, such as concrete and glass, yield better performance than absorptive materials like wood. This is because more reflections allow better performance for \name\'s reflection model.}

% \begin{table}[t]
% \centering
% \small
% \setlength{\tabcolsep}{6pt}
% \renewcommand{\arraystretch}{1.15}
% \begin{tabular}{|l|c|c|}
% \hline
% \textbf{Material} & \nameLoS{$\uparrow$} & \name{$\uparrow$} \\
% \hline
% Concrete & 0.251 & 0.371 \\
% Glass    & 0.236 & 0.364 \\
% Brick    & 0.232 & 0.357 \\
% Marble   & 0.226 & 0.328 \\
% Wood     & 0.227 & 0.311 \\
% \hline
% \end{tabular}
% \caption{Material sensitivity across five materials.}
% \label{tab:material_sensitivity}
% \end{table}

\red{\noindent $\blacksquare$ \textbf{Robustness to RSSI error.}
We added Gaussian noise to the RSSI measurements with a mean equal to the noise floor (in dB) and a variance of $4$\,dB. We vary the noise floor levels ranging from $-80$\,dB to $-130$\,dB across the 6 floorplans shown in Fig. \ref{fig:ph2}. The SNR at a receiver is computed as the difference between the received signal power and the noise floor (e.g., a received power of $-70$\,dB with a noise floor of $-80$\,dB results in an SNR of $10$\,dB).
We report the mean \texttt{Wall\_IoU} for \nameLoS\ and \name\ in Table \ref{tab:noise_robustness}.
% Both models produce legible floorplans down to $\approx 30$\,dB SNR, and deteriorate when noise approaches signal power.
The performance drops with decreasing SNR; both \nameLoS\ and {\name}’s floorplans are still legible till $30$\,dB, but below that (when noise power becomes comparable to the signal power) they break down, leading to missing and illegible walls.
Our results are conservative; when we report a specific SNR level (e.g., 40dB), it represents the highest SNR (best-case) among all receivers in that scenario, meaning other receivers experience even lower SNRs.
% , making our evaluation conservative.
In practice, WiFi SNR ranges from 30-60dB depending on the distance from the router, with close-proximity measurements often exceeding 50-60dB.
For average real-world SNR conditions around 45dB, the corresponding best-case SNR would be 60+dB, which aligns with the top rows where {\name} demonstrates strong performance.}

% \begin{table}[t]
% \centering
% \small
% \setlength{\tabcolsep}{6pt}
% \renewcommand{\arraystretch}{1.15}
% \begin{tabular}{|l|c|c|c|}
% \hline
% \textbf{SNR (dB)} & \nameLoS{$\uparrow$} & \name{$\uparrow$} & \textbf{Qualitative} \\
% \hline
% $\infty$ (no noise) & 0.251 & 0.371 & Legible \\
% 60                  & 0.246 & 0.336 & Legible \\
% 50                  & 0.231 & 0.292 & Legible \\
% 40                  & 0.226 & 0.298 & Legible \\
% 30                  & 0.207 & 0.241 & Missing walls \\
% 10                  & 0.090 & 0.140 & Illegible \\
% \hline
% \end{tabular}
% \caption{Noise robustness across SNR levels.}
% \label{tab:noise_robustness}
% \end{table}

\begin{table}[t]
\centering

\begin{minipage}[t]{0.43\linewidth}
\centering
\footnotesize
\setlength{\tabcolsep}{5pt}
\renewcommand{\arraystretch}{1.10}
\resizebox{\linewidth}{!}{%
\begin{tabular}{|l|c|c|}
\hline
\textbf{Material} & \nameLoS{$\uparrow$} & \name{$\uparrow$} \\
\hline
Concrete & 0.251 & 0.371 \\
Glass    & 0.236 & 0.364 \\
Brick    & 0.232 & 0.357 \\
Marble   & 0.226 & 0.328 \\
Wood     & 0.227 & 0.311 \\
\hline
\end{tabular}}
\vspace{0.1in}
\caption{Sensitivity across materials.}
\label{tab:material_sensitivity}
\end{minipage}
\hfill
\begin{minipage}[t]{0.53\linewidth}
\centering
\footnotesize
\setlength{\tabcolsep}{4pt}
\renewcommand{\arraystretch}{1.08}
\resizebox{\linewidth}{!}{%
\begin{tabular}{|l|c|c|c|}
\hline
\textbf{SNR (dB)} & \nameLoS{$\uparrow$} & \name{$\uparrow$} & \textbf{Qualitative} \\
\hline
$\infty$ (no noise) & 0.251 & 0.371 & Legible \\
60                  & 0.246 & 0.336 & Legible \\
50                  & 0.231 & 0.292 & Legible \\
40                  & 0.226 & 0.298 & Legible \\
30                  & 0.207 & 0.241 & Missing walls \\
10                  & 0.090 & 0.140 & Illegible \\
\hline
\end{tabular}}
\vspace{0.1in}
\caption{Noise robustness across SNR levels.}
\label{tab:noise_robustness}
\end{minipage}
\vspace{-0.25in}
\end{table}
\vspace{-0.05in}

\section{Follow ups and Conclusion}
\vspace{-0.05in}
\noindent\textbf{\hl{Follow-ups.}} 
\label{sec:limitations}
% (1) The Sionna simulator does not model through-wall signal penetration, hence we have placed a $Tx$ in each room.
% In reality (or in expensive ray tracing simulators \cite{remcom_wireless_insite}), WiFi signals will penetrate walls.
% This is an advantage since signals from a single $Tx$ will be measurable across the whole home.
% However, voxel opacities will no longer be bimodal ($0$ or $1$), hence {\name} will need to assign $\delta \in [0,1]$ to match the measured RSSI.
% This will require modifications to our models.
(1) The ability to model $2^{nd}$ order reflections will boost {\name}'s accuracy, allowing it to sharpen the scene and decode smaller objects. 
For short range applications, such as non-intrusive medical imaging, $2^{nd}$ and $3^{rd}$ order reflections would be crucial.
This remains an important direction for follow-on research.
(2) Extending {\name} to 3D floorplans is also of interest, and since it is undesirable to increase the number of measurements, effective 3D priors, or 2D-to-3D post-processing, may be necessary. 
Such post-processing tools exist \cite{cedreo2025} but we have not applied them since our goal is to improve NeRF's inherent inverse solver.
\red{(3) Expanding evaluations beyond ZInD, which contains largely unfurnished, rectangular rooms, to richer datasets such as HM3D \cite{hm3d} and MVL \cite{mvl} would also be valuable:
furniture and clutter can introduce significant multipath effects that complicate RF signal modeling, and additional research is needed to understand these effects.
We defer this to future work, as our focus in this paper is not so much to understand the limits of RF-based NeRFs, but to establish the feasibility of such frameworks.}
% A naive extension to 3D would lead to increased computational load, necessitating careful tweaks to {\name}'s core architecture. 
% (3) Fusing additional signal modalities, such as audio (from music loudspeakers), or overhearing from many WiFi enabled devices at home, could lead to increased measurement density. 
% % further enhance inference by capturing more nuanced room layouts. 
(4) Finally, {\name} floorplans can offer valuable spatial context to Neural RIR synthesizers like \cite{zhao2023nerf2, newrf, brunetto2024neraf, AV-NeRF, NACF}. 
Synthesized RIR could in-turn aid {\name}'s floorplan inference, forming the basis for an alternating optimization strategy. We leave these ideas to follow-up research. \\
\noindent\textbf{Conclusion.} 
% In summary, we propose {\name} to infer indoor floorplans from RF signals by adapting the classic NeRF framework to tackle inverse problems. 
In summary, we re-design the NeRF framework so it can learn to "see" its environment by leveraging both line-of-sight (LoS) paths and multipath reflections. 
While such reflections bring to the sensor more information about the surroundings, their mixtures with the LoS path also complicates the core inverse problems.
{\name} takes a step towards solving this inverse problem, but also leaves room for further research in neural wireless imaging and varies downstream applications.  

\subsection*{Acknowledgments}
We would like to thank Prof. Shenlong Wang for his guidance in the initial phase of this work.
% and Prof. Srihari Nelakuditi for providing computational guidance. 
We also thank the anonymous reviewers for their valuable feedback.
This work was partially supported by NSF \#2008338, \#1909568, \#2148583, and \#MRI-2018966.
This work used DELTA at NCSA through allocation  CIS230230 from the Advanced Cyberinfrastructure Coordination Ecosystem: Services \& Support (ACCESS) program, which is supported by U.S. National Science Foundation grants \#2138259, \#2138286, \#2138307, \#2137603, and \#2138296

\clearpage 
\bibliographystyle{plain} 
\bibliography{main} 

\newpage

% \clearpage
\begin{center}
    \Large \textbf{Appendix for}\\[0.3em]
    \Large \textbf{Can NeRFs \hspace{0.01pt} ``See'' \hspace{0.01pt} without Cameras?}
\end{center}
\vspace{1em}

% \section*{Appendix}

\appendix

\section{Background on Channel Impulse Response}
\label{sec:cirbackground}
When a wireless signal propagates, it is typically influenced by multipath effects such as reflections and scattering, as well as attenuation caused by the surrounding environment—collectively referred to as the channel. The overall impact of these phenomena on the signal is characterized by a linear model known as the Channel Impulse Response (CIR) \cite{cir}. 

Mathematically, the CIR is expressed as a sum of scaled and delayed impulses as shown in Eqn \ref{eqn:CIR}.
\begin{align} \label{eqn:CIR}
   h(t) = \sum_{i=1}^N a_i e^{j\phi_i} \delta(t - \tau_i), 
\end{align}
where $N$ is the number of multipath components, $a_i$ denotes the amplitude (attenuation factor) of the $i$-th path, $\phi_i$ represents the phase shift of the $i$-th path, and $\tau_i$ is the delay of the $i$-th path.

For an input signal $x(t)$ transmitted through the channel $h(t)$, the output signal measured at a \Rx, $y(t)$ is obtained by the convolution:
\begin{align}
    y(t) = x(t) * h(t) + w(t),
\end{align}
where $w(t)$ represents zero-mean additive noise. For a simple two-path channel with a line of sight (LoS) path and one reflected path, the CIR $h(t)$ is given as
\begin{align}
h(t) = a_1 \delta(t) + a_2 e^{j\phi_2} \delta(t - \tau_2),
\end{align}

The received signal $y(t)$ would then be:
\begin{align}
y(t) = x(t) * \left[ a_1 \delta(t) + a_2 e^{j\phi_2} \delta(t - \tau_2) \right] + w(t)
\end{align}

\section{Modelling Wideband Multipath Signal Power}
\label{sec:modelling_signal_power}

This section shows how received power in multipath scenarios can be approximated as a sum of powers of LoS and all other multipaths. In frequency domain, the received signal $y(t)$ at a particular receiver can be expressed as 

\begin{align}
    Y(f_k) = H(f_k) X(f_k) + W(f_k)
\end{align}

Here, $X(f_k)$ and $W(f_k)$ represent the discrete Fourier transforms of the signal $x(t)$ and the additive noise $w(t)$ at subcarrier frequency $f_k$ with $k \in \{0, 1, ..., K\}$. The channel can be written as $H(f_k) = \sum_{l=0}^{L}  a_{lk} \exp(-j2\pi f_k\tau_l)$ where $l \in \{1, 2, ..., L\}$ is an index over separate multipaths. Here, $a_{lk}$ and $\tau_l$ represents the attenuation and phase of the $l$-th multipath component at subcarrier $k$. We assume that the channel $H(f_k)$ and the signal $X(f_k)$ are independent. 

The received power at each frequency $k$ is given by $\psi_{Y_k} = \E[\left|Y(f_k)\right|^2]$. 
\begin{align}
    \E[\left|Y(f_k)\right|^2] &= \E[\left|H(f_k) X(f_k) + W(f_k)\right|^2] \\ \nonumber
    &= \E[\left|H(f_k)\right|^2\left|X(f_k)\right|^2] + \E[\left|W(f_k)\right|^2] + 2\text{Re}\{\E[H(f_k)X(f_k)W(f_k)]\} \\ \nonumber
    &= \E[\left|H(f_k)\right|^2]\E[\left|X(f_k)\right|^2] + \E[\left|W(f_k)\right|^2] + 2\text{Re}\{\E[H(f_k)X(f_k)]\E[W(f_k)]\} \\ \nonumber
    &= \E[\left|H(f_k)\right|^2]\E[\left|X(f_k)\right|^2] + \E[\left|W(f_k)\right|^2], \because \E[W(f_k)] = 0 \\ \nonumber
    &= \psi_{X_k} \E[\left|H(f_k)\right|^2] + \psi_{W_k}
\end{align}

where $\psi_{X_k} = \E[\left|X(f_k)\right|^2]$ and $\psi_{W_k} = \E[\left|W(f_k)\right|^2]$. Next,

\begin{align}
    \E[\left|H(f_k)\right|^2] &= \E[H(f_k)H^*(f_k)] \text{\footnotemark} \nonumber \\ 
    &= \E\left[\left(\sum_{l=0}^{L} a_{lk} \exp \left(-j2\pi f_k\tau_l\right)\right) \left(\sum_{m=0}^{L} a_{mk}^* \exp \left(j2\pi f_k\tau_m\right)\right)\right] \nonumber \\ 
    &= \E\left[\left(\sum_{l=0}^{L}\sum_{m=0}^{L} a_{lk}a_{mk}^* \exp \left(-j2\pi f_k\left(\tau_l - \tau_m\right)\right)\right)\right] \nonumber \\ 
    &= \E\left[\sum_{l=0}^{L} |a_{lk}|^2 + \sum_{l=0}^{L}\sum_{\substack{m=0 \\ m \neq l}}^{L} a_{lk}a_{mk}^* \exp \left(-j2\pi f_k\left(\tau_l - \tau_m \right)\right) \right] \nonumber \\ 
    &= \E\left[\sum_{l=0}^{L} |a_{lk}|^2\right] + \sum_{l=0}^{L}\sum_{\substack{m=0 \\ m \neq l}}^{L} \E\left[a_{lk}a_{mk}^* \right] \exp \left(-j2\pi f_k\left(\tau_l - \tau_m \right)\right) \nonumber \\
    &= \E\left[\sum_{l=0}^{L} |a_{lk}|^2\right] \nonumber \
\end{align}
\footnotetext{We use * to denote complex conjugate}
We assume that channel gains between different multipaths $l, m \ \text{with} \ l \neq m$ have vanishingly small correlation. Therefore, the total received power $\psi$ can be computed by summing all individual subcarrier powers. 
\begin{align}
    \psi &= \sum_{k = 1}^{K} \psi_{Y_k} \nonumber \\
        &= \sum_{k = 1}^{K} \psi_{X_k} \E\left[\sum_{l=0}^{L} |a_{lk}|^2\right] + \psi_{W_k} \nonumber \\
        &= \sum_{l=0}^{L} \sum_{k = 1}^{K} \E\left[\psi_{X_k} |a_{lk}|^2 \right] + \psi_{W_k} \nonumber
        % &= \psi_{LoS} + \sum_{l=1}^{L} \psi_{ref_l} + L\psi_W
\end{align}

Here, we separate out $\sum_{k = 1}^{K} \E\left[\psi_{X_k} |a_{0k}|^2 \right] = \psi_{LoS}$, to represent the LoS power over all subcarriers and $\sum_{k = 1}^{K} \E\left[\psi_{X_k} |a_{lk}|^2 \right] = \psi_{ref_l}, 1 \leq l \leq L$, to denote the power of $l$-th order reflections. 

%\section{$\psi_{ref_1}$'s Contribution}
% \section{Received power from reflections}
\section{Approximating Channel with First-Order Reflections}
\label{sec:highorder}

\name\ models the total received power at the {\Rx} as the combination of the LoS power \nameLoS, and the contributions from all the first-order reflections.
To validate the contribution of the achievable power from \name\ when compared to the total received power $\psi$, we evaluate the relative contributions of these signals to the total power using the NVIDIA Sionna simulator \cite{hoydis2022sionna}. To this end, we compute the ratios of the LoS signal $\psi_{LoS}$, LoS with the first order reflections $\psi_{LoS} + \psi_{ref_1}$, and LoS with the first two orders of reflections $\psi_{LoS} + \psi_{ref_1} + \psi_{ref_2}$. These are compared to the total received power $\psi$, which is approximated as the sum of the LoS power and the power from the first ten reflections.

% alongside the total received power $\psi$ ( approximated as the sum of the LoS power and the power from the first ten reflections)

\begin{figure}[!t]
  %\centering{0.45\columnwidth} % Use \columnwidth to make it fit within the single column
    \centering
    \includegraphics[width=1.0\columnwidth]{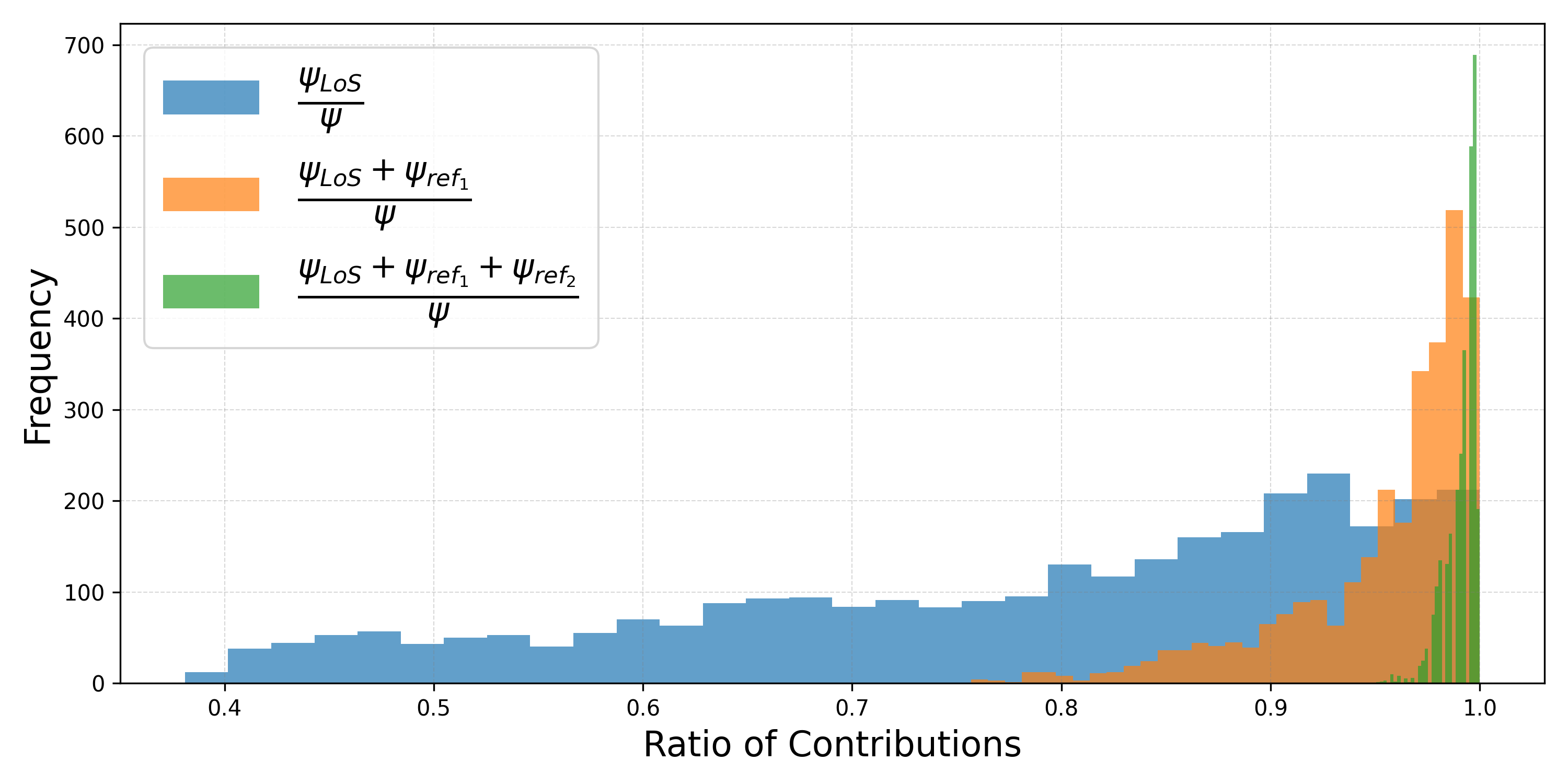}
    \caption{Histograms illustrating the contribution ratios from line-of-sight (LoS), LoS combined with first-order reflections, and LoS combined with first- and second-order reflections. The orange graph highlights the significant contribution of first-order reflections to the total power, supporting \name's approach of modeling only the first reflection alongside the LoS power.}
    % \caption{Histograms showing ratios of contribution from LoS, LoS and first-order reflection and LoS and first and second-order reflections.}
    % \caption{}
    \vspace{-0em}
    \label{fig:supp-contributions}
\end{figure}

Fig \ref{fig:supp-contributions} shows path power contribution ratio from different paths in histogram. While the $\psi_{LoS}$ power alone only accounts for approximately 70\% of the total received power and is more spread out, $\psi_{LoS} + \psi_{ref_1}$ accounts to 95\% of the total power, with a reduced spread. Moreover, secondary reflections $\psi_{ref_2}$ only contribute to less than 3\% of the total power.
Hence, \name\ models the first-order reflections along with the line-of-sight.

% Relying solely on modeling LoS power is insufficient to model the total received power accurately, and hence \name\ also models the first-order reflections that cannot be attributed by line of sight alone.

% \section{Romit's: Relaxing assumption on \Tx\ location}
% We relax the assumption that \Tx\ locations are known. 
% The main idea is to apply a maximum likelihood estimate (MLE).
% Briefly, among all the measured signal powers $\{\psi_j^*\}$ from a given \Tx\, we identify the $P$ strongest signal powers and their corresponding received locations.
% The motivation to choose the strongest powers is that they would be significantly dominated by the LoS component, hence easy to model using Friss' equation \cite{antennaBook}.
% The likelihood equation for all these $P$ measurements can be written as:
% \[
% p(\psi_1^*, \psi_2^*, \dots, \psi_P^* | \mathbf{Tx}) = \prod_{i=1}^{N} p(\psi_i^* | \mathbf{Tx})
% \]
% We assume independence among the measurements since the received LoS power across locations, for a given a \Tx\ location, are independent.

\vspace{0.3cm}
\section{Relaxing \Tx\ Assumptions}
\label{sec:relax}
\begin{figure*}[h]
  \centering
    \begin{tabular}{ @{\hskip 5pt} l @{\hskip 5pt} c@{ } c @{ } c @{ } c @{ } c @{ } c }
    % \hline
    \parbox[c][0.6cm][c]{0.1\textwidth}{\centering \vspace{-2cm}Ground Truth
    } & 
    \includegraphics[width=0.14\textwidth]{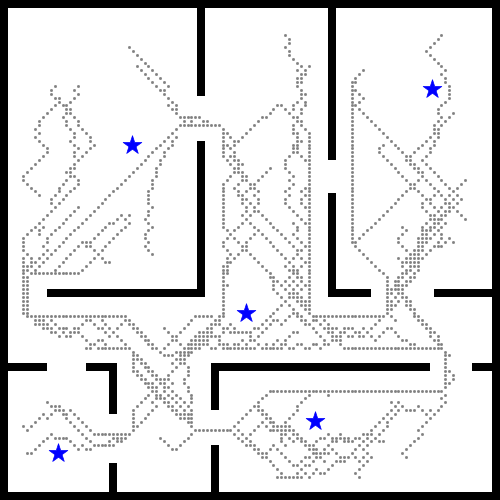}&
    \includegraphics[width=0.14\textwidth]{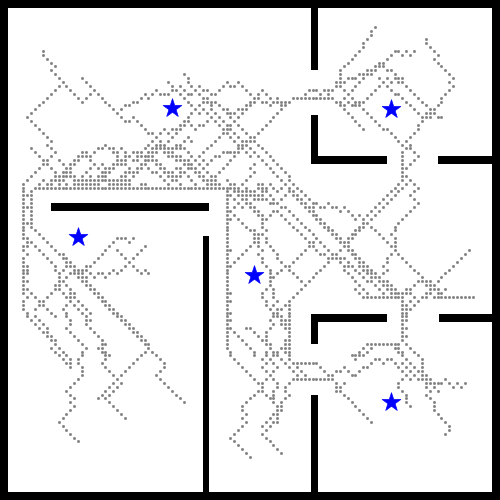}&
    \includegraphics[width=0.14\textwidth]{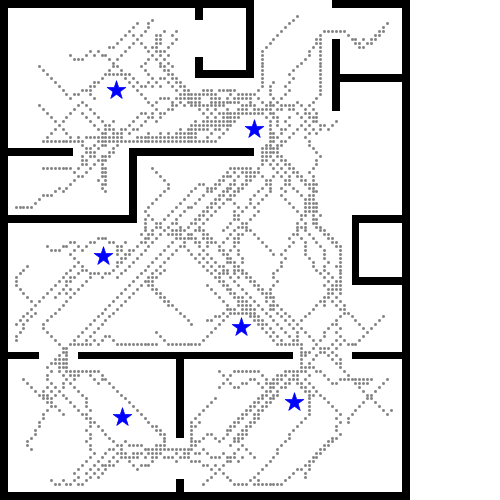}&
    \includegraphics[width=0.14\textwidth]{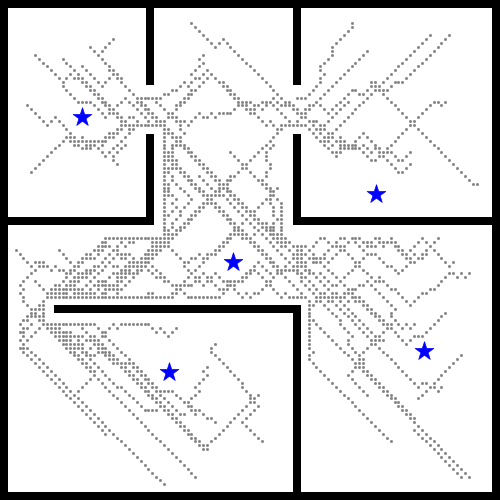}& 
    \includegraphics[width=0.14\textwidth]{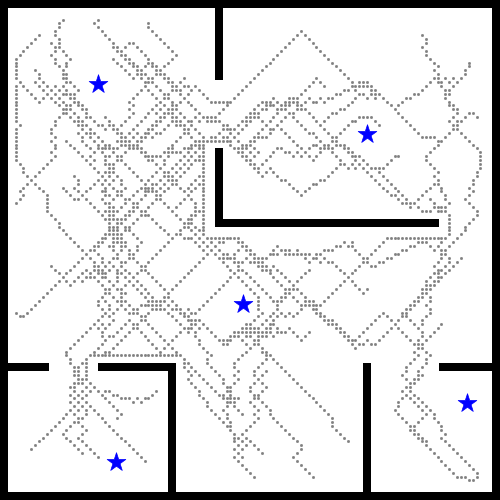}&
    \includegraphics[width=0.14\textwidth]{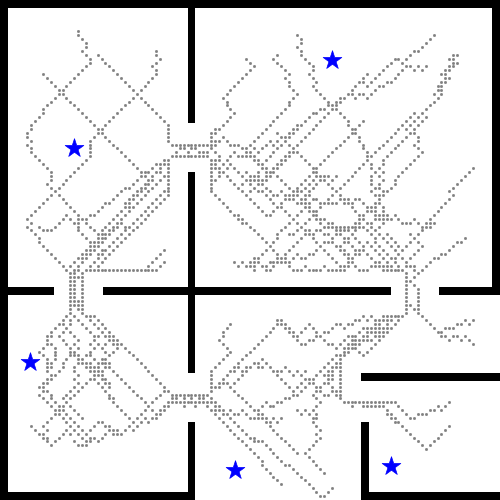}
    \vspace{0.3cm}
    \\
    \parbox[c][1.2cm][c]{0.1\textwidth}{\centering \vspace{-2cm} \texttt{EchoNeRF LoS}} & 
    \includegraphics[width=0.14\textwidth]{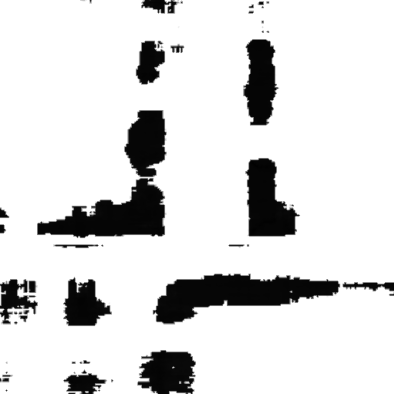}&
    \includegraphics[width=0.14\textwidth]{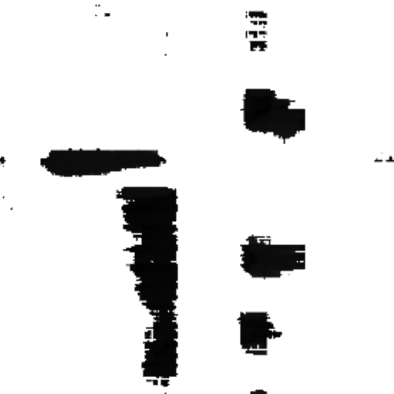}&
    \includegraphics[width=0.14\textwidth]{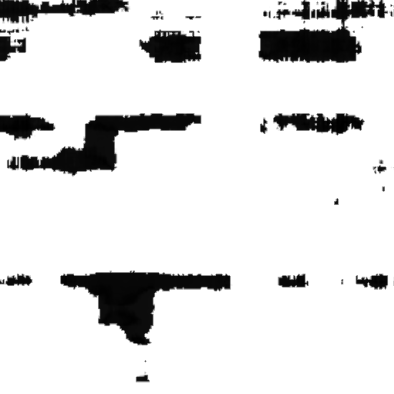}&
    \includegraphics[width=0.14\textwidth]{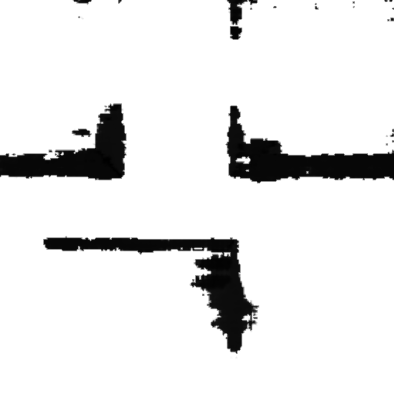}&
    \includegraphics[width=0.14\textwidth]{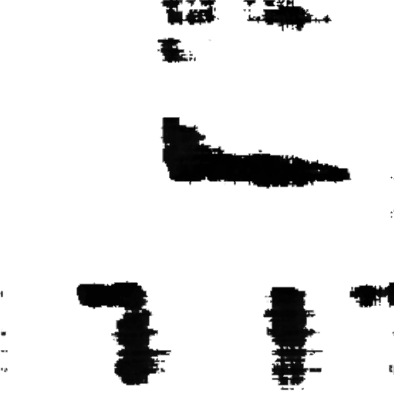}&
    \includegraphics[width=0.14\textwidth]{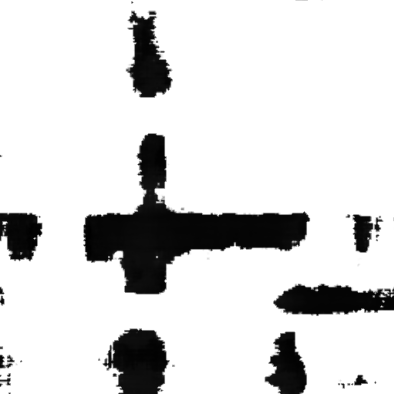}
    \\
    \parbox[c][.6cm][c]{0.1\textwidth}{\centering \vspace{-2cm} \name} &
    \includegraphics[width=0.14\textwidth]{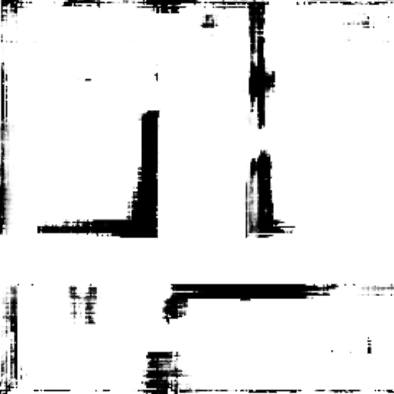}&
    \includegraphics[width=0.14\textwidth]{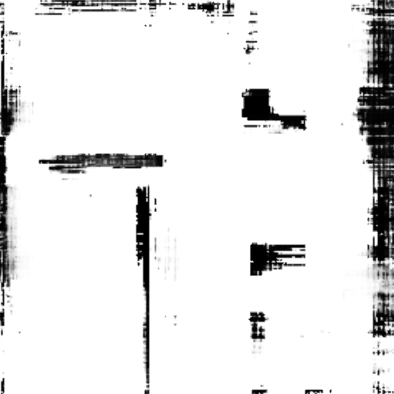}&
    \includegraphics[width=0.14\textwidth]{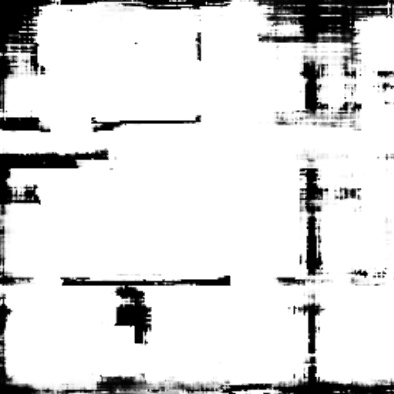}&
    \includegraphics[width=0.14\textwidth]{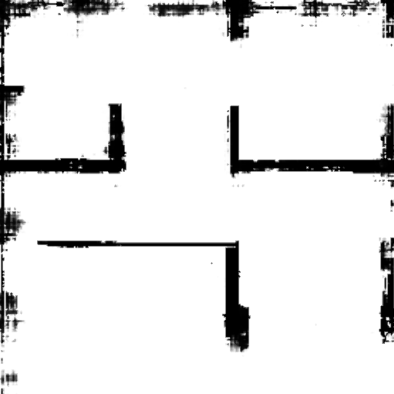}&
    \includegraphics[width=0.14\textwidth]{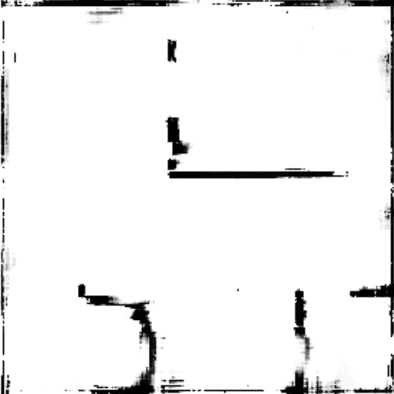}&
    \includegraphics[width=0.14\textwidth]{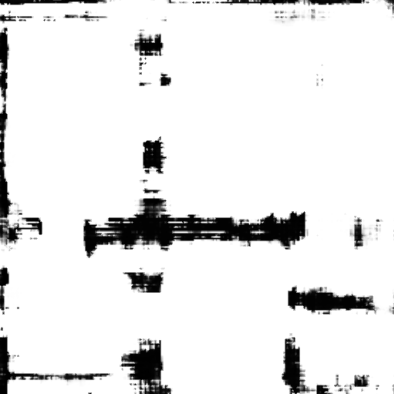}
    \vspace{1cm}
    \\
    % \hline
  \end{tabular}
  % \vspace{-0.15in}
  \caption{Qualitative comaprison of \nameLoS\ and \name\ when \Tx\ locations are unknown, and are estimated. The top row shows the ground truth floorplan, \Rx\ locations along with the estimated {\Tx}s in blue. The second and third row displays the performance of \nameLoS\ and \name\ respectively. Despite the \Tx\ locations being unknown, our methods accurately estimate them, leading to performance comparable to the case where \Tx\ locations are known.}
  
  \label{fig:supp-txresults}
  % \vspace{-0.1in}
\end{figure*}

\begin{figure*}[h]
    \centering
    \includegraphics[width=1\textwidth]{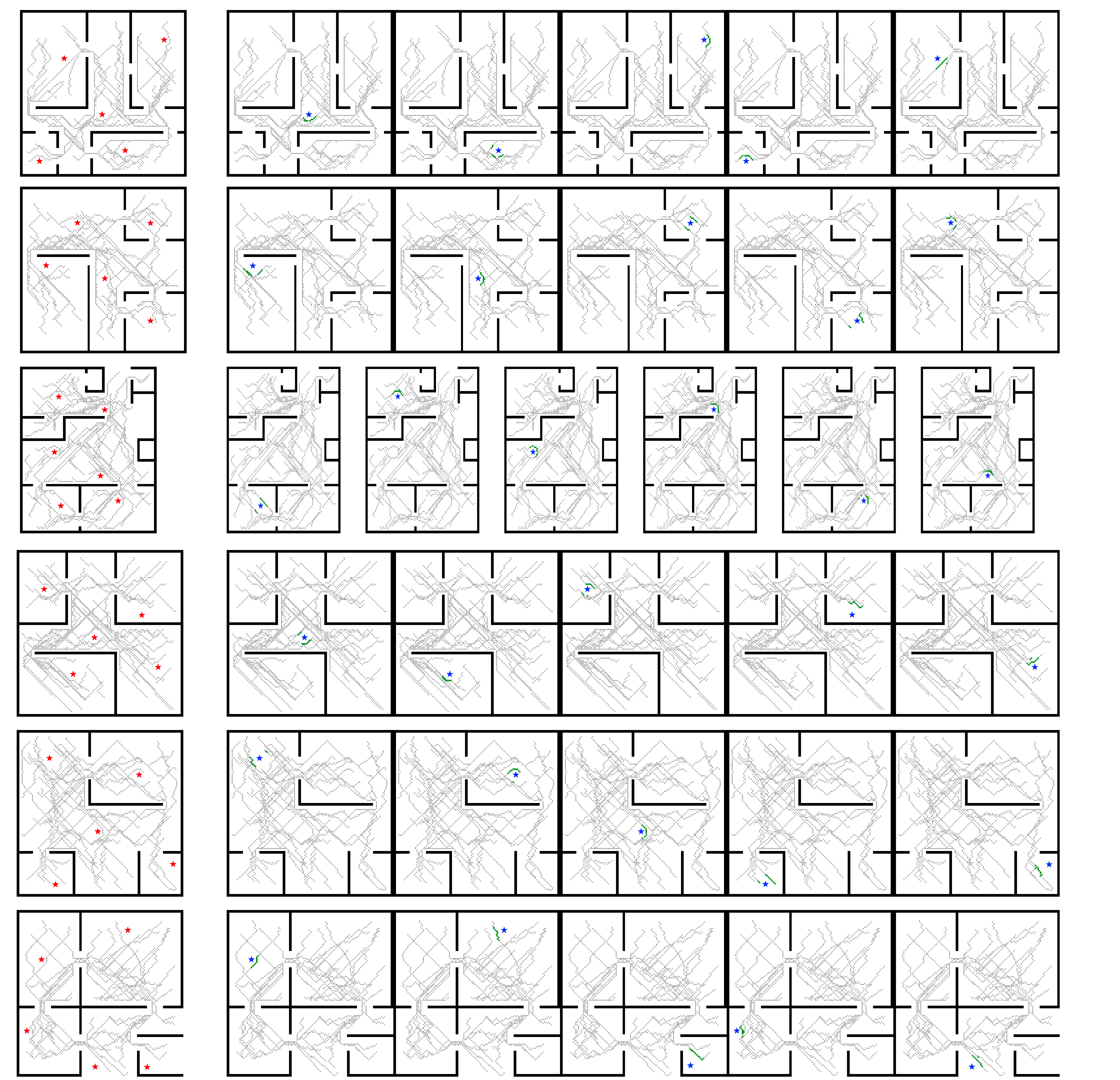}
    \caption{Comparison of Ground truth \Tx\ locations indicated in red in the first column with the estimated \Tx\ locations shown in blue from starting from column two. The \Rx\ positions used for the estimation are marked in green.}
  \label{fig:supp-txestimates}
\end{figure*}

We relax the assumption that \Tx\ locations are known. 
Given the set of receiver locations $\{\mathbf{Rx}_i\}$ and the signal powers $\{\psi_i\}$, the goal is to estimate the transmitter location \(\mathbf{Tx} = (\mathbf{Tx_{x}}, \mathbf{Tx_{y}})\).
% \Tx\ location $\mathbf{Tx}$, and $d_i = ||\mathbf{Tx} - \mathbf{Rx}_i||$,
% using the observed power \(\psi_i^*\) just based on the line-of-sight model.
To achieve this, we apply a maximum likelihood estimate (MLE). 
Briefly, among all the measured signal powers $\{\psi_i\}$ from a given \Tx\, we identify the $P$ strongest signal powers and their corresponding received locations.
The rationale behind selecting the strongest powers is that they are significantly influenced by the LoS component, allowing us to model them effectively only using the Friss' equation \cite{antennaBook}.
We assume independence among the measurements since the received LoS power across locations, for a given a \Tx\ location, are independent.
So, the likelihood equation for all these $P$ measurements can be written as:
\[
p(\psi_1, \psi_2, \dots, \psi_P | \mathbf{Tx}) = \prod_{i=1}^{P} p(\psi_i | \mathbf{Tx})
\]
% Since the observed power values \(\psi^*\) are independent given the true \Tx\ location, the likelihood function for all observations can be written as the product of individual likelihoods for each \(\psi_i^*\)
% \[
% p(\psi_1^*, \psi_2^*, \dots, \psi_N^* | \mathbf{Tx}) = \prod_{i=1}^{N} p(\psi_i^* | \mathbf{Tx})
% \]
We approximate that the \(\psi_i\) is normally distributed with a mean modeled by the line-of-sight power \(\frac{K}{d_i^2}\) and variance \(\sigma^2\) where $d_i = ||\mathbf{Tx} - \mathbf{Rx}_i||$ is the distance between $\mathbf{Tx}$ and $\mathbf{Rx}_i$.
The likelihood function for each observation \(\psi_i\) is thus given by:
\[
p(\psi_i | \mathbf{Tx}) = \frac{1}{\sqrt{2\pi \sigma^2}} \exp \left( -\frac{(\psi_i - \frac{K}{d_i^2})^2}{2\sigma^2} \right)
\]
Maximizing log-likelihood  $L$ of \(\{\psi_i \}\) \(\forall\ i \in \{1,\ldots, P\}\)
\[
\log L(\mathbf{Tx}) = \sum_{i=1}^{P} \log \left( \frac{1}{\sqrt{2\pi \sigma^2}} \exp \left( -\frac{(\psi_i - \frac{K}{d_i^2})^2}{2\sigma^2} \right) \right)
\]

\[
\log L(\mathbf{Tx}) = -\frac{P}{2} \log(2\pi \sigma^2) - \frac{1}{2\sigma^2} \sum_{i=1}^{P} \left(\psi_i - \frac{K}{d_i^2}\right)^2
\]
Minimizing the second term gives the optimal \(\mathbf{Tx}^*\) as:

\[
{\mathbf{Tx}^*} = \argmin_{\mathbf{Tx}} \sum_{i=1}^{P} \left(\psi_i - \frac{K}{\|\mathbf{Tx} - \mathbf{Rx}_i\|^2_2}\right)^2
\]
% Optimizing for the $\mathbf{Tx}$ and simplifying yields the following set of non-linear equations.
% \[
% \psi_i^* \left( \|\mathbf{Tx} - \mathbf{Rx}_i \|^2_2 \right) = K  \quad \forall\ i
% \]
% Thus, the optimization problem is reduced to solving this system of non-linear equations for $\mathbf{Tx}$. 
% We first sort the received signals \(\psi_i^*\)  in descending order of the power and use the first $P$ values to estimate the \Tx\ location.
We use Scipy's `minimize' with the BFGS method to numerically solve for $\mathbf{Tx}^*$. 
% \widehat{\mathbf{Tx}}
Fig \ref{fig:supp-txestimates}. visualizes the ground truth and the estimated \Tx\ locations across 6 floorplans. The estimated \Tx\ positions closely match the ground truth, and we report the \Tx\ location error to be 2.08 pixels. 
% should we compare the 
Fig \ref{fig:supp-txresults}. demonstrates the performance of \nameLoS\ and \name\ using the estimated \Tx\ locations for the 6 floorplans in Fig \ref{fig:supp-txestimates}. 
Performance is comparable to that achieved with ground truth \Tx\ locations, highlighting robustness.
% We note that the results are similar  

% \vspace{0.2cm}
\section{Details on Model Training}
\label{sec:trainingdetails}
\begin{figure}[h]
    \centering
    \begin{tabular}{ccc}  % 3 columns (adjust width as needed)
        \includegraphics[width=0.232\textwidth]{figs/eval/main/2_2000.png} & 
        \includegraphics[width=0.232\textwidth]{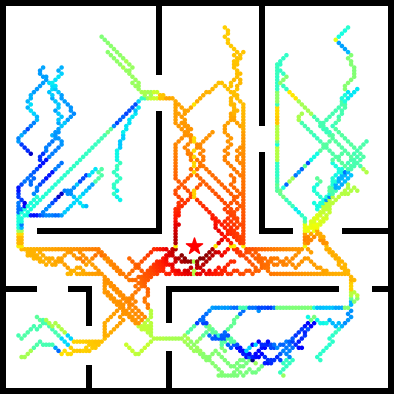} & 
        \includegraphics[width=0.232\textwidth]{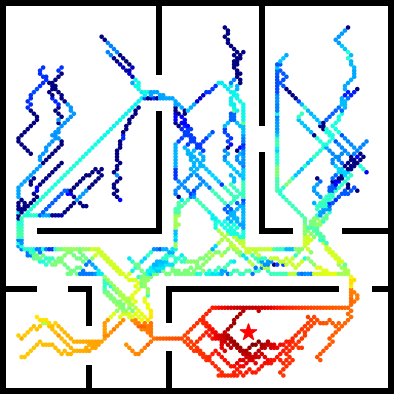} \\
        
        \includegraphics[width=0.232\textwidth]{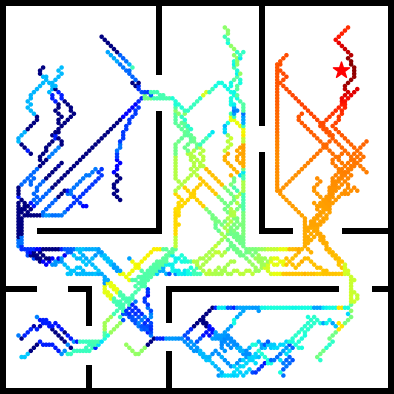} & 
        \includegraphics[width=0.232\textwidth]{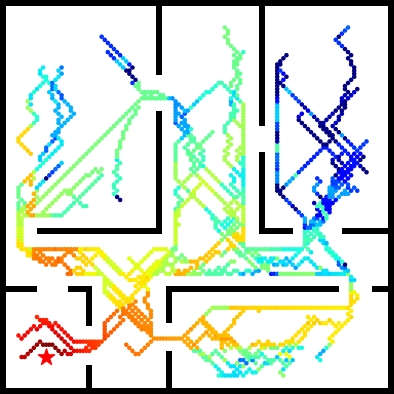} & 
        \includegraphics[width=0.232\textwidth]{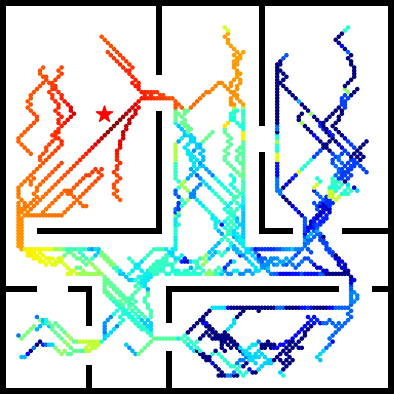} \\
    \end{tabular}
    \caption{Observed signal power at the \Rx s . The top left figure shows the positions of the \Rx s and \Tx s, followed by the power at the receivers from each of the five transmitters. The colormap ranges from red showing stronger signals to blue for weaker signals.}  % Main figure caption
    \label{fig:rssi_plots}
    \vspace{-1em}
\end{figure}

The signal power measured at the receiver is typically represented in a logarithmic scale. RSSI values generally range from -50 dB to -120 dB, where a higher value (e.g., -50 dB) corresponds to a stronger signal. Fig \ref{fig:rssi_plots} illustrates a typical input to \name\, where measurements have been collected from approximately 2000 \Rx s positioned in the floorplan, with data gathered from five \Tx s.
%The log-scaled RSSI values are plotted. As expected, receivers closer to the transmitters receive higher signal power.

\subsection{Linear-Scale RSSI Loss:}
% \noindent\textbf{Linear-Scale RSSI Loss:}
For the training of \name, we optimize on the linear-scale RSSI values. Linear loss ensures that the receivers that capture stronger signals are given more importance during training.
% first convert the log-transformed RSSI values back to their corresponding linear scale. 
% This unnormalization step ensures that the receivers which capture stronger signals are given more importance during training.
We partition our dataset into an 80-20 split, using 80\% of the data for model training, including baselines.

For \name’s network, we employ a simple 8-layer MLP with a hidden dimension of 256 units. For each voxel \(v_j\), the outputs from the final layer are passed through a sigmoid activation to obtain the opacity \(\delta\), and through a Gumbel softmax \cite{gumbel} layer to sample the output normal $\omega$ from one of the possible $K_{\omega}$ orientations.
%and through a softmax-like activation to sample from one of the possible \(K_{\omega}\) orientations for the normal \(\omega\). Specifically, we use the Gumbel softmax layer to sample the orientation from the output of the MLP. 
This sampled orientation is then used in the subsequent stages of training, such as for calculating the direction of the reflected signal $M$. For the learnable baselines, such as \MLP\ and \NeRFs, we adopt the same architecture as used in \name.

\subsection{Supervising Voxel Orientations}
% \noindent\textbf{Supervising Voxel Orientations:} 
\name\ leverages the spatial gradient of a voxel's opacity, \(\nabla \delta\), to supervise its orientation during the multi-stage training process. To compute this gradient, we evaluate the opacities of neighboring voxels along each of the \(K_{\omega}\) directions and apply finite difference methods. We found that this approach yielded superior results compared to using the gradient available via autograd.

\begin{figure}[!t]
\centering    \includegraphics[width=0.6\columnwidth]{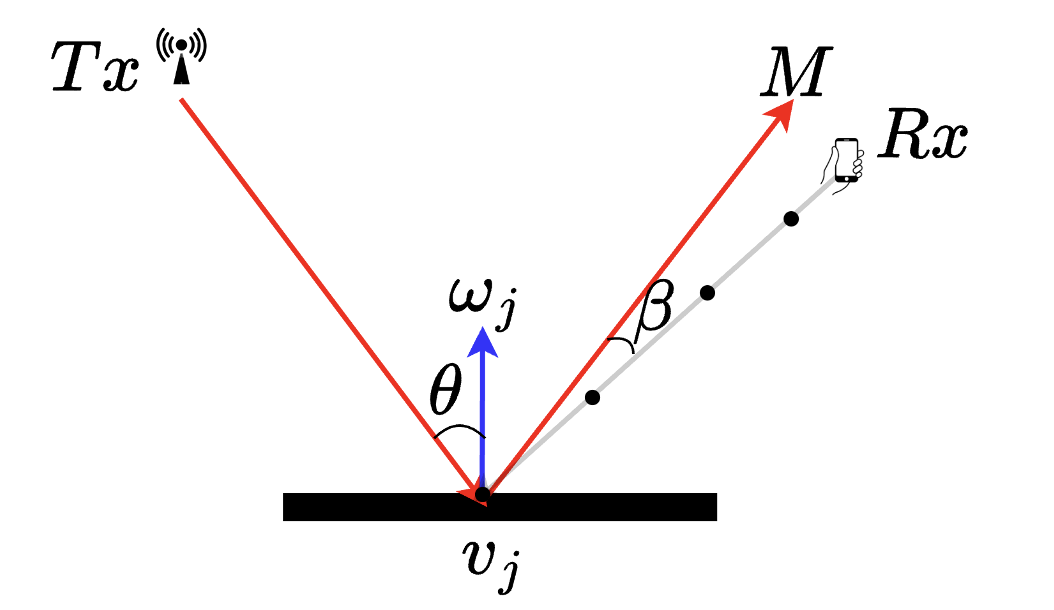}
\caption{An incoming ray from a transmitter {\Tx} reflecting around voxel $v_j$ and arriving at receiver {\Rx}. The incoming ray makes an incident angle $\theta$ with the normal $\omega_j$ to the reflecting surface. The ray after reflection passes a receiver \Rx\ at a certain distance making an angle $\beta$.}
\vspace{-1.5em}
\label{fig:supp-reflection}
\end{figure}

In general, the power from reflections depends not only on the total distance traveled but also on the angle at which the reflection occurs at the voxel \(v_j\), and whether the reflected ray reaches the receiver (\Rx). We model this behavior through $\theta$ and $\beta$ respectively which parameterize a nonlinear function $f$. The incidence angle $\theta$ measures the angle between the \Tx\ and the orientation $\omega_j$, and $\beta$ denotes the angle \Rx\ makes with the reflected ray $M$ (see fig \ref{fig:supp-reflection}). Note that if $v_j \in \mathcal{V}$, and if $\omega_j$ is correct, $\beta = 0$. The reflected ray $M$ can be computed as shown in Eqn \ref{eqn:Mcomp}.
\begin{align} \label{eqn:Mcomp}
M = (v_j - Tx) - 2 \left[ (v_j - Tx).\omega_j \right]\omega_j
\end{align}
Here $\omega_j$ is a unit vector.

\subsection{Detailed network parameters}
\label{sec:implementation}
We assume the floorplan is an unknown shape inside a $512 \times 512$ grid.
For any ray or ray segment, we uniformly sample $n_r=64$ voxels on it. 
We choose $\lambda_1 = \lambda_2 = 0.01$ for LoS training followed by $\lambda_1 = \lambda_2 = 0.1$ for training the {\name} model. 
We find that discretized opacity values to $\{0,1\}$ improve our LoS model. 
We use the straight-through estimator \cite{straightthrough} to avoid the unavailability of the gradient at the discretization step. To help optimization and to encourage sparsity of the number of reflections, we use only the top-$k$ contributions ($k$=$10$) while training the reflection model. 
We use the ADAM optimizer \cite{adam} with $1.0^{-4}$ learning rate. 
We train our models on NVIDIA A100 GPUs. %We train the LoS model for N epochs and the full modem of M epochs.
\vspace{-0.05in}

\subsection{Heatmap Segmentation Implementation Details}
\label{sec:segmentation}
% \noindent\textbf{\ZYBase\ details:}
The raw trajectory signal power values are first interpolated to obtain a heatmap that provides a smoother representation of the input measurements. Of course, interpolating for regions without any data can lead to incorrect results, especially in larger unseen areas. A rule-based classifier is then applied for segmentation, using two criteria: (1) RSSI values above a threshold to identify potential room areas, and (2) smoothness of the RSSI signal, assessed through the second-order derivative, to ensure continuity within rooms. The initial segmentation is refined using morphological operations (via dilation and erosion) with a 3x3 kernel to smooth rough edges and eliminate small components. Overlapping regions are resolved by comparing gradient magnitudes, followed by additional morphological processing and connected component analysis to obtain the final, refined segmentation.

\vspace{1em}
\noindent\textbf{Code:} We plan to release our code, data, and baselines soon. 
In the meantime, Section~\ref{sec:trainingdetails} provides sufficient details to allow readers to reproduce our results, especially since our network components are simple MLPs.

\clearpage

\onecolumn
\section{Evaluation on Additional Floorplans}

% We show more results of our method on additional floorplans and compare them with other baselines.  

%\begin{minipage}[c]{\textwidth}
\begin{figure*}[!h]
  \centering
    \begin{tabular}{ @{\hskip 5pt} l @{\hskip 5pt} c@{ } c @{ } c @{ } c @{ } c @{ } c }
    % \hline
    \parbox[c][0.6cm][c]{0.1\textwidth}{\centering \vspace{-2cm}Ground Truth
    } & 
    \includegraphics[width=0.14\textwidth]{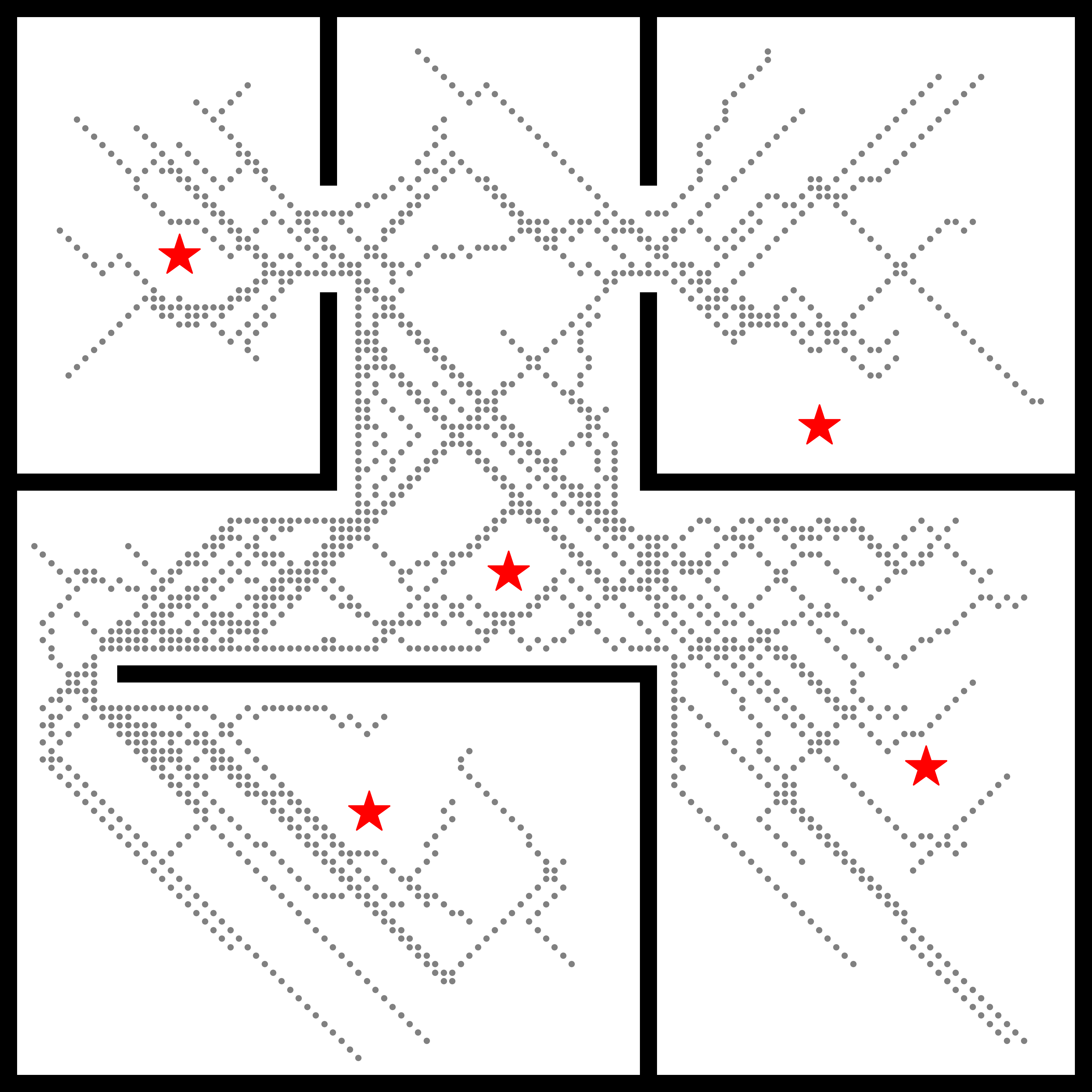}&
    \includegraphics[width=0.14\textwidth]{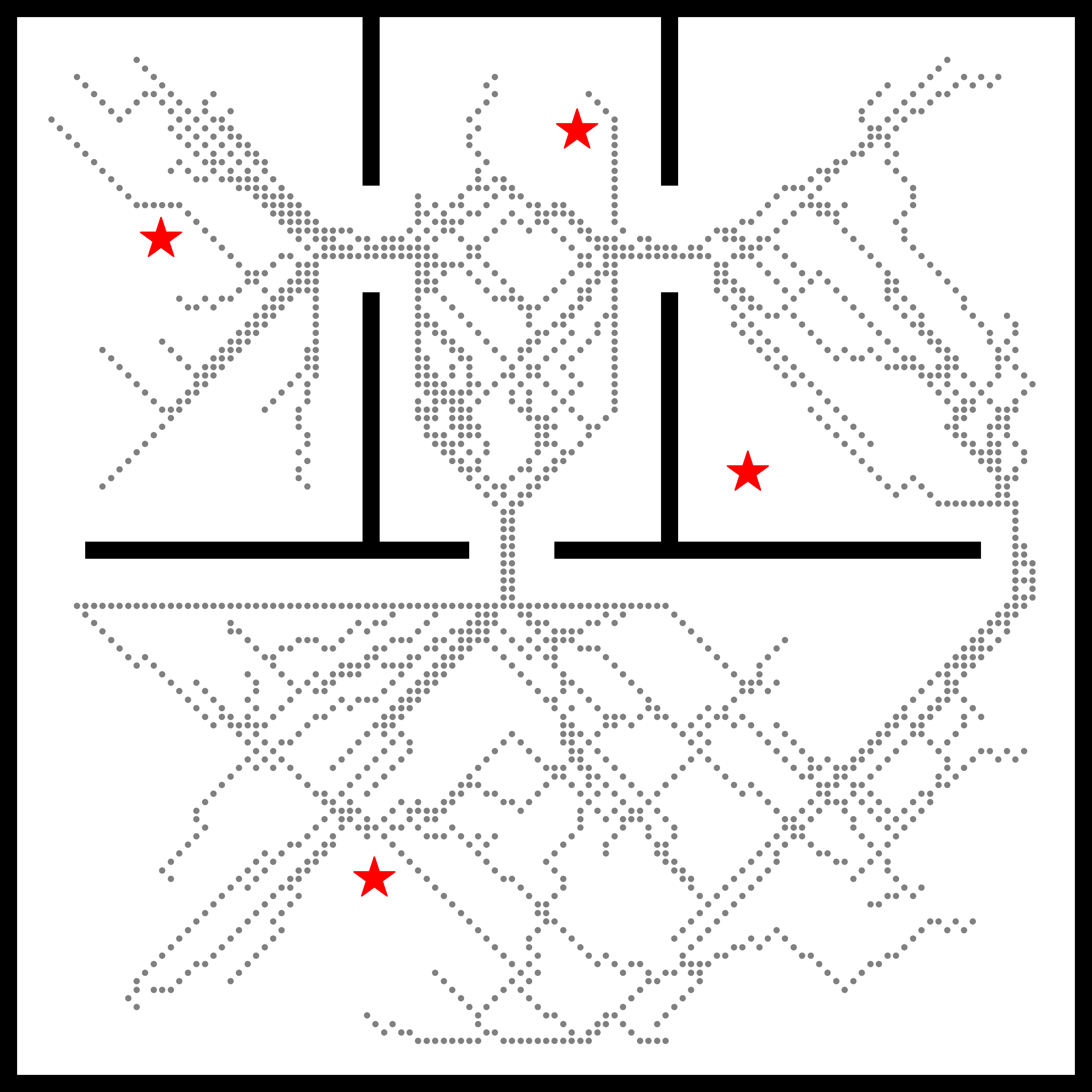}&
    \includegraphics[width=0.14\textwidth]{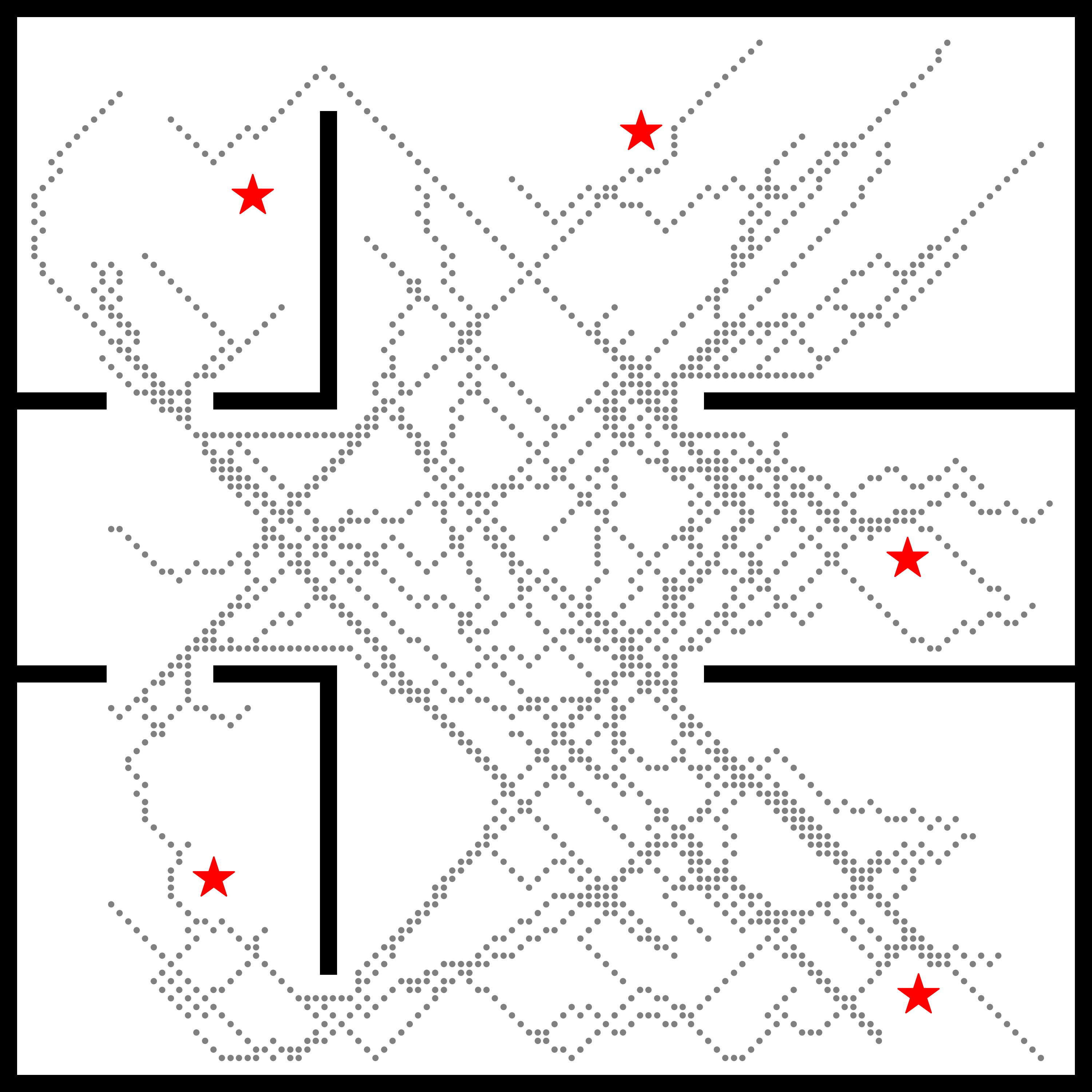}&
    \includegraphics[width=0.14\textwidth]{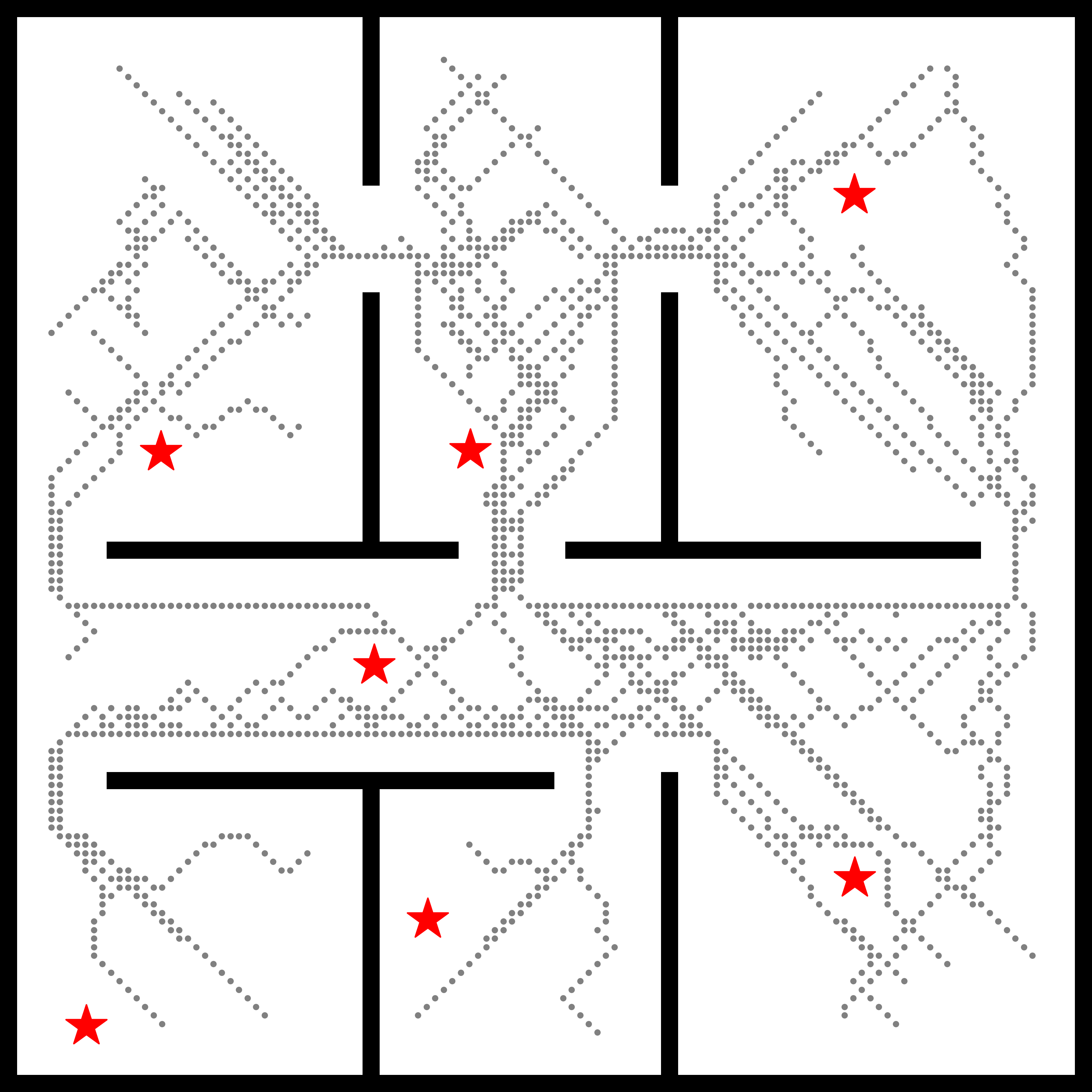}& 
    \includegraphics[width=0.14\textwidth]{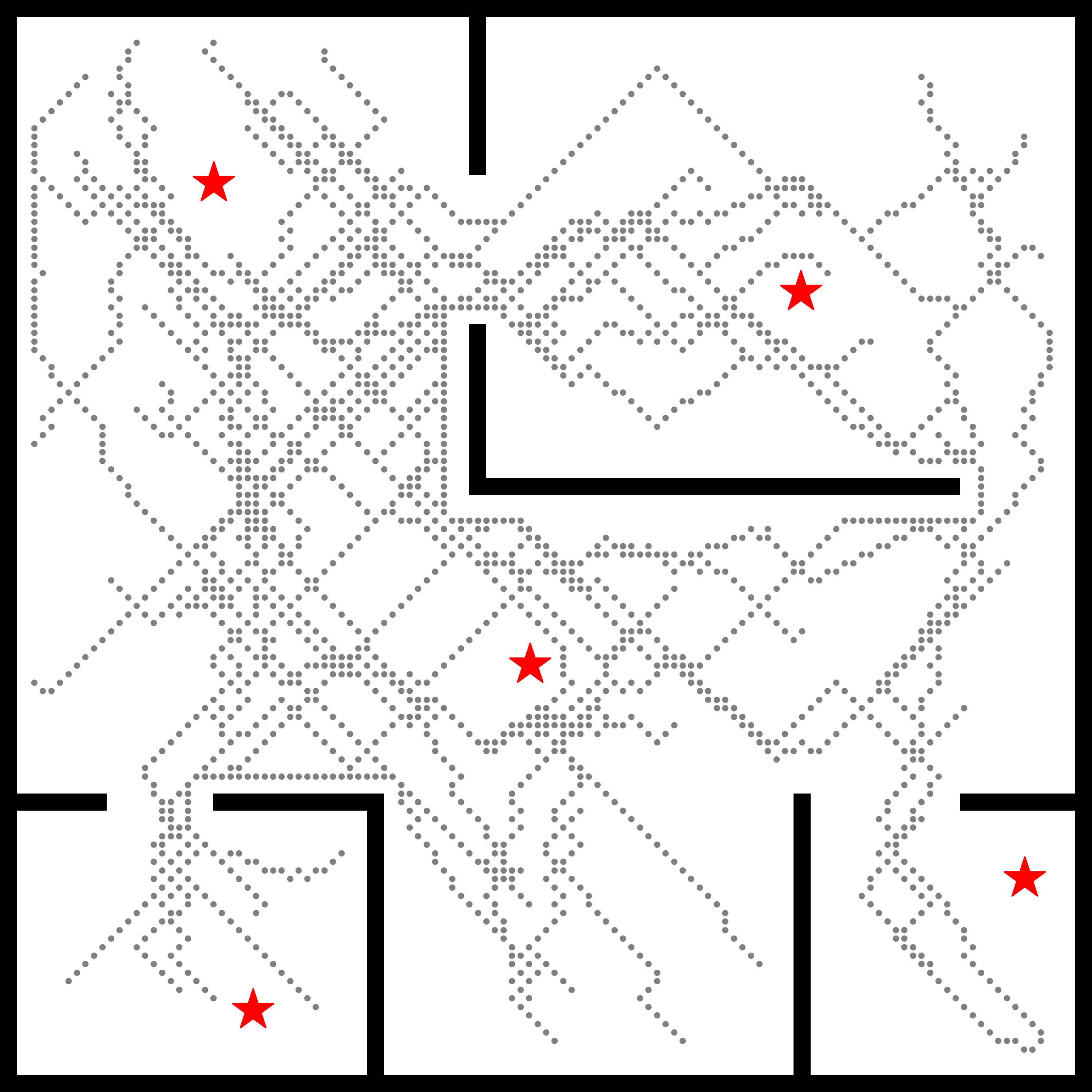}&
    \includegraphics[width=0.14\textwidth]{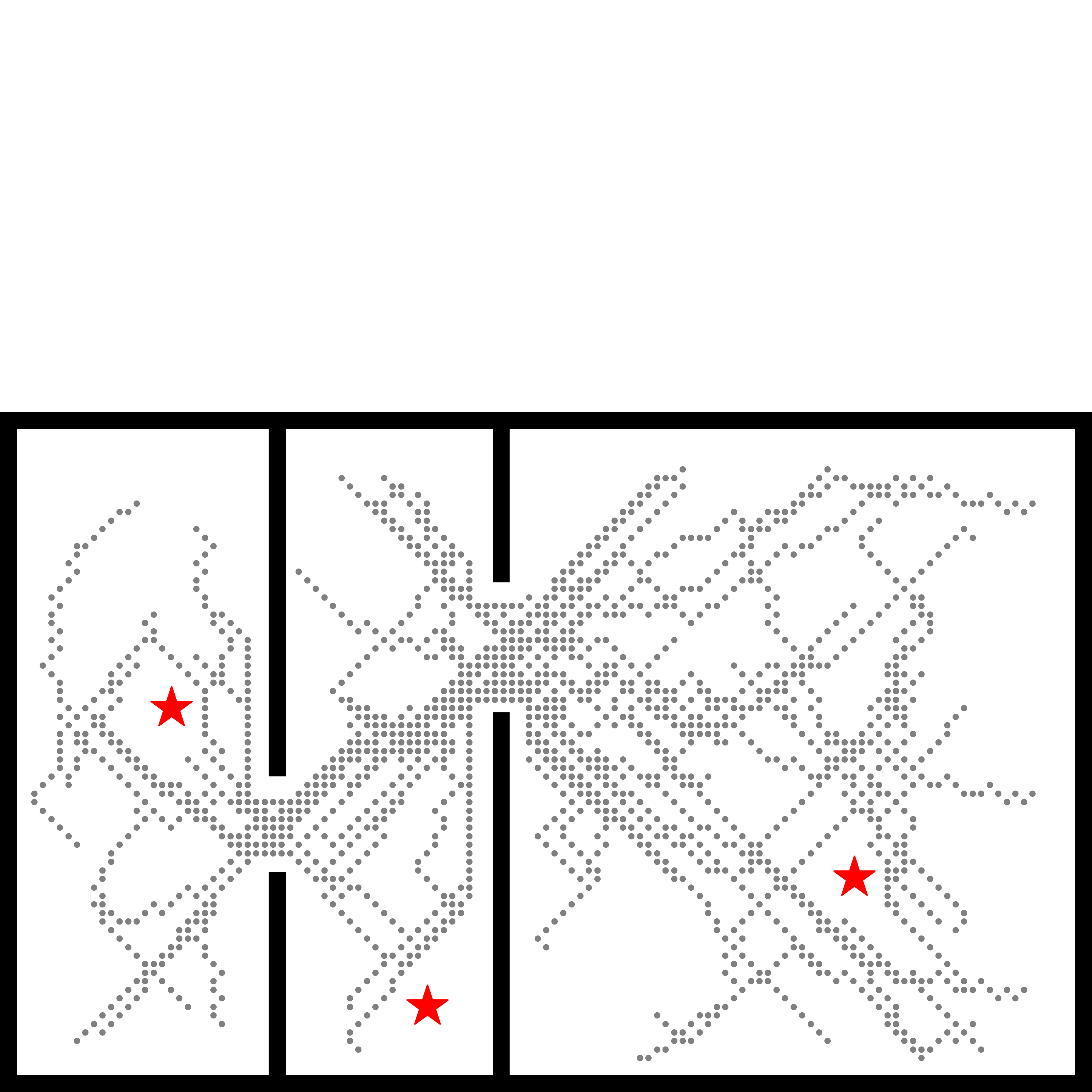}
    \vspace{0.3cm}
    \\
    %\hline
    \parbox[c][.6cm][c]{0.1\textwidth}{\centering \vspace{-2cm} \texttt{Heatmap Seg.}} & 
    \includegraphics[width=0.14\textwidth]{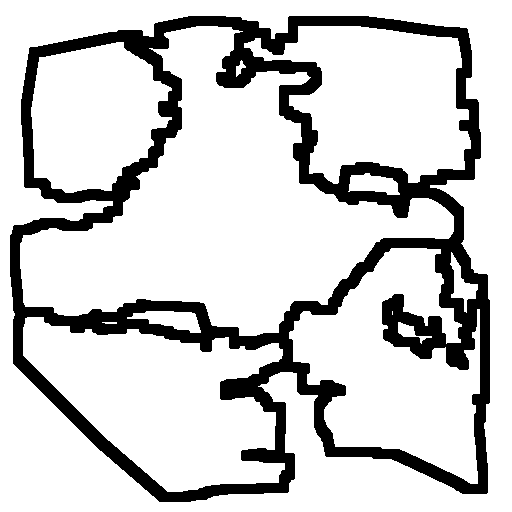}&
    \includegraphics[width=0.14\textwidth]{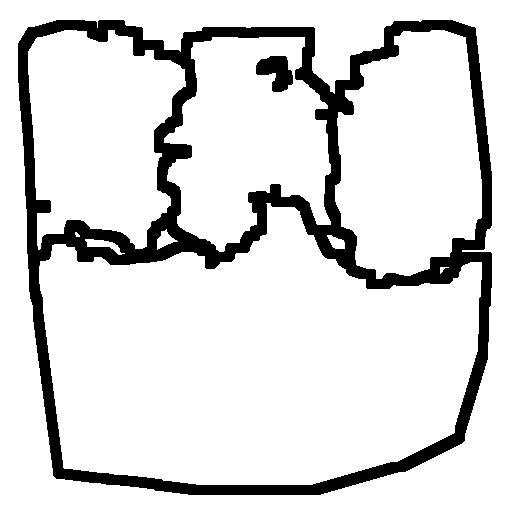}&
    \includegraphics[width=0.14\textwidth]{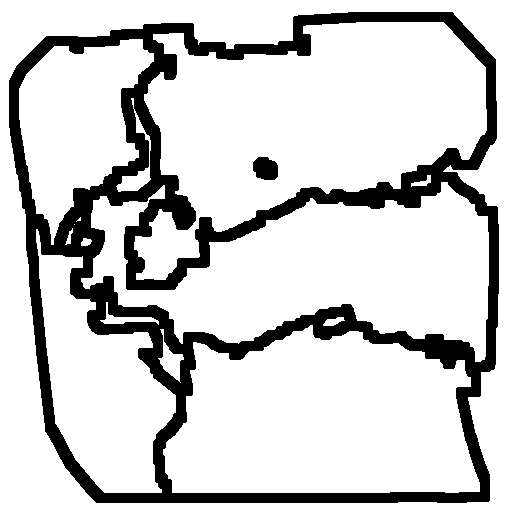}&
    \includegraphics[width=0.14\textwidth]{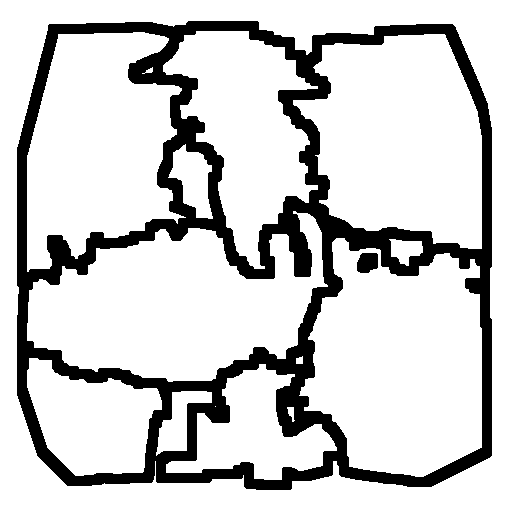}&
    \includegraphics[width=0.14\textwidth]{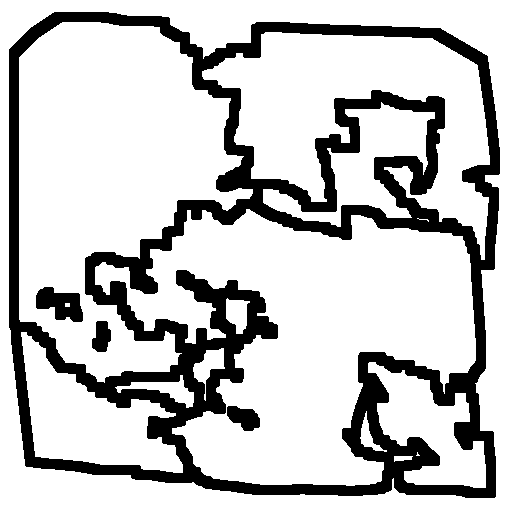}&
    \includegraphics[width=0.14\textwidth]{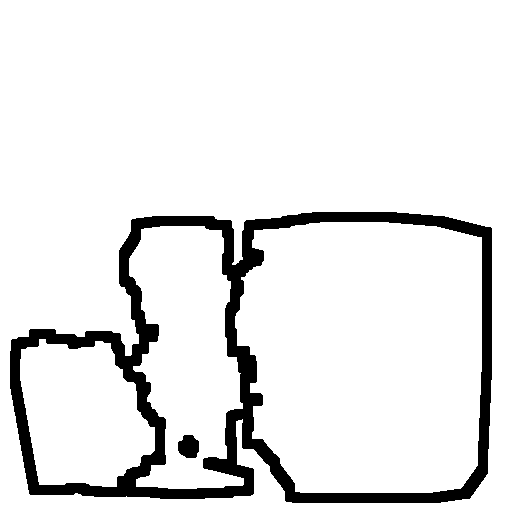}
    \\
    \parbox[c][.6cm][c]{0.1\textwidth}{\centering \vspace{-2cm} \NeRFs} & 
    \includegraphics[width=0.14\textwidth]{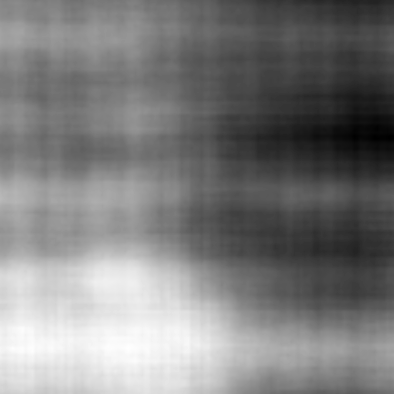}&
    \includegraphics[width=0.14\textwidth]{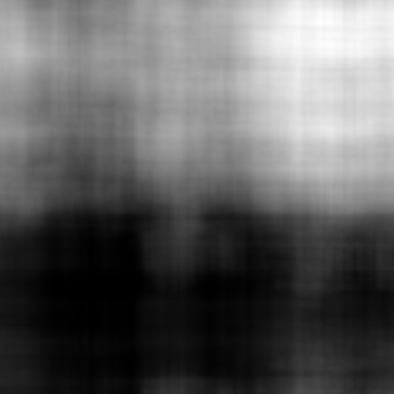}&
    \includegraphics[width=0.14\textwidth]{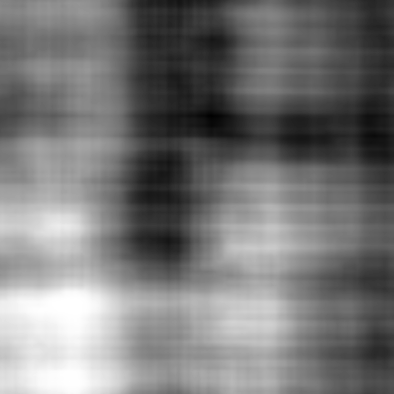}&
    \includegraphics[width=0.14\textwidth]{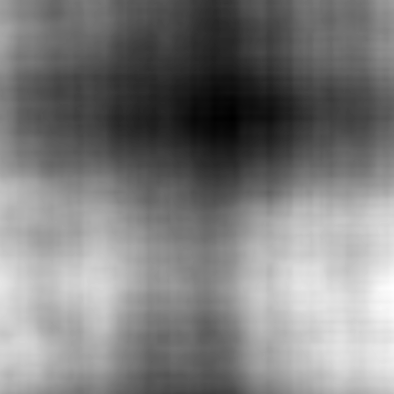}&
    \includegraphics[width=0.14\textwidth]{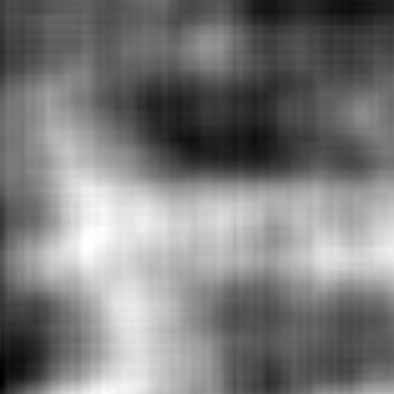}&
    \includegraphics[width=0.14\textwidth]{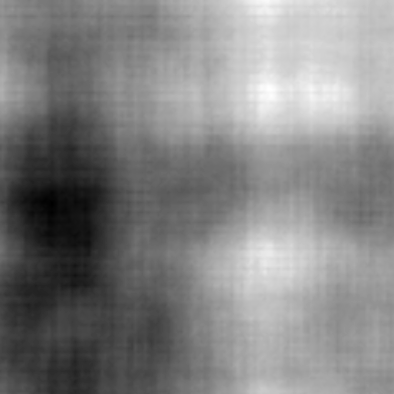}
    \\
    \parbox[c][1.2cm][c]{0.1\textwidth}{\centering \vspace{-2cm} \texttt{EchoNeRF LoS}} & 
    \includegraphics[width=0.14\textwidth]{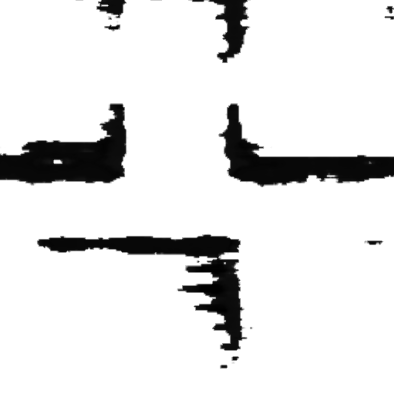}&
    \includegraphics[width=0.14\textwidth]{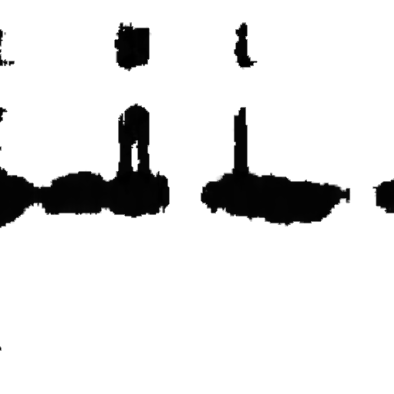}&
    \includegraphics[width=0.14\textwidth]{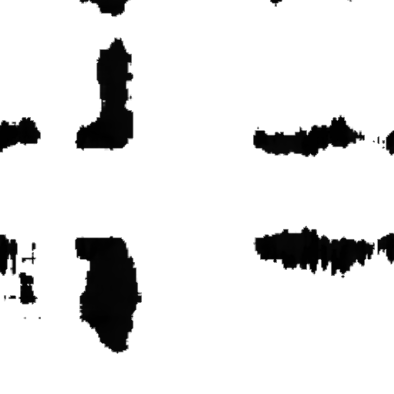}&
    \includegraphics[width=0.14\textwidth]{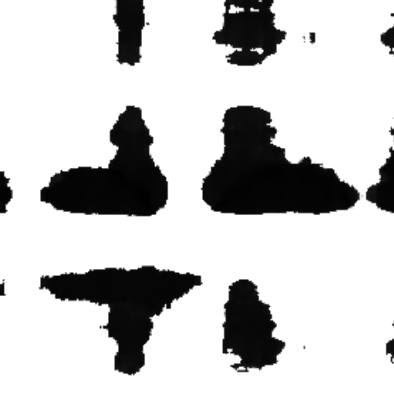}&
    \includegraphics[width=0.14\textwidth]{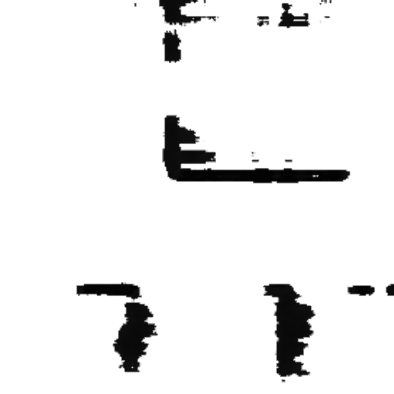}&
    \includegraphics[width=0.14\textwidth]{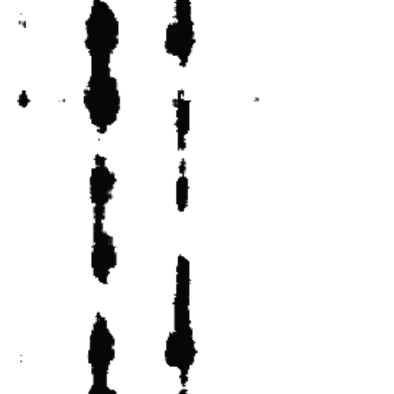}
    \\
    \parbox[c][.6cm][c]{0.1\textwidth}{\centering \vspace{-2cm} \name} &
    \includegraphics[width=0.14\textwidth]{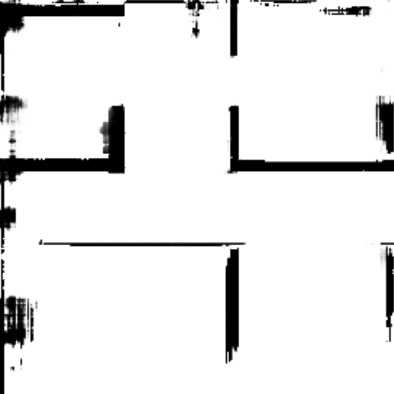}&
    \includegraphics[width=0.14\textwidth]{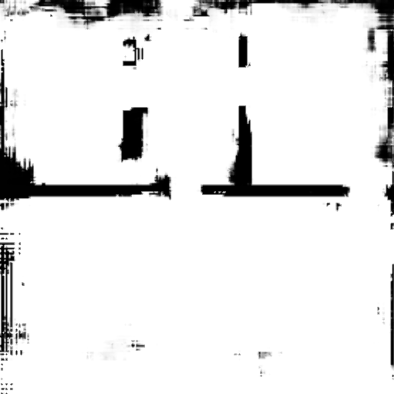}&
    \includegraphics[width=0.14\textwidth]{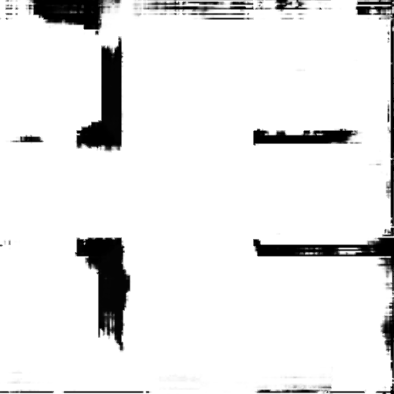}&
    \includegraphics[width=0.14\textwidth]{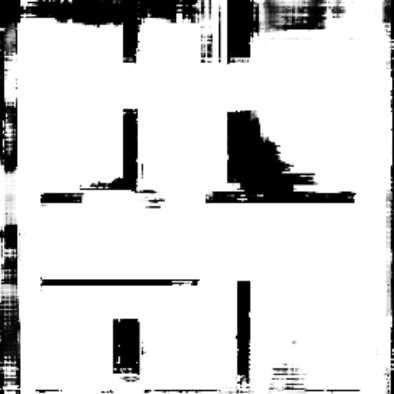}&
    \includegraphics[width=0.14\textwidth]{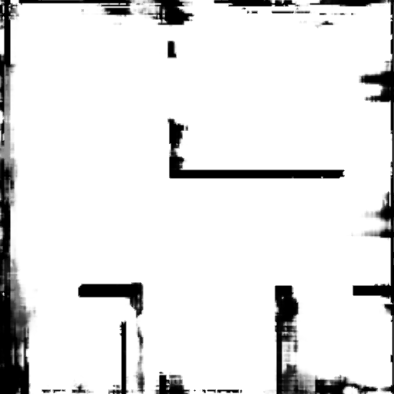}&
    \includegraphics[width=0.14\textwidth]{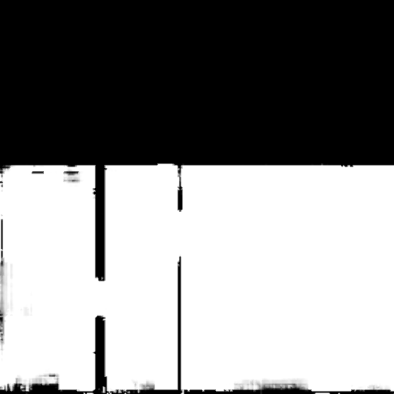}
    \vspace{1cm}
    \\
    % \hline
    % \vspace{1cm}
    \parbox[c][.6cm][c]{0.1\textwidth}{\centering \vspace{-2cm} \name\ Signal Prediction} &
    \includegraphics[width=0.14\textwidth]{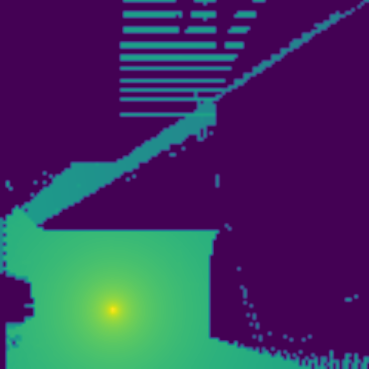}&
    \includegraphics[width=0.14\textwidth]{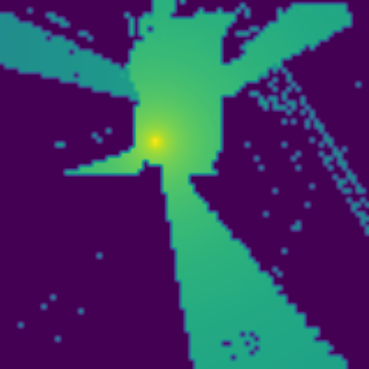}&
    \includegraphics[width=0.14\textwidth]{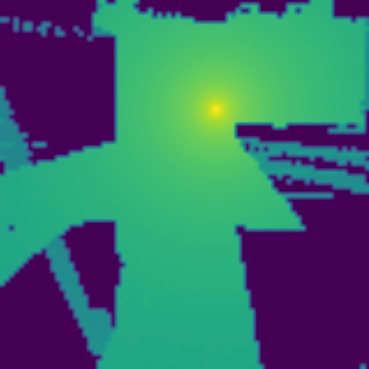}&
    \includegraphics[width=0.14\textwidth]{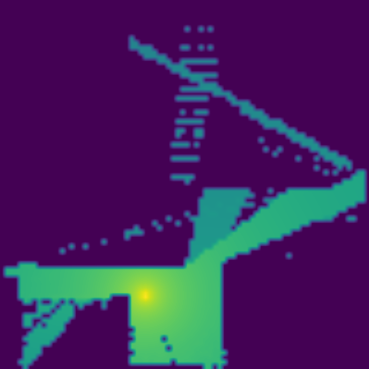}&
    \includegraphics[width=0.14\textwidth]{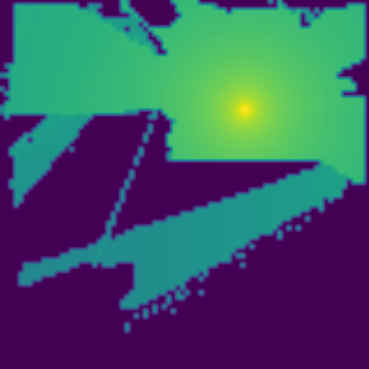}&
    \includegraphics[width=0.14\textwidth]{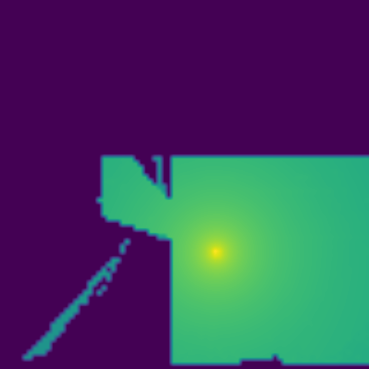}
  \end{tabular}
  % \vspace{-0.15in}
  \caption{Qualitative comparison of Ground Truth floorplans against those inferred by baselines
  {\ZYBase} and {\NeRFs}. We note that while \NeRFs\ is unable to predcit any reasonable floorplan, \ZYBase\'s shape is limited by the (convex hull of the) trajectory data (see bottom left of the second row, first column). Additionally, it fails to capture critical details, such as door openings. The 4th and 5th rows show floorplans by our proposed models {\nameLoS} and {\name}. \nameLoS\ captures the rough shape of the floorplan, especially the interior walls, while \name\ further improves these walls by adjusting their thickness and accurately correcting their shape. \name\ also correctly identifies the floorplan boundary, as evidenced in the last column, where the exact boundary is captured just from the reflections. To understand the signal propagation captured by \name, we place one {\Tx} in each floor plan randomly (that is not present in the training data) and evaluate the signal power at discrete receivers. These {\Rx}s are placed on a 2D grid at equal intervals and the predicted signal power is converted into a heatmap. The bottom row shows these inferred signal power heatmaps with  the brightest point indicating the Tx location (as the {\Rx} closest to the {\Tx} receives the highest power). \name\ is not only able to predict the signals well across the floorplan, but also capture the propagation paths i.e., LoS signal and the first-order reflections. For instance, in the first column, the left portion of the center hall receives power only due to the wall reflection from the left wall.}

  \label{fig:supp-res1}
  % \vspace{-0.1in}
\end{figure*}

\begin{figure*}[t]
  \centering
    \begin{tabular}{ @{\hskip 5pt} l @{\hskip 5pt} c@{ } c @{ } c @{ } c @{ } c @{ } c }
    % \hline
    \parbox[c][0.6cm][c]{0.1\textwidth}{\centering \vspace{-2cm}Ground Truth
    } & 
    \includegraphics[width=0.14\textwidth]{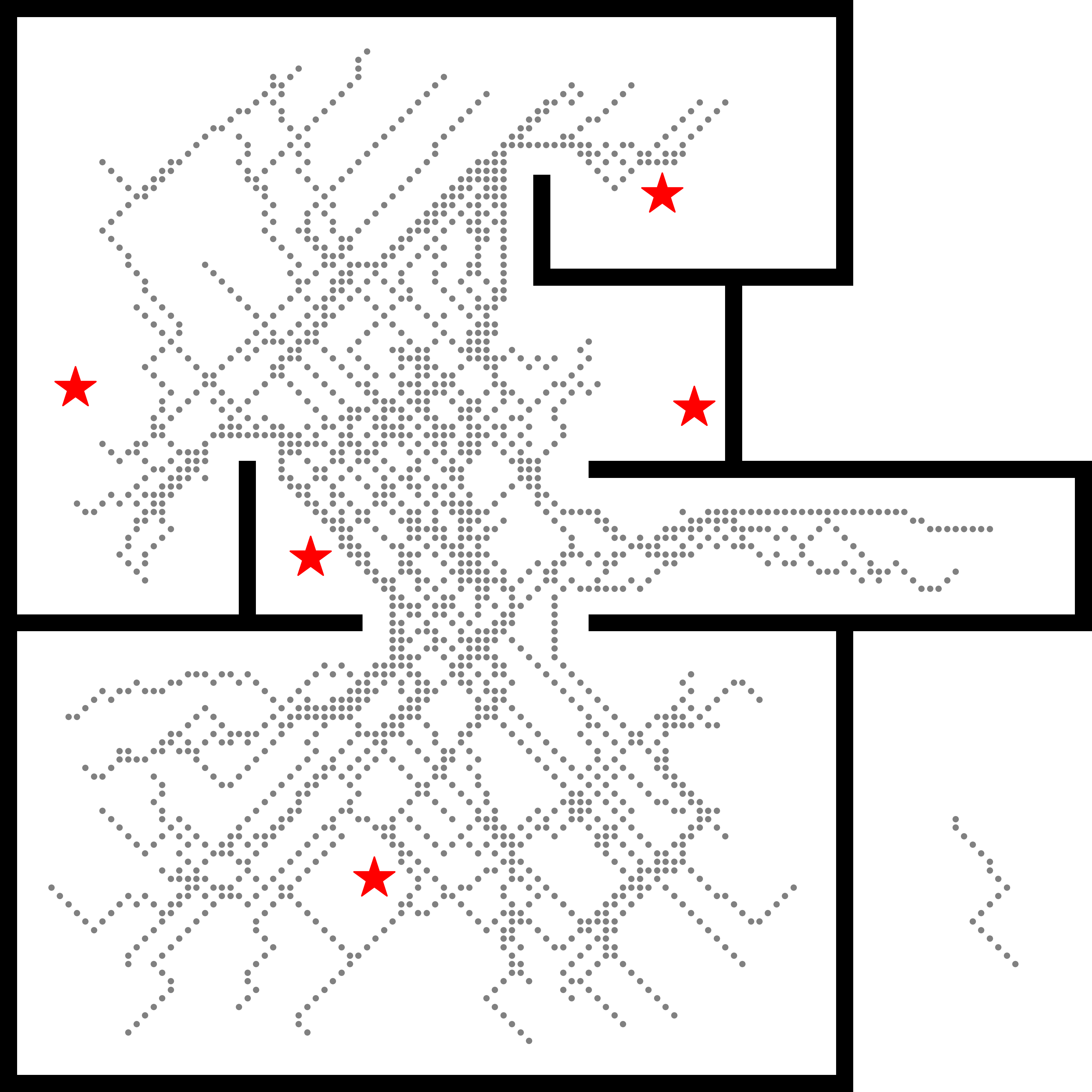}& 
    \includegraphics[width=0.14\textwidth]{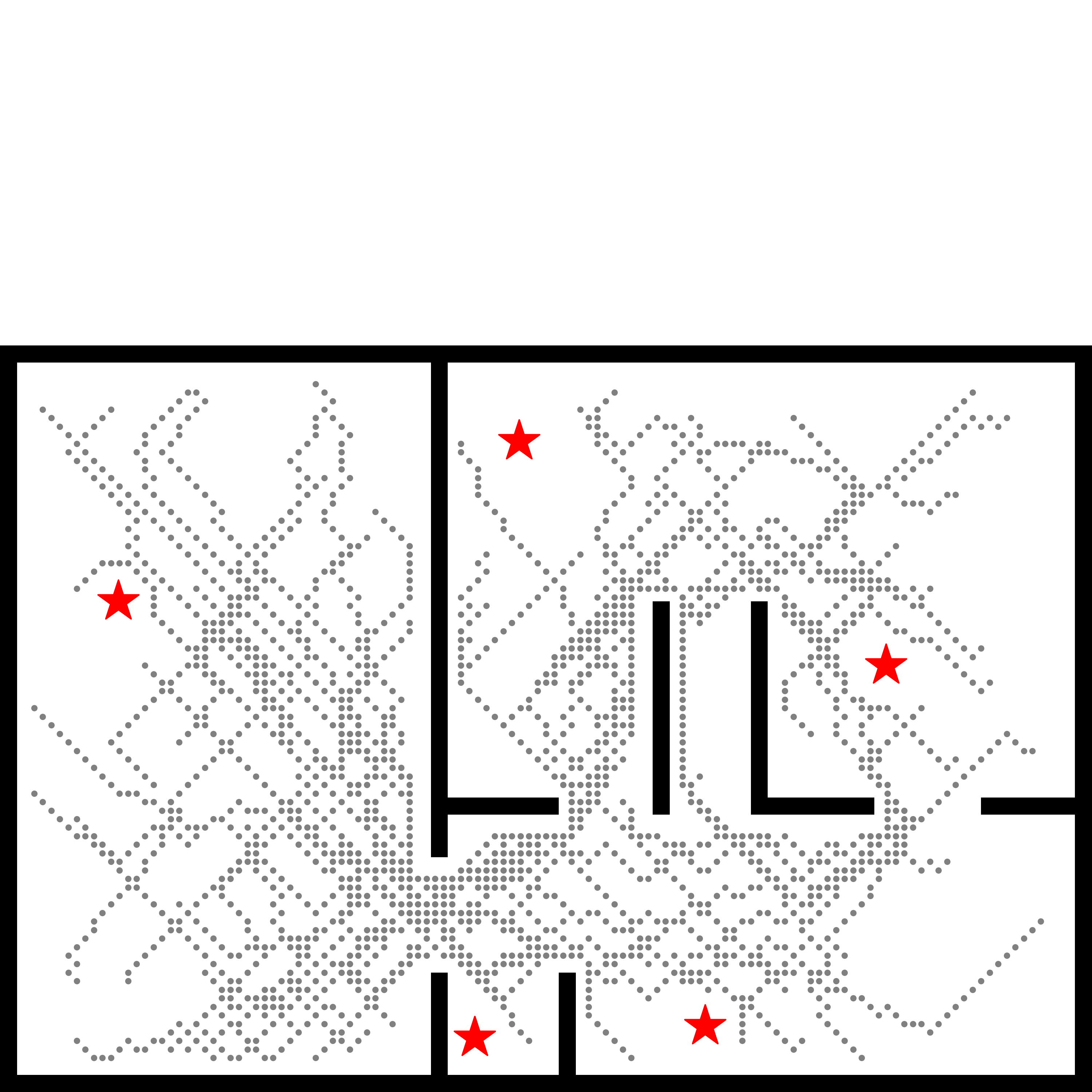}&
    \includegraphics[width=0.14\textwidth]{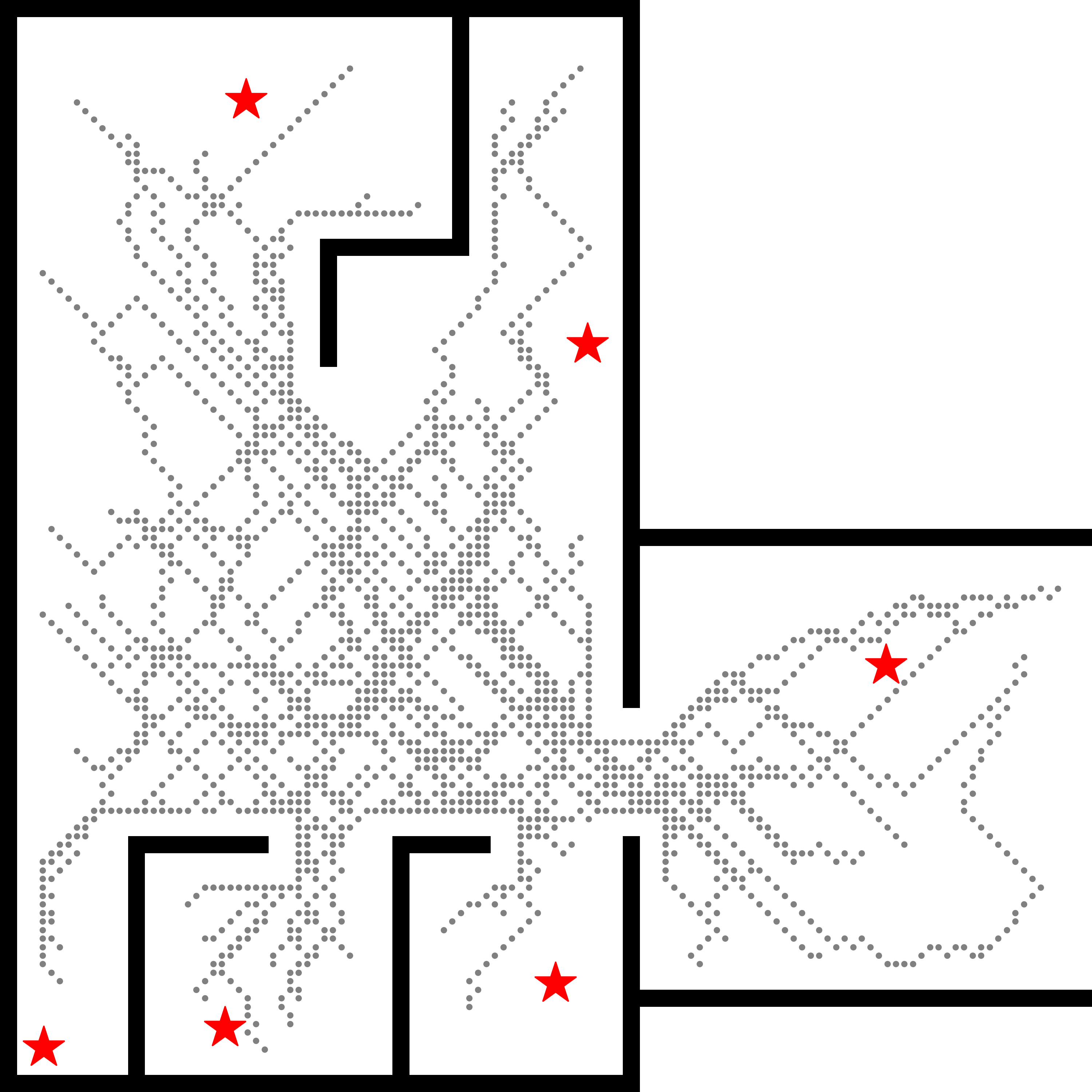}&
    \includegraphics[width=0.14\textwidth]{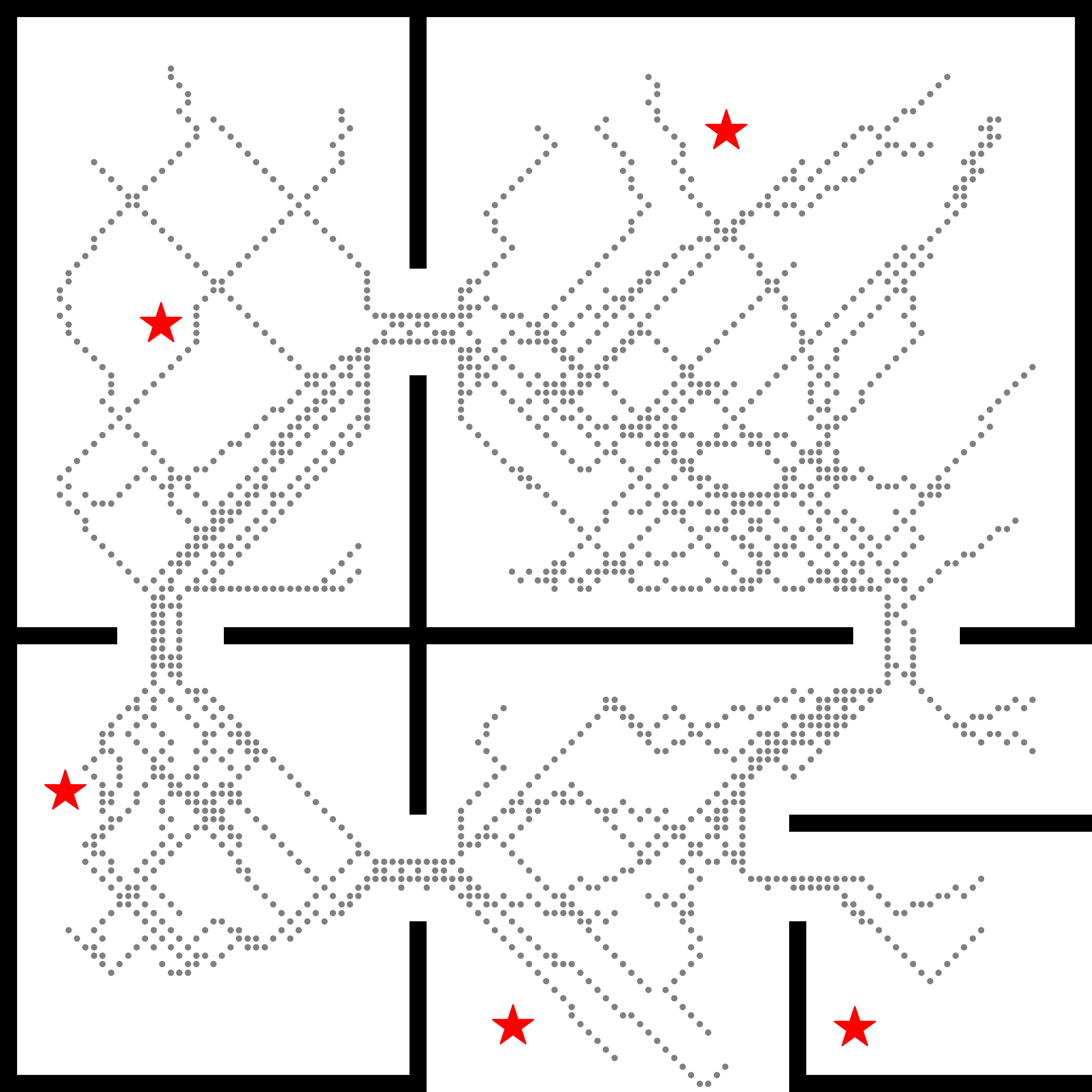}&
    \includegraphics[width=0.14\textwidth]{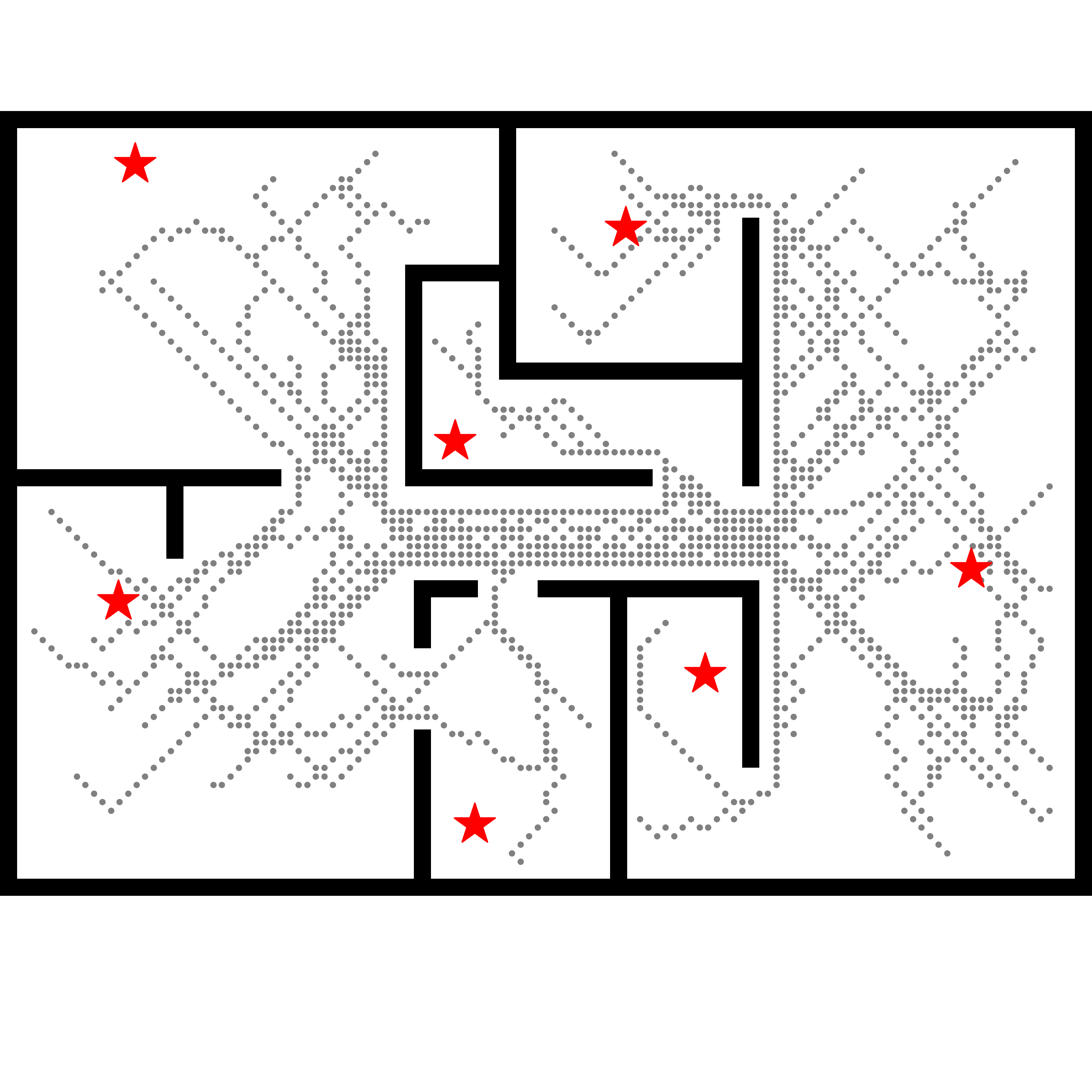}&
    \includegraphics[width=0.14\textwidth]{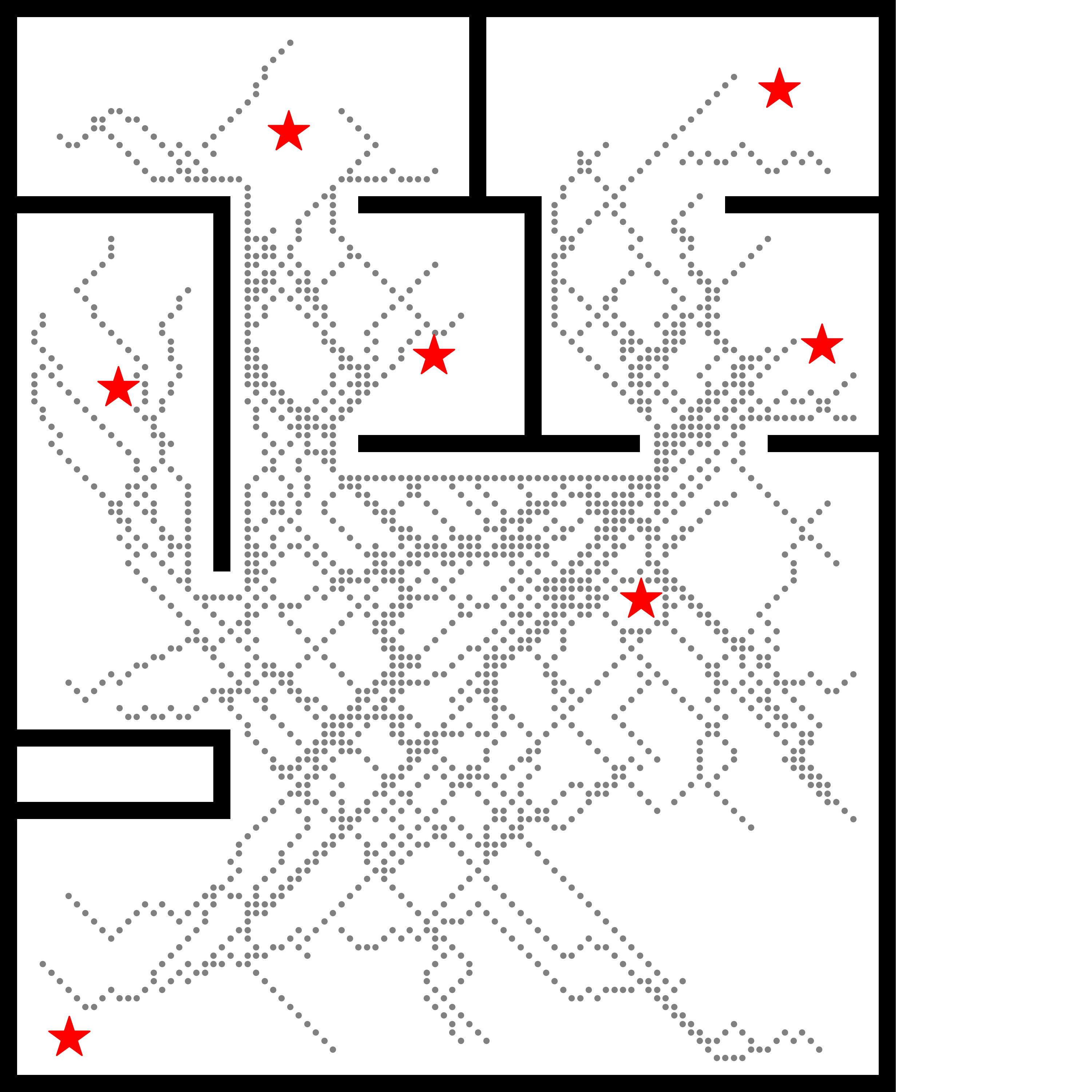}
    \vspace{0.3cm}
    \\
    %\hline
    \parbox[c][.6cm][c]{0.1\textwidth}{\centering \vspace{-2cm} \texttt{Heatmap Seg.}} & 
    \includegraphics[width=0.14\textwidth]{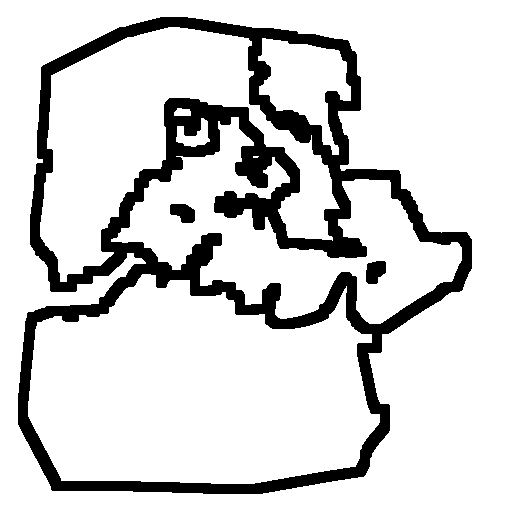}&
    \includegraphics[width=0.14\textwidth]{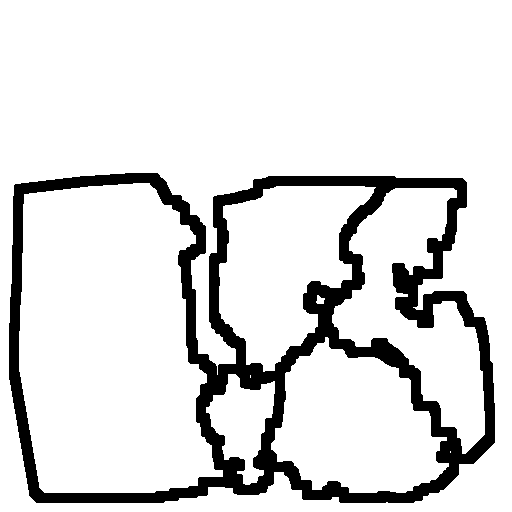}&
    \includegraphics[width=0.14\textwidth]{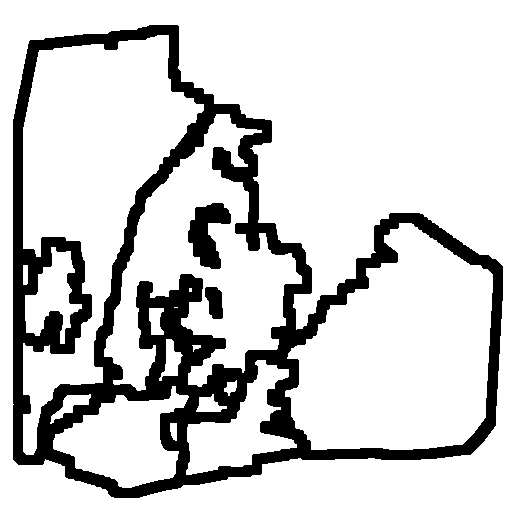}&
    \includegraphics[width=0.14\textwidth]{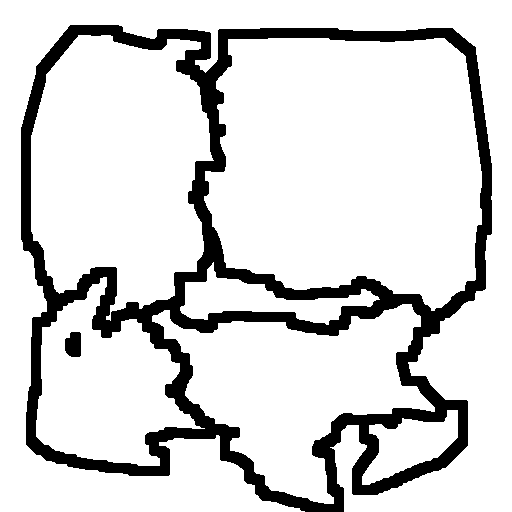}&
    \includegraphics[width=0.14\textwidth]{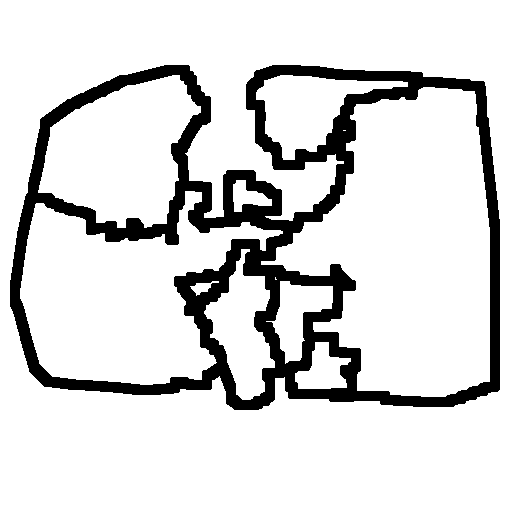}&
    \includegraphics[width=0.14\textwidth]{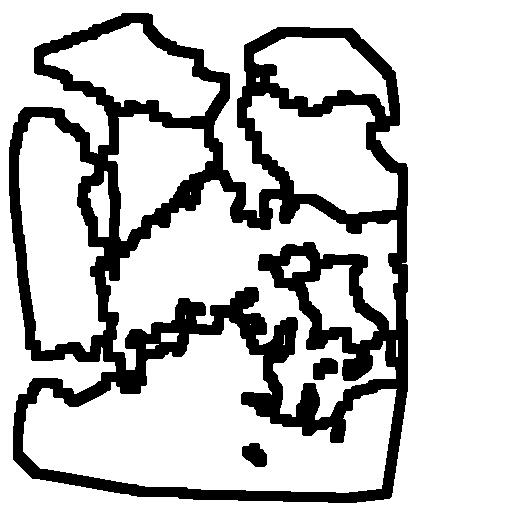}
    \\
    \parbox[c][.6cm][c]{0.1\textwidth}{\centering \vspace{-2cm} \NeRFs} & 
    \includegraphics[width=0.14\textwidth]{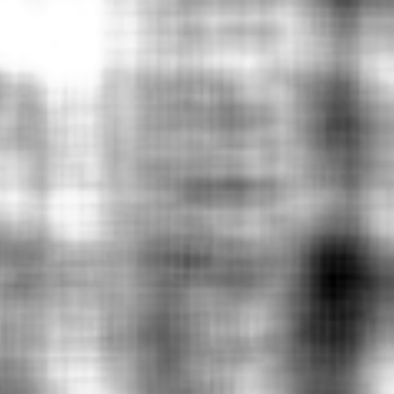}&
    \includegraphics[width=0.14\textwidth]{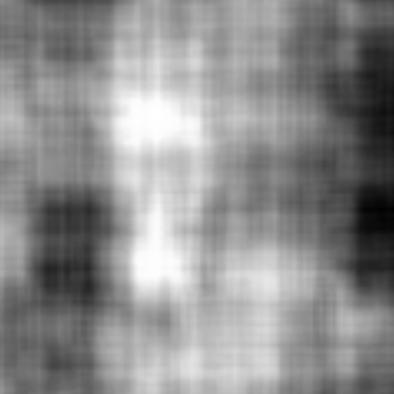}&
    \includegraphics[width=0.14\textwidth]{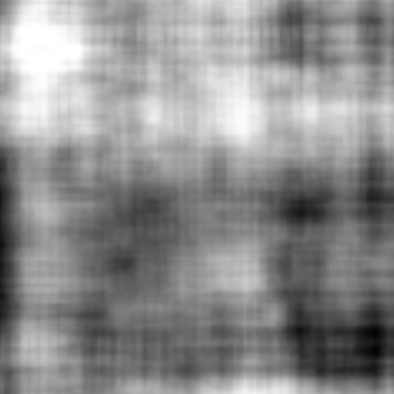}&
    \includegraphics[width=0.14\textwidth]{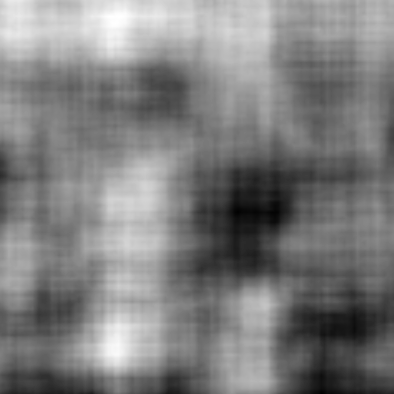}&
    \includegraphics[width=0.14\textwidth]{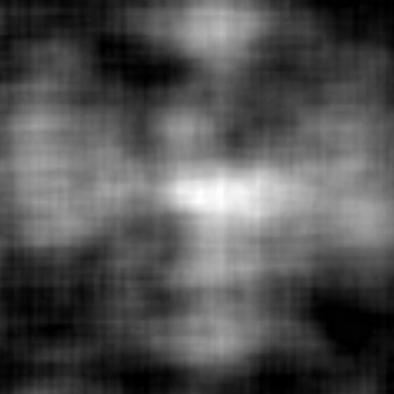}&
    \includegraphics[width=0.14\textwidth]{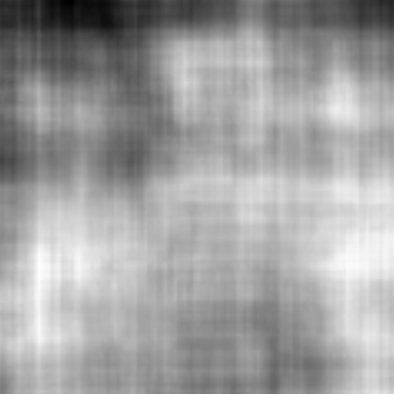}
    \\
    \parbox[c][1.2cm][c]{0.1\textwidth}{\centering \vspace{-2cm} {\texttt{EchoNeRF LoS}}} & 
    \includegraphics[width=0.14\textwidth]{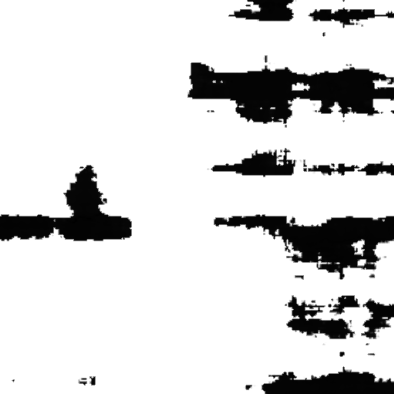}&
    \includegraphics[width=0.14\textwidth]{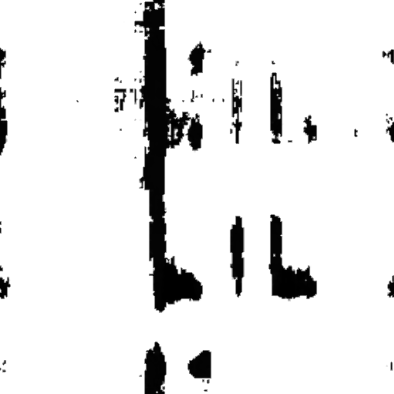}&
    \includegraphics[width=0.14\textwidth]{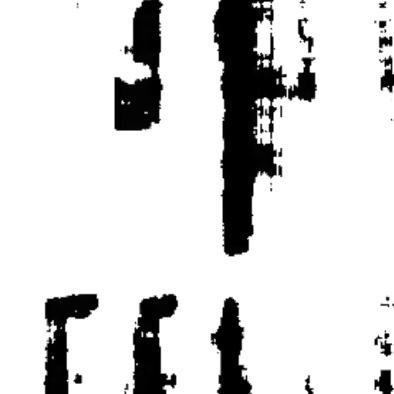}&
    \includegraphics[width=0.14\textwidth]{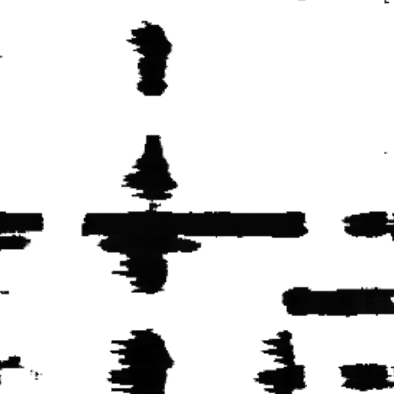}&
    \includegraphics[width=0.14\textwidth]{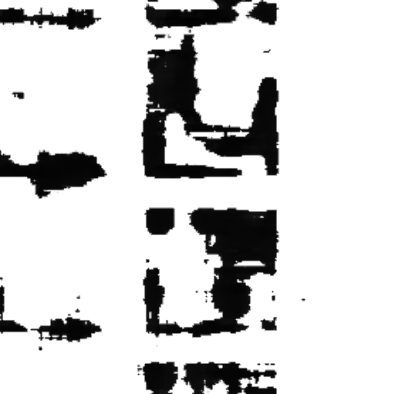}&
    \includegraphics[width=0.14\textwidth]{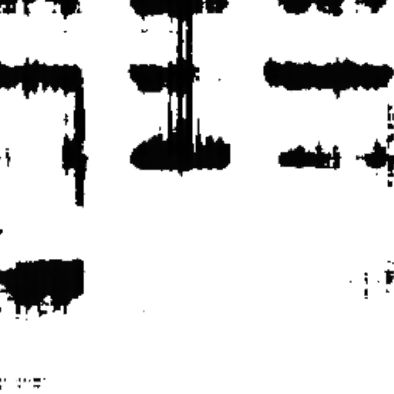}
    \\
    \parbox[c][.6cm][c]{0.1\textwidth}{\centering \vspace{-2cm} \name} &
    \includegraphics[width=0.14\textwidth]{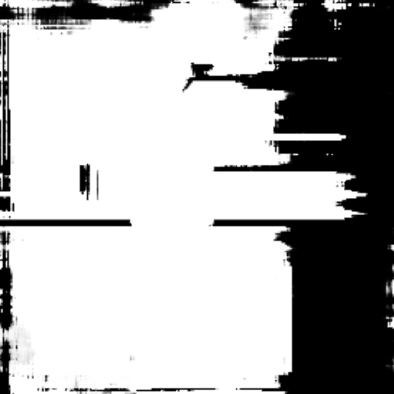}&
    \includegraphics[width=0.14\textwidth]{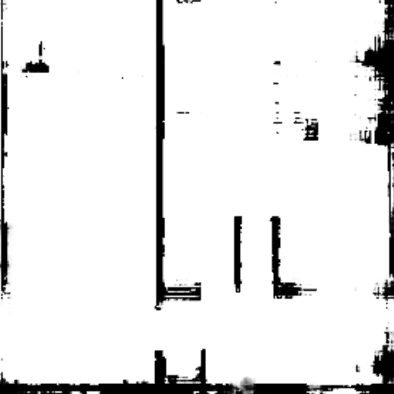}&
    \includegraphics[width=0.14\textwidth]{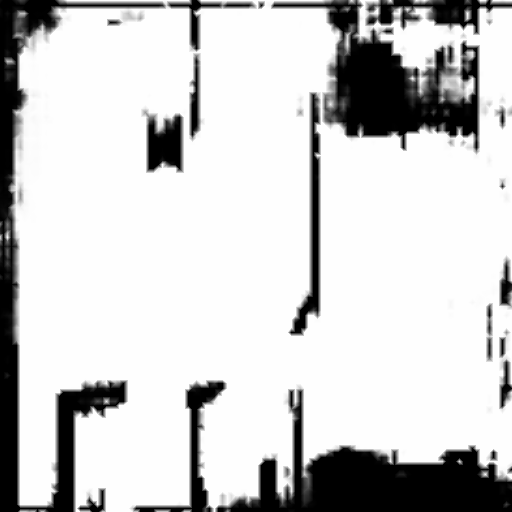}&
    \includegraphics[width=0.14\textwidth]{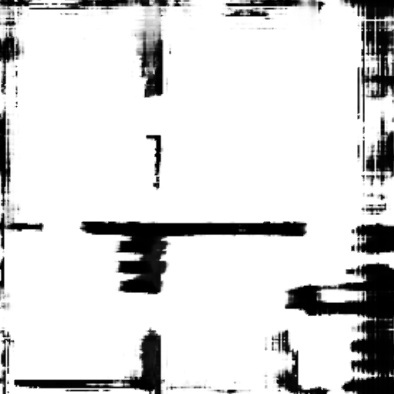}&
    \includegraphics[width=0.14\textwidth]{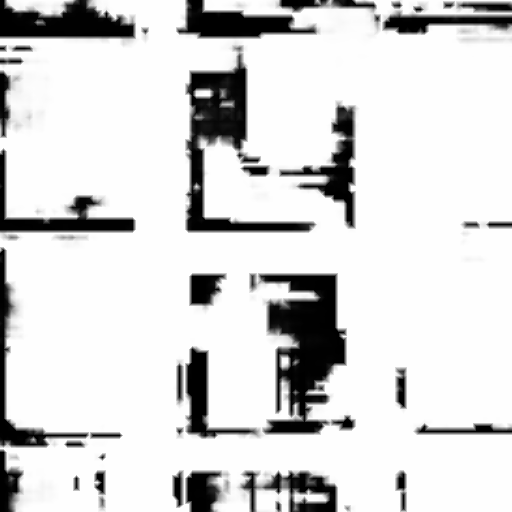}&
    \includegraphics[width=0.14\textwidth]{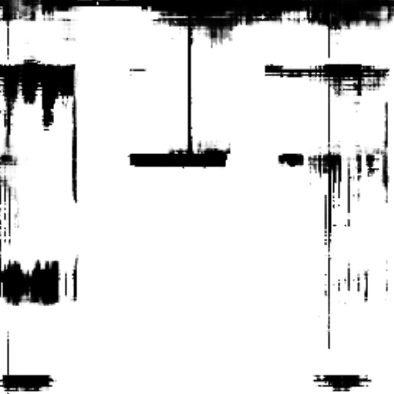}
    \vspace{1cm}
    \\
    \parbox[c][.6cm][c]{0.1\textwidth}{\centering \vspace{-2cm} \name\ Signal Prediction} &
    \includegraphics[width=0.14\textwidth]{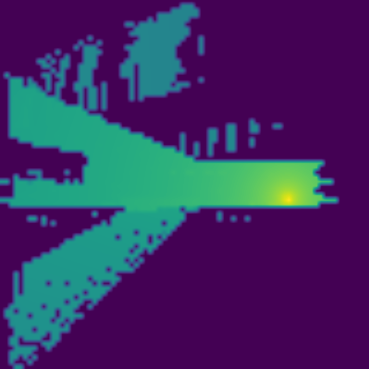}&
    \includegraphics[width=0.14\textwidth]{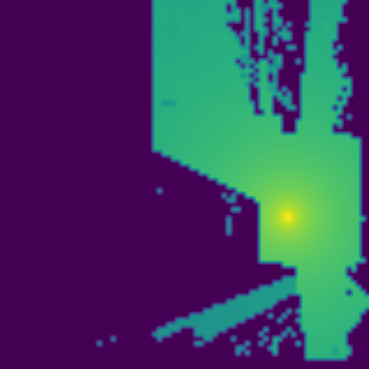}&
    \includegraphics[width=0.14\textwidth]{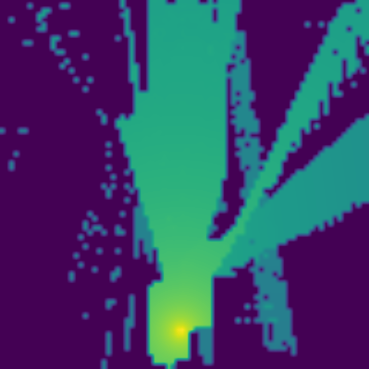}&
    \includegraphics[width=0.14\textwidth]{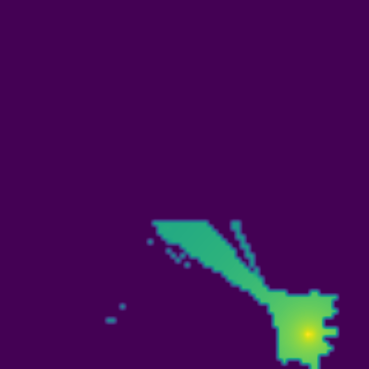}&
    \includegraphics[width=0.14\textwidth]{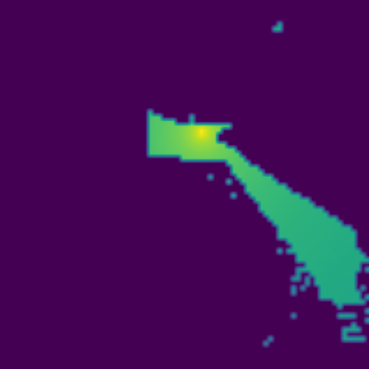}&
    \includegraphics[width=0.14\textwidth]{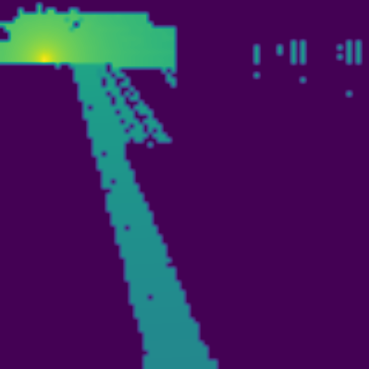}
  \end{tabular}
  \vspace{-0.15in}
  \caption{Additional qualitative comparisons of Ground Truth floorplans against those inferred by baselines
  {\ZYBase} and {\NeRFs}. The 4th and 5th rows show floorplans by our proposed models {\nameLoS} and {\name} with clearly identified walls and boundaries. The bottom row shows inferred signal power heatmaps demonstrating \name's capability to learn accurate signal propagation.}
  \label{fig:supp-res2}
  \vspace{-0.1in}
\end{figure*}

\newpage

% \end{document}

\section{Societal Impact Statement}
\label{sec:statement}
We acknowledge that NeRFs hold significant potential for positive societal impact. Applications span AR/VR, medical imaging, airport security, and education, where accurate 3D reconstructions can greatly enhance functionality and understanding.
However, our work on EchoNeRF also introduces potential risks. In particular, the ability to infer detailed spatial layouts from limited sensory input could be misused to access private or sensitive floorplan information. We emphasize the importance of responsible use.

\end{document}